\documentclass[
]{cup-journal}

\usepackage{amsmath}
\usepackage[nopatch]{microtype}
\usepackage{booktabs}
\usepackage[most]{tcolorbox}
\usepackage{listings}
\usepackage[utf8]{inputenc}
\usepackage{subcaption}
\usepackage[T1]{fontenc}
\usepackage[utf8]{inputenc}
\usepackage[T1]{fontenc}
\usepackage{textcomp}
\usepackage{amsfonts}
\usepackage{graphicx}

\DeclareUnicodeCharacter{0302}{^{}}
\DeclareUnicodeCharacter{0307}{.{}}

\tcbset{
colback=green!3,
colframe=black!10,
boxrule=0.4pt,
arc=2pt,
left=6pt,right=6pt,top=6pt,bottom=6pt,
enhanced
}
\newtcblisting{examplebox}{
listing only,
colback=green!3,
colframe=black!10,
boxrule=0.4pt,
arc=2pt,
left=8pt,right=8pt,top=6pt,bottom=6pt,
listing options={basicstyle=\ttfamily\small, columns=fullflexible}
}

\title{Optimal Turkish Subword Strategies at Scale: Systematic Evaluation of Data–Vocabulary–Morphology Interplay}

\author{Duygu Altinok}
\affiliation{Independent Researcher, Berlin, Germany}
\email[F. Author]{duygu@turkish-nlp-suite.com}


\begin{document}

\begin{abstract}
Tokenization is a pivotal design choice for neural language modeling in morphologically rich languages (MRLs) such as Turkish, where productive agglutination challenges both vocabulary efficiency and morphological fidelity. Prior studies have explored tokenizer families and vocabulary sizes, but they typically (i) vary vocabulary without systematically controlling the tokenizer's training corpus, (ii) provide limited intrinsic diagnostics of segmentation quality, and (iii) evaluate a narrow slice of downstream tasks. We present the first and only comprehensive, principled study of Turkish subword tokenization— a subwords manifest —that jointly varies vocabulary size and tokenizer training corpus size (data–vocabulary coupling), compares multiple tokenizer families under matched parameter budgets (WordPiece, morphology-level, and character baselines), and evaluates across semantic (NLI, STS, sentiment analysis, NER), syntactic (POS, dependency parsing), and morphology-sensitive probes. To explain why tokenizers succeed or fail, we introduce a morphology-aware diagnostic toolkit that goes beyond coarse aggregates to boundary-level micro/macro-F1 over gold morpheme boundaries, decoupled lemma atomicity vs\. surface boundary hits, over-/under-segmentation indices, character/word edit distances (CER/WER), continuation rates, and affix-type coverage and token-level atomicity. Our contributions are fourfold: (i) a systematic investigation of the vocabulary–corpus–success triad for Turkish, including larger data regimes than prior work; (ii) a unified, morphology-aware evaluation framework linking intrinsic diagnostics to extrinsic outcomes; (iii) extensive, controlled comparisons that identify when character-level and morphology-level tokenization pay off; and (iv) full open-source release of evaluation code, tokenizer training pipelines, and interim Transformer checkpoints for reproducibility. As the first and only work of its kind, this subwords manifest delivers actionable, prescriptive guidance for building effective tokenizers in MRLs and establishes a reproducible foundation for future research and deployment.
\end{abstract}

\section{Introduction}
Since the advent of Transformer architectures—from encoder-only models like BERT \autocite{devlin-etal-2019-bert} to decoder-only large language models\autocite{minaee2024largelanguagemodelssurvey}, tokenizers have drawn sustained attention, especially their underlying algorithms. For non-English and, in particular, morphologically rich languages (MRLs), tokenization becomes even more consequential: how vocabularies are compressed, how subwords align with linguistic units (morphemes), and how compression choices affect downstream task performance are central design questions.

Tokenization is the first—and often most consequential—interface between raw text and neural language models. Its design determines how linguistic structure is exposed to the model, how parameters are allocated, and how efficiently sequences are represented. For MRLs such as Turkish, where productive agglutination creates long and sparse surface forms \autocite{Oflazer1994}, tokenization is not merely a preprocessing choice but a core modeling decision. Word-level tokenization explodes the vocabulary and invites out-of-vocabulary failures; character- or byte-level tokenization lengthens sequences and obscures morpheme boundaries; and off-the-shelf subword tokenizers (e.g. BPE \autocite{sennrich2016neuralmachinetranslationrare}, WordPiece \autocite{devlin-etal-2019-bert}, Unigram \autocite{kudo-2018-subword}) often fragment stems or smear affixes, weakening the model's access to syntactic and morphological cues.

In Turkish specifically, the central practical question is how to represent and leverage morphology. Naturally Transformer type tokenizers such as  WordPiece and BPE attracted attention from research side. Recent work has examined aspects of this problem in Turkish: suffix-preserving tokenizers have shown modest, consistent gains on select tasks; RoBERTa-scale comparisons have tied vocabulary size to downstream performance; and WordPiece/BPE have remained strong baselines. Yet, these studies largely vary vocabulary size without systematically controlling or scaling the tokenizer's training corpus, provide limited intrinsic diagnostics of segmentation quality, and evaluate a narrower slice of tasks—often omitting morphology- and syntax-sensitive settings where boundary fidelity matters most. As a result, practitioners lack prescriptive guidance on when to favor larger vocabularies, morphology-aware tokenizers, or byte/character regimes, and how tokenizer training data interacts with these choices.

Motivated by these gaps, we argue that Turkish subword modeling and morphological structure have not been dissected in sufficient depth. This paper advances tokenization for Turkish from ad hoc exploration to principled design. We present the first comprehensive study that jointly varies vocabulary size and tokenizer training corpus size (data–vocabulary coupling), compares multiple tokenizer families under matched parameter budgets (WordPiece, morphology-level, and character baselines), and evaluates across a broad suite of downstream tasks spanning semantic (NLI, STS, NER, sentiment analysis), syntactic (POS, dependency parsing), and morphology-sensitive probes. To explain \emph{why} tokenizers succeed or fail, we introduce a morphology-aware diagnostic toolkit that moves beyond coarse aggregates (e.g., fertility) to boundary-level micro/macro-F1 over gold morpheme boundaries, decoupled lemma atomicity etc\. surface boundary hits, over-/under-segmentation indices, character/word edit distances (CER/WER), continuation rates, and affix-type coverage and token-level atomicity.

Our contributions are fourfold:
(i) A systematic investigation of the vocabulary–corpus–success triad for Turkish tokenization, including larger data regimes than prior work.
(ii) A unified, morphology-aware evaluation framework that links intrinsic segmentation diagnostics to extrinsic task outcomes, enabling causal interpretation rather than post hoc correlation.
(iii) Extensive, controlled comparisons across tokenizer families and parameter allocations, including conditions where character-level tokenization is competitive (e.g., NER) and when morphology-level tokenization pays off.
(iv) Open-source release of all evaluation code, tokenizer training pipelines, and interim Transformer checkpoints to ensure full reproducibility and to facilitate further research and deployment in Turkish NLP.

Positioned as a manifest in Turkish subwords, our study is the first and only of its kind to integrate large-scale tokenizer data sweeps, fine-grained morphological diagnostics, and the widest coverage of morphology- and syntax-sensitive tasks. We transform fragmented observations into actionable, prescriptive rules for building tokenizers that actually work for morphologically rich languages—grounded in evidence, reproducible by design, and immediately useful to both researchers and practitioners. We offer all our work freely, all evaluation and training scripts under our Github repository \footnote{\url{https://github.com/turkish-nlp-suite/Turkish-subwords-research}} and Transformer models under our Hugging Face repository\footnote{\url{https://huggingface.co/collections/turkish-nlp-suite/turkish-subwords-research}}. 

\section{Related Work}
In this section, we present works directly related to our study and compare them to our contributions.

\autocite{erencan-boun} investigates how subword tokenization interacts with Turkish morphology and whether morphology-aware tokenization improves modeling and downstream tasks. The study compares BPE, WordPiece, and Unigram across corpus and vocabulary sizes on Turkish data (OSCAR Turkish split \autocite{oscar2}), and introduces a suffix-preserving tokenizer (“morphosubwords”) that keeps affixes atomic while letting roots be learned. With ELECTRA \autocite{clark2020electra} pretraining, morphosubwords reduces pseudo-perplexity and delivers small but consistent gains on POS, NER, QA, and sentiment versus a WordPiece baseline, with faster convergence. Intrinsically, they report trends with data and vocabulary size (fertility increases, average token length decreases, single-word rate decreases, morphology-compatibility increases, and diminishing returns at larger scales). Their morphology metrics include fertility, average token length, single-word token rate, word-level morphology-compatible segmentation, suffix precision/recall, and a “root tokens” rate. 

Common concepts with our work: both target Turkish MRL tokenization; both analyze corpus- and vocabulary-size effects; both compare standard subword tokenizers to a morphology-aware variant; both evaluate on POS, NER, QA, and sentiment; and both use morphology-sensitive metrics alongside intrinsic/extrinsic evaluation. Where we differ: our evaluation is more granular and diagnostic (boundary-level micro/macro-F1 over all morpheme boundaries; decoupled lemma atomicity vs.\ surface boundary hits; over-/under-segmentation indices; CER/WER sequence edits; continuation rate; affix-type coverage and token-level atomicity). We empirically connect vocabulary size and tokenizer training corpus size to downstream success (not just intrinsic metrics), extend to larger-data regimes (e.g., $\sim$80\,GB), and include pre-transformer analyses and interpretability diagnostics, yielding a broader and more explanatory treatment.

\autocite{toraman} analyze tokenization granularity for Turkish by pretraining RoBERTa-medium \autocite{zhuang-etal-2021-robustly} on OSCAR-TR with five tokenizers (character, BPE, WordPiece, morphological, word) and evaluating six downstream tasks (news, hate speech, sentiment, NER, STS, NLI), while sweeping vocabulary size via embedding-parameter allocation. Results show WordPiece/BPE strongest overall, a morphology-level tokenizer competitive but slightly behind, word-level hurt by UNKs, and character-level underperforming at this scale. Increasing vocabulary size yields monotonic gains with quicker saturation for BPE/WordPiece and larger relative gains for morphology/word-level; they recommend allocating 20\% of parameters to embeddings for de facto tokenizers and 40\% for morphology/word-level.

Similar to our work, \autocite{toraman} (i) targets Turkish and agglutinative morphology; (ii) compares de facto subword tokenizers against a morphology-level alternative; and (iii) systematically studies vocabulary size and ties it to downstream performance across multiple tasks. Where we go further, our study (i) adds syntactic evaluations (POS, dependency parsing) that are particularly sensitive to morpheme boundaries; (ii) provides explainability via rich morphology-aware metrics (boundary micro/macro-F1 over gold morpheme boundaries, lemma boundary hits and lemma single-token rates, over-/under-segmentation indices, CER/WER, continuation rate, affix coverage/atomicity) rather than only extrinsic scores; (iii) examines the vocabulary–corpus–success triad by varying tokenizer training corpus size (including larger data regimes), which \autocite{toraman} does not; and (iv) offers a nuanced view on character/byte models (e.g., character-level can be competitive on NER under some settings), whereas \autocite{toraman} reports a broadly negative conclusion at their scale.

\autocite{kaya} study tokenization granularity for Turkish by pretraining Turkish BERT variants with WordPiece tokenizers at multiple vocabulary sizes (32k–256k) on the BERTurk corpus \autocite{stefan_schweter_2020_3770924} and evaluating on NER, QA, and sentiment. They also test normalization and simple morphology-injection schemes (tags and inflectional groups). Their results show that larger vocabularies steadily improve tokenization “success” for token-level tasks (NER, QA), with saturation around 128k–256k, while sentiment—driven by the [CLS] representation—does not benefit from larger vocabularies. Morphology-tag tokenization increases sequence length and generally harms downstream performance under a 512-token limit. Overall, they confirm a task-dependent trade-off between vocabulary size and granularity for Turkish.

This study is similar to our study in spirit, but our study is substantially broader and more diagnostic. First, we expand task coverage beyond NER/QA/SA to include NLI and syntax-sensitive evaluations (POS, dependency parsing), which are critical for assessing morpheme-boundary fidelity. Second, we sweep vocabulary sizes much more finely, including very small regimes (e.g., 1–8k) where segmentation behavior and sequence-length pressure are most revealing in agglutinative languages, and we compare multiple tokenizer families (WordPiece, BPE, morphology-level, and character/byte baselines) under matched parameter budgets. Third, unlike \autocite{kaya}—who train tokenizers on a fixed large corpus—we explicitly vary tokenizer training corpus size and domain to study data–vocabulary coupling. Finally, we add intrinsic, morphology-aware diagnostics (boundary F1, lemma boundary hits, affix coverage/atomicity, over-/under-segmentation indices, continuation rates) and analyze parameter-allocation trade-offs, yielding prescriptive guidance on when larger vocabularies or morphology-level tokenizers pay off and why.

In summary, earlier studies illuminate pieces of the puzzle—tokenizer choice, vocabulary size, or individual tasks—but stop short of a holistic, data-coupled, diagnostic account for Turkish. Our work is the first and only of its kind, a manifest in Turkish subwords: we systematically couple vocabulary with tokenizer training data across larger regimes, evaluate the broadest suite of semantic, syntactic, and morphology-sensitive tasks, and contribute the most comprehensive morphology-aware diagnostics to date. This manifest transforms scattered observations into actionable, prescriptive rules for building effective tokenizers in morphologically rich languages.

\section{Datasets}
We evaluate Turkish subword strategies across diverse corpora and tasks that map to our three axes: data scale (for tokenizer training), vocabulary size (via controlled tokenizer variants), and morphology-aware evaluation (via syntactic and morphological benchmarks). This section details the datasets used for benchmarking and, where relevant, how they interface with our diagnostics.

\subsection{Benchmarking}
We adopt a broad evaluation suite covering semantic, named entity recognition, and syntax/morphology-sensitive tasks:

\paragraph{Semantic (TrGLUE)}
We use the newly released TrGLUE \autocite{altinok2025introducingtrgluesentiturcacomprehensive, wang-etal-2018-glue} benchmark for Turkish, which aggregates multiple semantic tasks to probe representation quality across inference, similarity, and classification. TrGLUE provides standardized splits and metrics, enabling consistent model comparison under matched conditions.

We evaluate model success on a focused yet linguistically diverse suite chosen to probe morphology–subword interactions without excessive redundancy. For syntax, we include POS tagging and NER, which are directly sensitive to morpheme segmentation, affix boundaries, and lemma integrity. TrCoLA \autocite{cola, altinok2025introducingtrgluesentiturcacomprehensive}
serves as a complementary stress test for morphosyntactic well‑formedness, capturing acceptability phenomena beyond token‑level labels. For sentence‑level semantics, TrMNLI \autocite{mnli, altinok2025introducingtrgluesentiturcacomprehensive} provides broad inference coverage across genres, TrMRPC \autocite{mrpc, altinok2025introducingtrgluesentiturcacomprehensive} tests paraphrase recognition under varying lexical overlap, TrSST‑2 \autocite{sst2, altinok2025introducingtrgluesentiturcacomprehensive} captures polarity composition in single sentences, and STS‑B \autocite{stsb, altinok2025introducingtrgluesentiturcacomprehensive} measures graded semantic similarity with continuous correlations. This mix balances syntax‑heavy supervision with semantic understanding and uses metric diversity (accuracy/F1, MCC, correlation) to reveal how subword choices trade off sequence length, OOV handling, and morphological fidelity across task families. 

Table \ref{tab:bench-size} gives the sizes and evaluation metrics per evaluation set.

\begin{table}[ht!]
\centering
\small
\caption{Sizes of TrGLUE datasets, task type,  size, eval metric and BERTurk performance on them.}
\begin{tabular}{lrrrrr}
\toprule
Dataset & Task &  Size & Eval metric & BERTurk success \\
\midrule
TrCoLA & acceptability & 9.9K & Matthews' corr. &  42 \\
TrMNLI & NLI & 202K & matched/mismatched acc. & 87.9/90.8 \\
TrMRPC & paraphrase & 5.18K & acc./F1 &  74.4/72.7  \\
TrSST-2 & sentiment & 78K & acc./F1 & 87.4/91.3  \\
TrSTS-B &  sentence similarity & 3.06K & Pearson/Spearman corr. & 71.3/69.9  \\
\bottomrule
\end{tabular}
\label{tab:bench-size}
\end{table}

For brevity, we omit the “Tr-” prefix in the remainder of the paper (e.g., TrCoLA → CoLA, TrMRPC → MRPC) and refer to the Turkish variants by their task names.

\paragraph{Named Entity Recognition (NER)}
We evaluate on the Turkish WikiNER dataset introduced by \citet{altinok-2023-diverse}. The corpus uses 19 entity tags, covering traditional categories (e.g., PERSON, LOC) and finer-grained types (e.g., TITLE, EVENT, TIME). The dataset comprises approximately 20k sentences, with 18k for training and 1k each for development and test. As a reference point, a BERTurk \autocite{stefan_schweter_2020_3770924} baseline attains an F1-score of 0.77 on the official test split.

\paragraph{Syntax and Morphology (BOUN Treebank)}
For POS tagging, dependency parsing \autocite{de-marneffe-etal-2021-universal}, and morphology-sensitive evaluation, we use the BOUN Turkish Treebank \autocite{boun} in UD format. The training set contains roughly 7.8k sentences, with development and test sets of about 1k sentences each. The treebank provides rich Turkish morphological annotation, enabling fine-grained diagnostics beyond POS and dependencies.

As a baseline, we evaluate a BERTurk-based model on BOUN across tasks. The model achieves 92.63 UPOS accuracy, 81.51 UAS, and 74.59 LAS. For morphology, overall micro-accuracy is 30.76, with substantial variation across features: higher accuracies for Reflex (56.20), Typo (50.81), VerbForm (49.12), Abbr (46.57), and Number[psor] (43.48); and lower performance for Voice (2.01), Mood (6.54), Aspect (11.41), Case (11.96), Tense (21.49), and PronType (15.35). These results suggest that while the baseline captures several orthographic and inflectional cues, categories with finer-grained semantics and skewed label distributions remain challenging—underscoring the value of morphology-aware tokenization and diagnostics.

\subsection{Morphological segmentation}
We curated a Turkish morphology evaluation set compiled specifically for tokenizer assessment. Each instance provides a surface word form alongside its gold morphological analysis: a lemma (base form) and a “+”-separated suffix sequence reflecting the inflectional or derivational chain. The collection is organized into five validation-only subsets to probe different usage regimes: Çekimli (general suffixed words), Common Nouns (suffixed nouns), Common Verbs (suffixed verbs), Lemma (lemma-only forms for lemma integrity), and Common Lemmas (frequent lemmas). Together, these splits cover a broad range of productive suffixation patterns in Turkish, enabling evaluation of whether subword segmentations align with morpheme boundaries and preserve lemma integrity. We collected analyses strings from a previous work,  the collection of analyses called Turkish morph dictionaries \autocite{duygu-statistical-morph}.

\noindent The dataset is distributed as JSON Lines (one analysis per line) with fields \texttt{word} (surface), \texttt{lemma}, and \texttt{suffixes} (a “+”-joined string; empty for lemma-only items). It emphasizes clean morphological structure over sentence context, focusing evaluation on subword behavior at the word level. This makes it well-suited for comparing tokenizers (e.g., BPE, WordPiece, Unigram) by boundary precision/recall/F1, subwords-per-word, and lemma atomicity, and for diagnosing over- and under-segmentation on nouns and verbs. The resource is intended for test-only use; no training split is provided.

\begin{examplebox}
word: kitaplarımızda | lemma: kitap | suffixes: lar+ımız+da
word: evlerden | lemma: ev | suffixes: ler+den
word: koşuyordum | lemma: koş | suffixes: uyor+du+m
word: güzelleştirmek | lemma: güzel | suffixes: leş+tir+mek
word: bilgisayarcılardan | lemma: bilgisayar | suffixes: cı+lar+dan
word: gelmeyecekti | lemma: gel | suffixes: me+yecek+ti
word: çocuk | lemma: çocuk | suffixes:
word: çalıştırdık | lemma: çalış | suffixes: tır+dı+k
\end{examplebox}

This dataset can be found on its Hugging Face repository\footnote{\url{https://huggingface.co/datasets/turkish-nlp-suite/turkish-morph-analysis}}.

\section{Tokenization Metrics}
\subsection{Tokenization Granularity and Fragmentation Metrics}

In this subsection we formalize two complementary metrics—fertility and token continuation rate—that together characterize how a subword tokenizer fragments text. These measures are model-agnostic yet diagnostic for downstream behavior: they govern pretraining sequence lengths, the visibility of morpheme boundaries, and the trade-off between compression (short sequences) and compositionality (interpretable subword structure).

\subsubsection{Fertility}
Fertility is basically the average number of subwords per word.
Let the corpus contain words $\{w_i\}_{i=1}^{N}$, and let $t(w_i)$ be the multiset of subword tokens for $w_i$. Define
\[
\text{Fertility} \;=\; \frac{1}{N} \sum_{i=1}^{N} \bigl|t(w_i)\bigr|.
\]
Lower values indicate stronger compression (more words realized as single tokens); higher values indicate greater fragmentation. In morphologically rich languages, very high fertility often reflects near character-level behavior; very low fertility can reflect whole-word memorization that may obscure productive morphology.

\subsubsection{Token continuation rate (cross-token span frequency)}
Many subword schemes mark continuations (e.g., WordPiece \#\#). Let $\{u_j\}_{j=1}^{M}$ be the subword stream and $\mathbb{1}_{\text{cont}}(u_j)$ indicate whether $u_j$ is a continuation (i.e., not the first subword aligned to its source word). Define
\[
\text{Continuation rate} \;=\; \frac{1}{M} \sum_{j=1}^{M} \mathbb{1}_{\text{cont}}(u_j).
\]
High continuation rates imply long intra-word chains; low rates indicate that most tokens start new spans, consistent with whole-word coverage or morpheme-sized chunks. Fertility captures sequence-length inflation; continuation rate captures the shape of segmentations within words.

Here's an example:

\begin{tcolorbox}[colback=blue!2!white,colframe=blue!40!black,title={Morpheme-aligned mid granularity}]
Sentence: \emph{Evlerimizden ayrıldık ve Ankara'ya döndük.}\\
evlerimizden $\rightarrow$ [ev, \#\#ler, \#\#imiz, \#\#den] (4)\\
ayrıldık $\rightarrow$ [ayrıl, \#\#dık] (2) \quad ve $\rightarrow$ [ve] (1)\\
Ankara'ya $\rightarrow$ [Ankara, \#\#', \#\#ya] (3)\\
döndük $\rightarrow$ [dön, \#\#dük] (2) \quad . $\rightarrow$ [.] (1)\\
Fertility: $(4+2+1+3+2+1)/6 = 13/6 \approx 2.17$.\\
Continuation rate: continuations are \#\#ler, \#\#imiz, \#\#den, \#\#dık, \#\#dük (5) out of total 13 subwords (period excluded), so $5/13 \approx 0.38$.
\end{tcolorbox}

\subsubsection{Interpreting the metric pair}
- Moderate fertility with a moderate continuation rate often signals morpheme-aligned segmentations.

- Very high fertility with very high continuation rate indicates over-fragmentation (near character-level).

- Very low fertility with very low continuation rate suggests whole-word memorization and potential loss of compositionality.

\subsection{Morphology-Aware Tokenization Metrics}
\label{subsec:morpho-metrics}
Turkish is morphologically rich and highly agglutinative. A single surface word typically consists of a lemma followed by a chain of suffixes. Standard subword tokenizers (e.g., BPE, WordPiece) optimize for frequent orthographic substrings rather than linguistically meaningful morphemes. We propose a suite of complementary metrics to assess whether tokenizers respect morphological structure. The metrics jointly evaluate boundary alignment, lemma integrity, segmentation granularity, sequence-level agreement, and vocabulary-to-affix correspondence.

\subsubsection{Setup and Notation}
Let the dataset contain $N$ surface words. For item $i\in\{1,\dots,N\}$:
\begin{itemize}
  \item The surface string is $w^{(i)}$ with character length $|w^{(i)}|$.
  \item The gold analysis consists of a lemma $\ell^{(i)}$ and a gold morpheme sequence $m^{(i)}_{1:k_i} = [m^{(i)}_1,\dots,m^{(i)}_{k_i}]$, where $m^{(i)}_1=\ell^{(i)}$ and $m^{(i)}_2,\dots,m^{(i)}_{k_i}$ are suffixes.
  \item A tokenizer maps $w^{(i)}$ to subwords $t^{(i)}_{1:n_i} = [t^{(i)}_1,\dots,t^{(i)}_{n_i}]$.
\end{itemize}
We interpret subword ends as predicted morpheme boundaries and compare them against gold morpheme ends. Using 1-indexed character offsets measured from the start of $w^{(i)}$:
\begin{align*}
B_{\text{gold}}^{(i)} &= \left\{ \sum_{j=1}^{u} \left|m^{(i)}_j\right| \;:\; u=1,\dots,k_i \right\},\\
B_{\text{pred}}^{(i)} &= \left\{ \sum_{j=1}^{u} \left|t^{(i)}_j\right| \;:\; u=1,\dots,n_i \right\}.
\end{align*}
For WordPiece-like tokenizers, continuation markers (e.g., \#\# are removed before computing lengths.

\begin{tcolorbox}[colback=blue!2!white,colframe=blue!40!black,title={Illustrative gold analyses and tokenizations}]
We use three running examples:
\begin{itemize}
  \item kitaplarımızda $\;=\;$ \textbf{kitap} + lar + ımız + da
  \item güzelleştirmek $\;=\;$ \textbf{güzel} + leş + tir + mek
  \item koşuyordum $\;=\;$ \textbf{koş} + uyor + du + m
\end{itemize}
We will compare different predicted tokenizations (good, over-segmentation, under-segmentation, and lemma splits) against these gold segmentations.
\end{tcolorbox}

\subsubsection{Core Metrics}

\paragraph{1) Mean subwords per word}
Average tokenization granularity:
\begin{equation}\label{eq:mean_subwords}
\text{Subwords/Word} \;=\; \frac{1}{N}\sum_{i=1}^{N} n_i.
\end{equation}
High values imply fine-grained segmentations (potential over-segmentation and longer sequences); low values imply coarse segmentations (potential under-segmentation).

\begin{tcolorbox}[colback=gray!3,colframe=gray!60!black,title={Example: Mean subwords/word}]
For \emph{kitaplarımızda}:\\
Tokenizer A: [kitap][lar][ımız][da] $\Rightarrow n=4$.\\
Tokenizer B: [kita][p][lar][ımız][da] $\Rightarrow n=5$ (more granular).
\end{tcolorbox}

\paragraph{2) Boundary precision, recall, and micro-F1}
Define true/false positives and false negatives at the character-boundary level:
\begin{align}
\text{TP} &= \sum_{i=1}^{N} \left| B_{\text{pred}}^{(i)} \cap B_{\text{gold}}^{(i)} \right|, \\
\text{FP} &= \sum_{i=1}^{N} \left| B_{\text{pred}}^{(i)} \setminus B_{\text{gold}}^{(i)} \right|, \\
\text{FN} &= \sum_{i=1}^{N} \left| B_{\text{gold}}^{(i)} \setminus B_{\text{pred}}^{(i)} \right|.
\end{align}
Then micro-averaged precision, recall, and F1 are
\begin{align}\label{eq:micro_prf}
P_{\mu} = \frac{\text{TP}}{\text{TP}+\text{FP}},\qquad
R_{\mu} = \frac{\text{TP}}{\text{TP}+\text{FN}},\qquad
F1_{\mu} = \frac{2P_{\mu}R_{\mu}}{P_{\mu}+R_{\mu}}.
\end{align}

\begin{tcolorbox}[colback=gray!3,colframe=gray!60!black,title={Example: Boundary micro-F1}]
Gold boundaries in \emph{kitap+lar+ımız+da} occur at offsets $\{5,8,12,14\}$.\\
Pred 1: [kitap][lar][ımız][da] $\Rightarrow B_{\text{pred}}=\{5,8,12,14\}$, so $P=R=1$.\\
Pred 2: [ki][tap][lar][ımız][da] adds an extra boundary at offset 2 $\Rightarrow \text{FP}\uparrow$, $P<1$, while recall may remain high.
\end{tcolorbox}

\paragraph{3) Boundary macro-F1}
Per-item precision/recall/F1:
\begin{align}
P^{(i)} &= \frac{\left| B_{\text{pred}}^{(i)} \cap B_{\text{gold}}^{(i)} \right|}{\left|B_{\text{pred}}^{(i)}\right| + \varepsilon},\\
R^{(i)} &= \frac{\left| B_{\text{pred}}^{(i)} \cap B_{\text{gold}}^{(i)} \right|}{\left|B_{\text{gold}}^{(i)}\right| + \varepsilon},\\
F1^{(i)} &= \frac{2P^{(i)}R^{(i)}}{P^{(i)}+R^{(i)}+\varepsilon},
\end{align}
with $\varepsilon>0$ a tiny constant to avoid division by zero.\footnote{E.g., single-morpheme items where $|B_{\text{gold}}^{(i)}|=0$.} The macro-F1 is
\begin{equation}\label{eq:macro_f1}
F1_{\text{macro}} \;=\; \frac{1}{N}\sum_{i=1}^{N} F1^{(i)}.
\end{equation}

\paragraph{4) Lemma boundary hit rate}
Let $L^{(i)} = |\ell^{(i)}|$ denote the lemma span length in characters within $w^{(i)}$. The lemma boundary hit indicator is
\begin{equation}\label{eq:lemma_hit_indicator}
h^{(i)} \;=\; \mathbb{1}\!\left[\, L^{(i)} \in B_{\text{pred}}^{(i)} \,\right],
\end{equation}
and the corpus-level rate is
\begin{equation}\label{eq:lemma_hit_rate}
\text{LemmaHit} \;=\; \frac{1}{N}\sum_{i=1}^{N} h^{(i)}.
\end{equation}

\begin{tcolorbox}[colback=green!2!white,colframe=green!40!black,title={Example: Lemma boundary hit}]
Gold: \emph{koş+yor+du+m}.\\
Pred A: [koş][uyor][dum] has a boundary at $|{\text{koş}}|$ $\Rightarrow h=1$.\\
Pred B: [ko][şuyor][dum] splits inside the lemma $\Rightarrow h=0$.
\end{tcolorbox}

\paragraph{5) Lemma single-token rate}
Tokenize each lemma in isolation; let $\tau(\cdot)$ be the tokenizer applied to a standalone string and $|\tau(\ell)|$ its subword count. Define
\begin{equation}\label{eq:lemma_single_token}
\text{Lemma1Tok} \;=\; \frac{1}{N}\sum_{i=1}^{N} \mathbb{1}\!\left[\, |\tau(\ell^{(i)})| = 1 \,\right].
\end{equation}
This reflects lemma atomicity in the vocabulary independent of surface affixation.

\paragraph{6) Over-/Under-segmentation indices}
Let $k_i$ be the number of gold morphemes and $n_i$ the number of predicted subwords for item $i$. We define:
\begin{align}\label{eq:over_under_seg}
\text{OverSeg} &= \frac{1}{N}\sum_{i=1}^{N} \frac{n_i}{k_i + \varepsilon},\\
\text{UnderSeg} &= \frac{1}{N}\sum_{i=1}^{N} \frac{k_i}{n_i + \varepsilon}.
\end{align}
Values $>1$ indicate, respectively, a tendency to split morphemes further (over-segmentation) or to fuse multiple morphemes (under-segmentation).

\begin{tcolorbox}[colback=orange!2!white,colframe=orange!60!black,title={Examples: Over/Under-segmentation}]
\emph{güzelleştirmek} (güzel+leş+tir+mek)\\
Good: [güzel][leş][tir][mek] $\Rightarrow n=4$, $k=4$, balanced.\\
Over-seg: [gü][zel][leş][tir][mek] $\Rightarrow n=5$, $k=4$, OverSeg $>1$.\\[4pt]
\emph{kitaplarımızda} (kitap+lar+ımız+da)\\
Under-seg: [kitaplarımızda] $\Rightarrow n=1$, $k=4$, UnderSeg $>1$.
\end{tcolorbox}

\paragraph{7) Sequence agreement: CER and WER}
Represent the gold morph sequence as a token string $S^{(i)}_{\text{gold}} = m^{(i)}_1{+}\cdots{+}m^{(i)}_{k_i}$ and the predicted subword sequence as $S^{(i)}_{\text{pred}} = t^{(i)}_1{+}\cdots{+}t^{(i)}_{n_i}$, where ``+'' is a literal separator. Let $\text{Edit}(\cdot,\cdot)$ be the Levenshtein distance.
\begin{align}
\text{CER} &= \frac{\sum_{i=1}^{N} \text{Edit}\big(\,\text{chars}(S^{(i)}_{\text{gold}}),\,\text{chars}(S^{(i)}_{\text{pred}})\,\big)}{\sum_{i=1}^{N} \left| \text{chars}(S^{(i)}_{\text{gold}}) \right|}, \label{eq:cer}\\
\text{WER} &= \frac{\sum_{i=1}^{N} \text{Edit}\big(\,\text{tokens}(S^{(i)}_{\text{gold}}),\,\text{tokens}(S^{(i)}_{\text{pred}})\,\big)}{\sum_{i=1}^{N} \left| \text{tokens}(S^{(i)}_{\text{gold}}) \right|}, \label{eq:wer}
\end{align}
where $\text{chars}(\cdot)$ returns a character sequence and $\text{tokens}(\cdot)$ returns the ``+''-separated units.

\begin{tcolorbox}[colback=purple!3!white,colframe=purple!60!black,title={Examples: CER/WER}]
Gold (\emph{koşuyordum}): ``koş+uyor+du+m''.\\
Pred: ``koş+uyor+dum'' $\Rightarrow$ small character edit (CER low), one token substitution (WER higher).
\end{tcolorbox}

\subsubsection{Supporting Metrics}

\paragraph{Affix coverage and atomicity}
Let $\mathcal{A}$ be a set of frequent suffix types (e.g., plural, case, possessive). For type-level coverage:
\begin{equation}\label{eq:affix_coverage}
\text{AffixCov} \;=\; \frac{1}{|\mathcal{A}|} \sum_{a \in \mathcal{A}} \mathbb{1}\!\left[\, \exists\, \text{subword type } u \text{ s.t. } u=a \,\right].
\end{equation}
For token-level atomicity, let $C(a)$ be the number of occurrences of affix $a$ in the corpus and $A(a)$ the number of those realized as a standalone predicted subword (a predicted boundary on both sides of $a$ in the surface word):
\begin{equation}\label{eq:affix_atomicity}
\text{AffixAtom} \;=\; \frac{\sum_{a \in \mathcal{A}} A(a)}{\sum_{a \in \mathcal{A}} C(a)}.
\end{equation}

\subsubsection{Worked examples}

\begin{tcolorbox}[colback=blue!1!white,colframe=blue!50!black,title={Example 1: kitaplarımızda}]
Gold: \textbf{kitap}+lar+ımız+da. Offsets after gold morphemes: $\{5,8,12,14\}$.

\textbf{Pred A (good)}: [kitap][lar][ımız][da]\\
$B_{\text{pred}}=\{5,8,12,14\}$. $\Rightarrow \text{TP}=4,\ \text{FP}=0,\ \text{FN}=0$; $P_{\mu}=R_{\mu}=F1_{\mu}=1$.\\
LemmaHit: boundary at $|\text{kitap}|=5$ exists $\Rightarrow 1$.\\
OverSeg $=n/k=4/4=1$; UnderSeg $=4/4=1$; WER/CER $=0$.

\textbf{Pred B (over-seg)}: [ki][tap][lar][ımız][da]\\
$B_{\text{pred}}=\{2,5,8,12,14\}$. Intersections $\{5,8,12,14\}$\\
$\Rightarrow \text{TP}=4,\ \text{FP}=1,\ \text{FN}=0$; $P_{\mu}=4/5,\ R_{\mu}=1,\ F1_{\mu}=\frac{2\cdot (4/5)\cdot 1}{(4/5)+1}=\frac{8}{9}\approx 0.889$.\\
LemmaHit: boundary at 5 exists $\Rightarrow 1$. OverSeg $=5/4=1.25$; UnderSeg $=4/5=0.8$.
\end{tcolorbox}

\begin{tcolorbox}[colback=blue!1!white,colframe=blue!50!black,title={Example 2: güzelleştirmek}]
Gold: \textbf{güzel}+leş+tir+mek. Suppose offsets are $\{5,8,11,14\}$ (illustrative).

\textbf{Pred A (good)}: [güzel][leş][tir][mek] $\Rightarrow$ perfect boundary match.

\textbf{Pred B (lemma split + over-seg)}: [gü][zel][leş][tir][mek]\\
$B_{\text{pred}}=\{2,5,8,11,14\}$; gold $\{5,8,11,14\}$.\\
$\Rightarrow \text{TP}=4,\ \text{FP}=1,\ \text{FN}=0$; $F1_{\mu}\approx 0.889$.\\
LemmaHit: boundary at $|\text{güzel}|=5$ exists $\Rightarrow 1$ (even though lemma was internally split).
\end{tcolorbox}

\begin{tcolorbox}[colback=blue!1!white,colframe=blue!50!black,title={Example 3: koşuyordum}]
Gold: \textbf{koş}+uyor+du+m. Let offsets be $\{3,7,9,10\}$.

\textbf{Pred A (good-ish)}: [koş][uyor][dum]\\
Assume subword char lengths $[3,4,3]$ on the surface, so $B_{\text{pred}}=\{3,7,10\}$; 

$\Rightarrow \text{TP}=3,\ \text{FP}=0,\ \text{FN}=1$; $P_{\mu}=1.00,\ R_{\mu}=0.75,\ F1_{\mu}\approx0.857$.\\
LemmaHit: boundary at $|\text{koş}|=3$ exists $\Rightarrow 1$.\\
Sequence strings: Gold ``koş+uyor+du+m'' vs Pred ``koş+uyor+dum'': CER$=0$, WER$=0.25$.
\end{tcolorbox}

High $F1_{\mu}$ and LemmaHit indicate that subword splits align with morphological seams; Lemma1Tok reflects vocabulary-level lemma atomicity. OverSeg/UnderSeg explain whether gains come from finer or coarser granularity. CER/WER summarize end-to-end sequence agreement, with CER sensitive to near-misses and WER stricter at the unit level. Affix coverage and atomicity connect the subword inventory to the language's productive suffix system.

\section{Pre-Transformer Tokenization Benchmarks}
We benchmark character-level tokenization, word-Level tokenization, and morphology-aware subwords on three task families: TrGLUE, NER, and POS–DEP–Morph. Each tokenizer has its own subsection with results and explainability; architectures are kept comparable across tokenizers for fair evaluation.

\paragraph{Tokenization schemes}
\begin{itemize}
  \item \textbf{Character-level tokenization}: words are segmented into characters; the first character is a standalone token and subsequent characters are prefixed with ``\#\#'' to denote continuation.
  \item \textbf{Word-level tokenization}: words remain intact as tokens.
  \item \textbf{Morphology-aware subwords}: words are segmented using Zeyrek \autocite{AkinAkin2007Zemberek} and spaCy Turkish \autocite{altinok-2023-diverse}; each subword is prefixed with ``\#\#'' to mark continuation.
\end{itemize}

\begin{examplebox}{Tokenization of the example word ``gittim'' under different schemes.}
\begin{tabular}{@{}ll@{}}
Character-level: & g \ \#\#i \ \#\#t \ \#\#t \ \#\#i \ \#\#m \\
Word-level: & gittim \\
Morphological subwords: & git \ \#\#ti \ \#\#m \\
\end{tabular}
\end{examplebox}

\paragraph{Modeling overview}
\begin{itemize}
  \item \textbf{TrGLUE:}
    \begin{itemize}
      \item Character-level: CNN encoder over character embeddings (no external pretrained embeddings).
      \item Word-level and morphology-aware subwords: BiLSTM encoders over embeddings initialized with word2vec (Google); embeddings are fine-tuned.
    \end{itemize}
  \item \textbf{NER:} BiLSTM encoder with a token-classification head (BIO). For word-level predictions, we compute one representation per word (direct word token or pooled subword/char states). In subword setups, only the first subword of each word can take B-/I- tags (others are masked or forced to O) to ensure consistent word-level span reconstruction. In character-level setups, BIO is predicted per character and contiguous labeled spans are merged back to words for F1.
  \item \textbf{POS–DEP–Morph:} A joint multi-task model over a shared BiLSTM. UPOS is predicted with a linear head; dependency parsing uses a deep-biaffine scorer for arcs and relations; morphology uses parallel per-attribute classifiers (including an explicit \textsc{None}). With subword/character inputs, word-level representations are formed via pooling (e.g., first-subword or mean) or by composing character states; all predictions and evaluation remain at the word level.
\end{itemize}

\noindent In the following subsections, we report results and provide tokenizer-specific explainability analyses for character-level tokenization, word-Level tokenization, and morphology-aware subwords, followed by key takeaways.

\subsection{Character-Level Tokenization}
\label{sec:char-baselines}
We evaluate character-level tokenization on TrGLUE, NER, and POS–DEP–Morph. Models operate directly over raw characters (first character standalone; subsequent characters marked with ``\#\#'').

\subsubsection{TrGLUE}
The character model achieves competitive performance on single-sentence sentiment (SST-2: $84.3\%$ accuracy) and a respectable baseline on natural language inference (MNLI: $67.1\%$ accuracy). In contrast, it underperforms on grammatical acceptability (CoLA: $\mathrm{MCC}=0.08$) and semantic textual similarity (STS-B: $\rho=0.12$ Pearson), with moderate results on paraphrase (MRPC: $62.1\%$ accuracy).

We observe three consistent phenomena:
\begin{itemize}
\item Tasks dominated by surface sentiment cues (e.g., polarity markers, intensifiers) are well captured by local character $n$-grams, yielding strong SST-2 performance without explicit lexical segmentation.
\item For inference (MNLI), the model recovers a non-trivial decision boundary from local sublexical patterns, suggesting that Turkish morphology conveys discriminative information even at the character level; however, the absence of longer-range compositional structure limits ceiling performance.
\item Tasks requiring fine-grained acceptability judgments or graded semantic similarity (CoLA, STS-B) degrade substantially. These tasks depend on syntactic well-formedness and precise lexical semantics, which are not reliably recoverable from short receptive fields and tokenization-free representations.
\end{itemize}

\subsubsection{NER}
Character-level NER achieved 0.70 F1 with our architecture, indicating that fine-grained subword representations and orthographic cues are sufficiently expressive to recover entity spans without reliance on pre-tokenized inputs. Despite the lack of explicit word boundaries and the longer effective sequence lengths characteristic of character-level processing, the encoder learns consistent patterns in morphology, affixation, and character-shape features that correlate with entity categories, yielding a balanced precision–recall profile. Typical error modes include boundary drift at span edges and confusions among semantically proximate labels (e.g., organization vs. location for institution names), reflecting the diffuse, local nature of character-level evidence. Nevertheless, this result constitutes a strong baseline for settings with noisy text, rich morphology, or ambiguous tokenization. 

Relative to English counterparts, character-level baselines on standard English NER benchmarks typically exceed this score when coupled with strong contextual encoders (e.g., BERT/SpanBERT) and token-level conditioning, often reporting F1 in the 0.90–0.95 range under well-tokenized, high-resource conditions. Purely character-driven models without large pretrained context generally trail those token-based systems in English by a substantial margin, as English benefits from comparatively stable tokenization, abundant labeled data, and capitalization cues. In contrast, for languages with richer inflection, noisier orthography, or less reliable tokenization, the gap narrows: character-level modeling better captures morphological regularities and exhibits robustness to OOV forms and spelling variation. Thus, while 0.70 F1 would be considered below state-of-the-art for English token-based pipelines, it is competitive for purely character-level architectures and is particularly promising in non-English or low-resource scenarios where subword granularity confers a larger advantage.

\subsubsection{POS-DEP-Morph}
Character-level baseline results on the BOUN treebank show strong performance on POS, morphology, and moderate performance on dependency parsing. The model attains 91.56 POS accuracy, 65.19 UAS, and 57.15 LAS. For morphological tagging, the overall micro-accuracy is 96.19, with uniformly high accuracies across individual features: Abbr (99.98), Echo (100.00), Typo (99.99), Reflex (99.72), NumType (99.48), Polarity (95.38), PronType (98.21), Evident (98.59), Mood (97.26), Tense (96.66), VerbForm (97.10), Voice (97.11), Person (91.04), Case (88.48), Number (89.25), and Number[psor]/Person[psor] around 94–95. These results indicate that, at the character level, the encoder effectively leverages orthographic regularities and inflectional morphology to recover rich feature structures, yielding near-ceiling performance on several categories that are often brittle in token-based setups. The comparatively lower UAS/LAS suggests that while local morphological cues are captured reliably, learning long-distance syntactic relations remains more challenging without strong contextual token-level priors.

Compared to BERTurk on the same dataset, the character-level baseline trades off syntactic attachment quality for morphological fidelity. Whereas the BERTurk baseline achieves higher UAS/LAS (81.51/74.59) and slightly better POS accuracy (92.63 vs. 91.56), its morphological micro-accuracy is substantially lower (30.76 vs. 96.19), with large gaps across many features (e.g., Voice: 2.01 vs. 97.11; Mood: 6.54 vs. 97.26; Case: 11.96 vs. 88.48). This divergence underscores different inductive biases: character-level modeling excels at capturing affixal and orthographic signals that directly encode morphological categories, while contextualized token encoders better model head–dependent structure and span-level semantics that support dependency parsing. In settings where accurate morphological annotation is paramount (e.g., downstream agreement, lemmatization, or generation), the character baseline establishes a strong reference point; conversely, for syntactic structure, pretrained token-based models retain an advantage.

\subsubsection{Key Findings}
\label{sec:key-findings-char}
Character-level models serve as tokenizer-free baselines that operate directly over raw characters, offering robustness to out-of-vocabulary (OOV) morphology, misspellings, and tokenization errors. Across TrGLUE, NER, and POS–DEP–Morph evaluations, they reveal where sublexical cues suffice and where explicit segmentation, longer-range composition, and lexicalization remain necessary.

Here are our key findings from this subsection:
\begin{itemize}
    \item \textbf{Strong tokenizer-free baselines on surface-driven tasks.}
    Character CNNs are competitive on single-sentence sentiment (e.g., SST-2) and deliver serviceable accuracy on MNLI, indicating that local character $n$-grams and morphotactics carry substantial signal without explicit tokenization.

    \item \textbf{Clear limits on structure- and meaning-sensitive tasks.}
    Large gaps on grammatical acceptability (CoLA) and graded semantic similarity (STS-B) highlight the need for longer-range composition and precise lexical semantics that character-only encoders struggle to capture.

    \item \textbf{Near-ceiling morphological fidelity; weaker long-distance syntax.}
    On BOUN, character-level baselines achieve very high morphological micro-accuracy with uniformly strong per-attribute scores, yet substantially lower UAS/LAS, indicating difficulty modeling head–dependent structure and long-distance attachments without strong token-level context.

    \item \textbf{Viable NER without tokenization.}
    Character-level NER attains an F1 of 0.70, showing that entity spans can be recovered from subword signals; primary errors involve boundary drift and confusions among closely related types.

    \item \textbf{Complementary to pretrained token/subword models.}
    Relative to BERTurk on BOUN, character models excel in morphological fidelity but trail in POS and dependency parsing, reflecting complementary inductive biases: local morpho-orthographic cues vs.\ broader contextual and lexical knowledge.

    \item \textbf{Practical advantages for agglutinative and low-resource settings.}
    Character models are robust to OOV forms and noisy orthography, avoid vocabulary explosion, and simplify preprocessing—benefits that are particularly salient for languages with rich morphology.

    \item \textbf{Methodological value as a clean floor.}
    Using characters establishes a clear baseline for attributing subsequent gains to subword segmentation and lexicalization, disentangling tokenizer coverage from genuine modeling improvements.
\end{itemize}

In summary, character-level modeling provides a robust, tokenizer-free baseline that excels at morphology and surface-driven classification, while exposing headroom on syntax and semantics. These complementary strengths motivate hybrid designs and systematic exploration of subword vocabularies to close the gaps on structure- and meaning-sensitive tasks.

\subsection{Word-Level Tokenization}
\label{sec:word-baselines}
We next consider the opposite extreme to subwording: tokens are whole words. In Turkish, rich inflection means many surface forms per lemma, so exact word forms in the test set are often unseen even when related inflections occur in training. This drives a substantial train–test mismatch via OOV items and motivates a careful measurement of vocabulary coverage and its impact on task performance.

For each task below, let $V$ denote the full word vocabulary extracted from training and sorted by frequency, and let $K$ be the size of the retained prefix (top-$K$ types). We report:
(i) training coverage: the fraction of training tokens accounted for by the top-$K$ types;
(ii) test coverage: the fraction of test tokens covered by the same top-$K$ types learned from training;
(iii) the relationship between coverage (or $K$) and final task scores.

This protocol quantifies how rapidly coverage saturates on train, how it decays on test, and how OOV exposure correlates with accuracy/F1 across TrGLUE, NER, and POS–DEP–Morph. It also provides a controlled knob ($K$) to compare word-level tokenization against character-level and morphology-aware subword baselines in subsequent analyses.

\subsubsection{TrGLUE}
We evaluate word-level tokenization task-by-task on TrGLUE, relating performance to training/test coverage, OOV rates, and the chosen top-$K$ vocabulary size.

\begin{figure}[ht!]
\centering
\begin{subfigure}{0.48\linewidth}
\centering
\includegraphics[width=\linewidth]{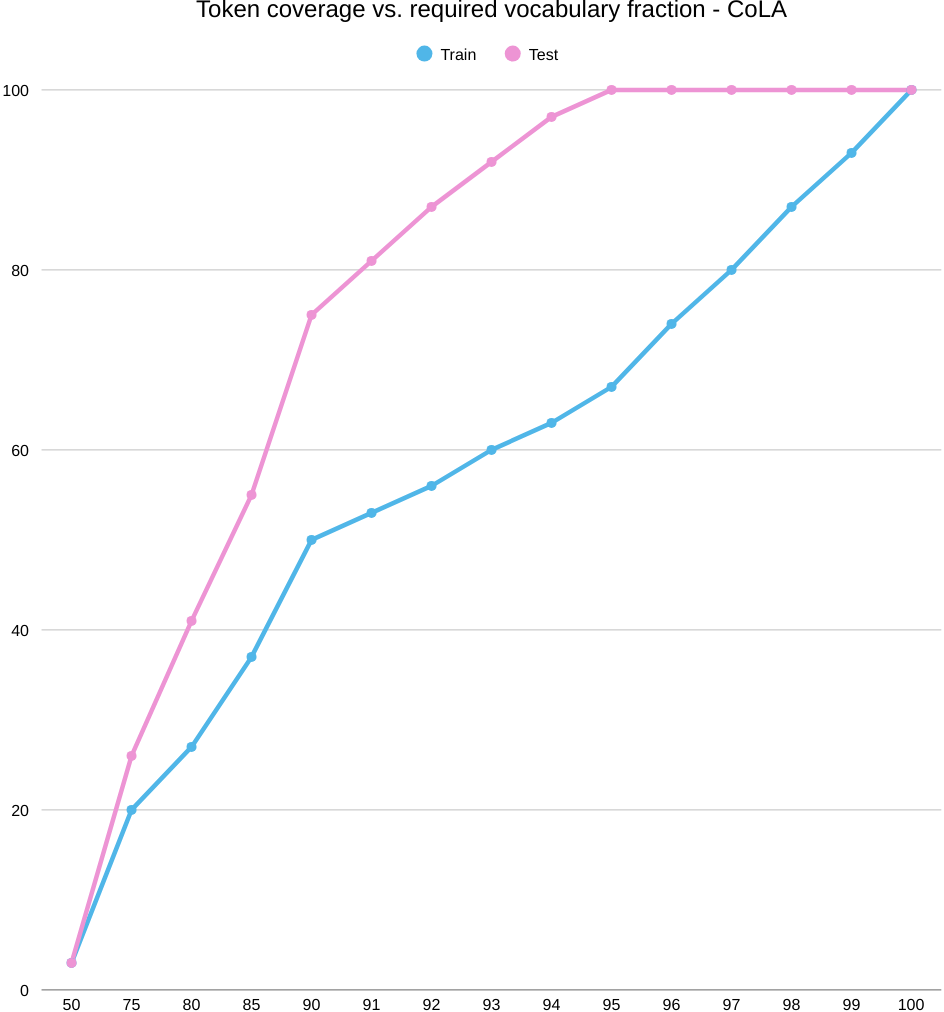}
\caption{Token coverage vs. required vocabulary fraction (K/|V|) for CoLA (Train/Test). Test coverage accumulates faster than train up to 95\% and then saturates, while train coverage increases gradually, indicating that many frequent train types do not transfer cleanly to the test distribution.}
\end{subfigure}
\hfill
\begin{subfigure}{0.48\linewidth}
\centering
\includegraphics[width=\linewidth]{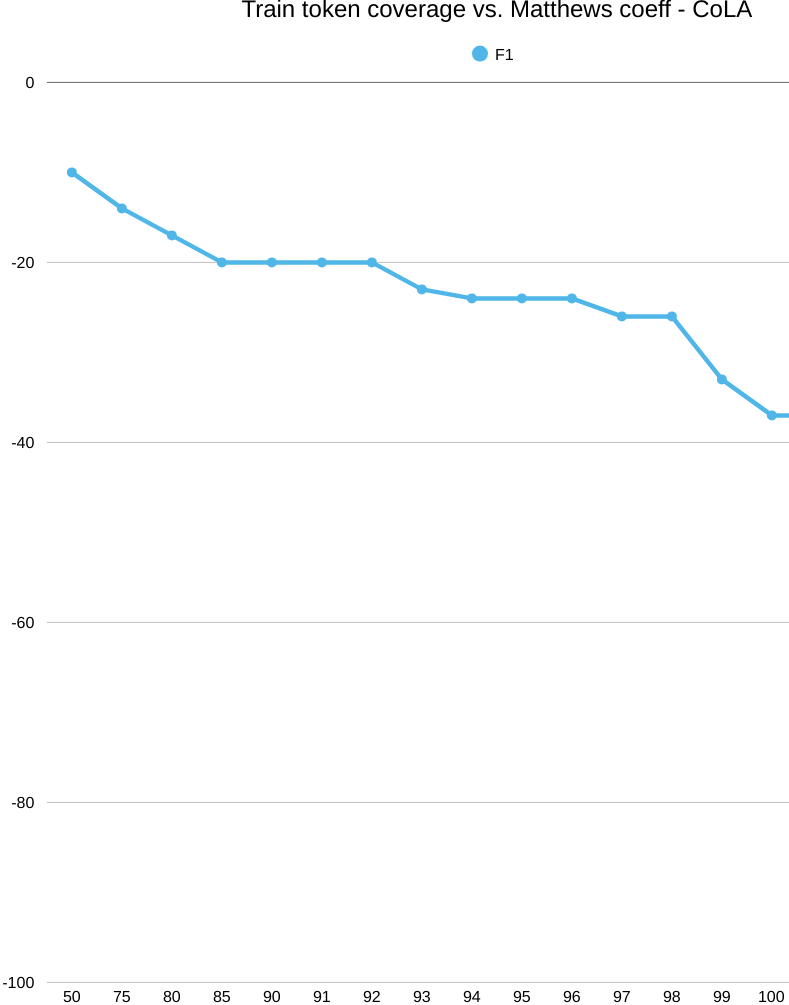}
\caption{Train token coverage vs. CoLA score (MCC). Each point uses the smallest top-K achieving the indicated train coverage. Performance is negative across the range and degrades slightly as coverage grows ($\approx$ -10 at 50\% to $\approx$ -36 at 100\%), suggesting the word-level model fails to capture grammatical acceptability even with full coverage.}
\end{subfigure}
\caption{CoLA with word‑level vocabularies: coverage efficiency and success. Left: achieving high token coverage requires retaining large fractions of the word list on train, with a different accumulation pattern on test. Right: increasing train token coverage does not improve CoLA performance and in fact trends downward, pointing to representation limits rather than coverage as the bottleneck.}
\label{fig:cola-word-success}
\end{figure}

\paragraph{CoLA} We examine CoLA under word-level vocabularies, relating token coverage to Matthews correlation (MCC).
As shown in Figure \ref{fig:cola-word-success} despite increasing coverage, the model's acceptability judgments remain below chance, indicating that lexical coverage alone is not sufficient for this task.

On CoLA, the word-level model never gets off the ground: MCC is negative across all coverages and drifts downward as we retain more of the vocabulary. The coverage–efficiency plot shows that train coverage grows slowly with K/|V| while test coverage saturates earlier, but the success curve makes clear that coverage isn't the bottleneck—representation is. Grammatical acceptability depends on abstract well‑formedness cues (long‑distance constraints, subcategorization, function words in context) that aren't captured by a sparse, surface‑form word inventory. The heavy morphological/orthographic tail further fragments evidence across many low‑count types, so adding more words mostly adds rare variants without improving generalization. In short, the model is effectively memorizing lexical patterns and label priors; as the vocabulary grows, this memorization amplifies noise and overfits train idiosyncrasies, yielding worse MCC. This aligns with the strong transformer gains on CoLA: contextualized subword representations encode the syntactic regularities that a word-level model misses.

\begin{figure}[ht!]
\centering
\begin{subfigure}{0.48\linewidth}
\centering
\includegraphics[width=\linewidth]{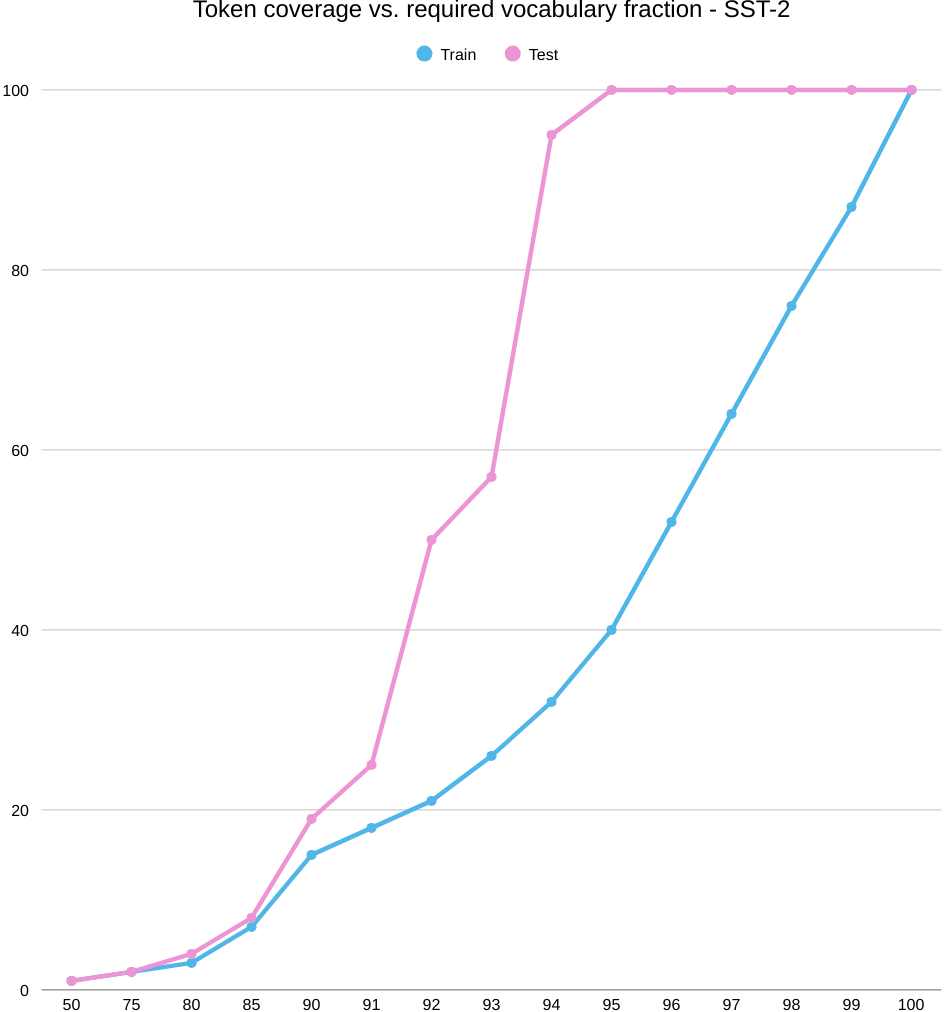}
\caption{Token coverage vs. required vocabulary fraction (K/|V|) for SST‑2 (Train/Test). Test coverage rises sharply and saturates around 93–95\% with a relatively small K, while train coverage increases gradually, indicating a mismatch in how the ranked word list accumulates coverage across splits.}
\end{subfigure}
\hfill
\begin{subfigure}{0.48\linewidth}
\centering
\includegraphics[width=\linewidth]{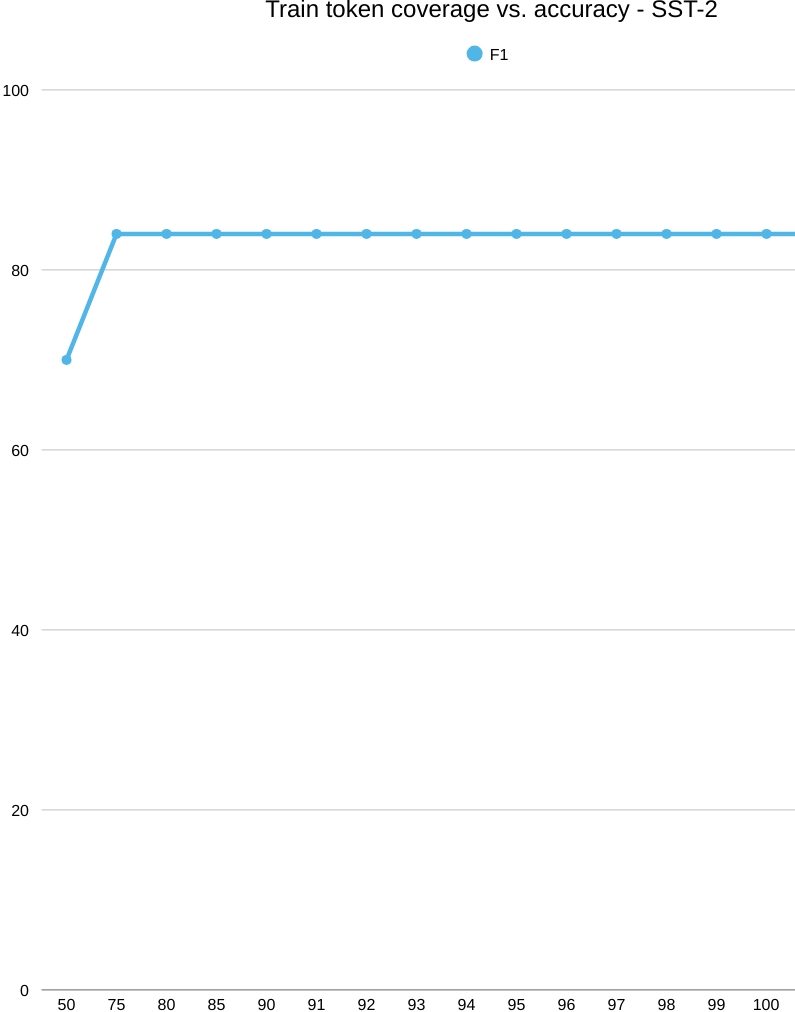}
\caption{Train token coverage vs. SST‑2 accuracy. Each point uses the smallest top‑K achieving the shown train coverage. Accuracy jumps from 69\% at 50\% coverage to 85\% by 75–80\%, then remains flat near 85\% through 100\% coverage.}
\end{subfigure}
\caption{SST‑2 with word‑level vocabularies: coverage efficiency and success. Left: coverage accumulates differently on train vs. test, with test saturating earlier. Right: performance exhibits an early “elbow,” reaching 85\% accuracy by 80\% train coverage and showing no gains with larger vocabularies.}
\label{fig:sst2-word-success}
\end{figure}

\paragraph{SST-2} Here we offer results for movie sentiment analysis binary task SST-2. Figure \ref{fig:sst2-word-success} shows the success vs vocabulary size.

For SST‑2, word‑level modeling shows a quick payoff and then a long plateau: accuracy climbs from 69\% at 50\% train coverage to 85\% by roughly 75–80\%, after which larger vocabularies do not help. The early gains occur once the top‑K list includes frequent polarity markers and phrases—e.g., “hiç,” “çok kötü,” “berbat,” “beğenmedim” on the negative side, and “mükemmel,” “harika,” “bayıldım” on the positive side. The coverage plot shows test coverage saturates earlier than train, but beyond this “elbow,” adding rarer types contributes little, consistent with sentiment being driven by a compact lexicon of strong cues.

For MNLI, MRPC, and STS‑B, vocabulary coverage behaves like CoLA (test coverage saturates with smaller K than train), while performance vs. coverage mirrors SST‑2 (early elbow followed by a broad plateau). Across coverage levels, scores fluctuate around roughly 0.70 accuracy for MNLI, 0.60 accuracy for MRPC, and 0.25 Pearson/Spearman for STS‑B, with no consistent gains near full coverage. Compared to the character baseline (MNLI 67.1, MRPC 62.1, STS‑B 0.12), word‑level modeling is slightly better on MNLI, roughly comparable on MRPC, and clearly stronger on STS‑B—but all three tasks exhibit the same diminishing‑returns pattern once the top‑K includes the frequent pairs and cues that drive each task.

\subsubsection{NER}
\begin{figure}[ht!]
\centering
\begin{subfigure}{0.48\linewidth}
\centering
\includegraphics[width=\linewidth]{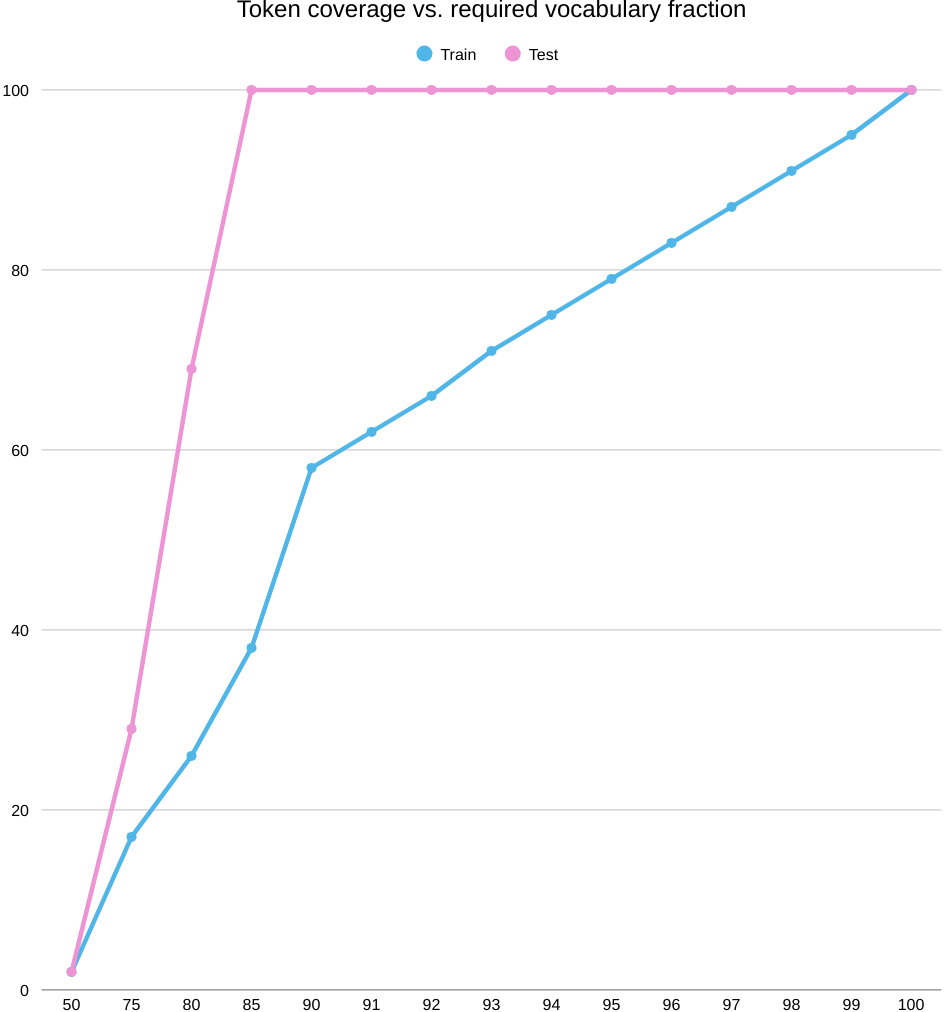}
\caption{Token coverage vs. required vocabulary fraction (K/|V|) for train and test. Curves show how much of the ranked word vocabulary must be retained to reach a given coverage on each split; test coverage lags train, indicating distribution shift.}
\end{subfigure}
\hfill
\begin{subfigure}{0.48\linewidth}
\centering
\includegraphics[width=\linewidth]{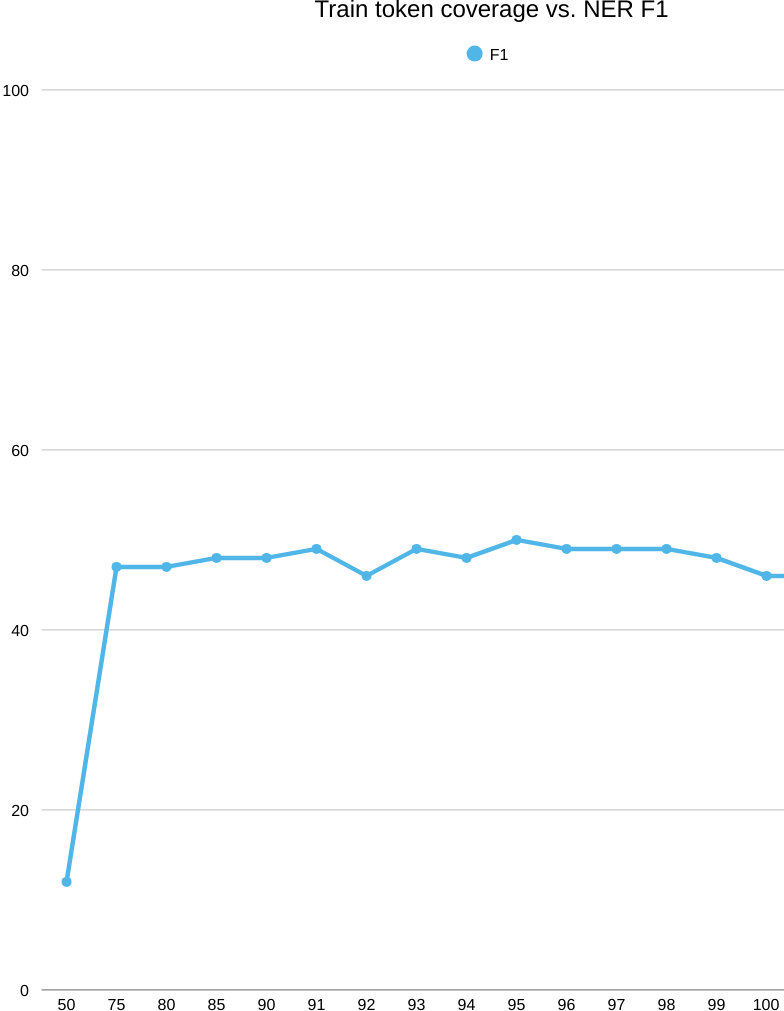}
\caption{Train token coverage vs. NER F1. Each point uses the smallest top‑K word list that achieves the indicated training token coverage; despite increasing coverage, F1 remains around 0.5.}
\end{subfigure}
\caption{Word‑level vocabulary efficiency and downstream NER performance. Left: achieving high coverage requires large fractions of the vocabulary, and test coverage accumulates more slowly than train, underscoring inefficiency and domain mismatch. Right: raising training token coverage from 75\% to 100\% yields only modest and unstable gains, with F1 saturating around 0.5.}
\label{fig:word-ner-curve}
\end{figure}

We analyze how efficiently a word-level vocabulary captures token mass in our data and whether increasing coverage translates into better NER performance. Figure \ref{fig:word-ner-curve} relates coverage to both the required vocabulary fraction and downstream F1 to diagnose efficiency and generalization.

Across the training data, token coverage increases only gradually as we retain larger fractions of the word vocabulary, with no early “head wins.” This unusually slow accumulation is consistent with a heavy‑tailed type distribution driven by morphology: many surface forms (inflections, casing, clitics, hyphenation) split a single lemma into numerous distinct word tokens, each occurring sparsely and counted as separate vocabulary items. As a result, even large expansions of K primarily mop up rare forms rather than consolidating mass under frequent entries, making the word vocabulary inefficient on both train and test (where coverage lags further due to domain shift and unseen variants). This inefficiency carries through to downstream performance: increasing training token coverage from 75\% toward 100\% yields only modest, unstable gains, with NER F1 saturating around 0.5—well below character/subword models ($\approx$ 0.70) and Transformer baselines ($\approx$ 0.77). In short, rich morphology fragments the word‑level representation, so chasing higher word coverage adds vocabulary without commensurate improvements in success.

\subsubsection{POS-DEP-Morph}
\begin{figure}[ht!]
\centering
\begin{subfigure}{0.48\linewidth}
\centering
\includegraphics[width=\linewidth]{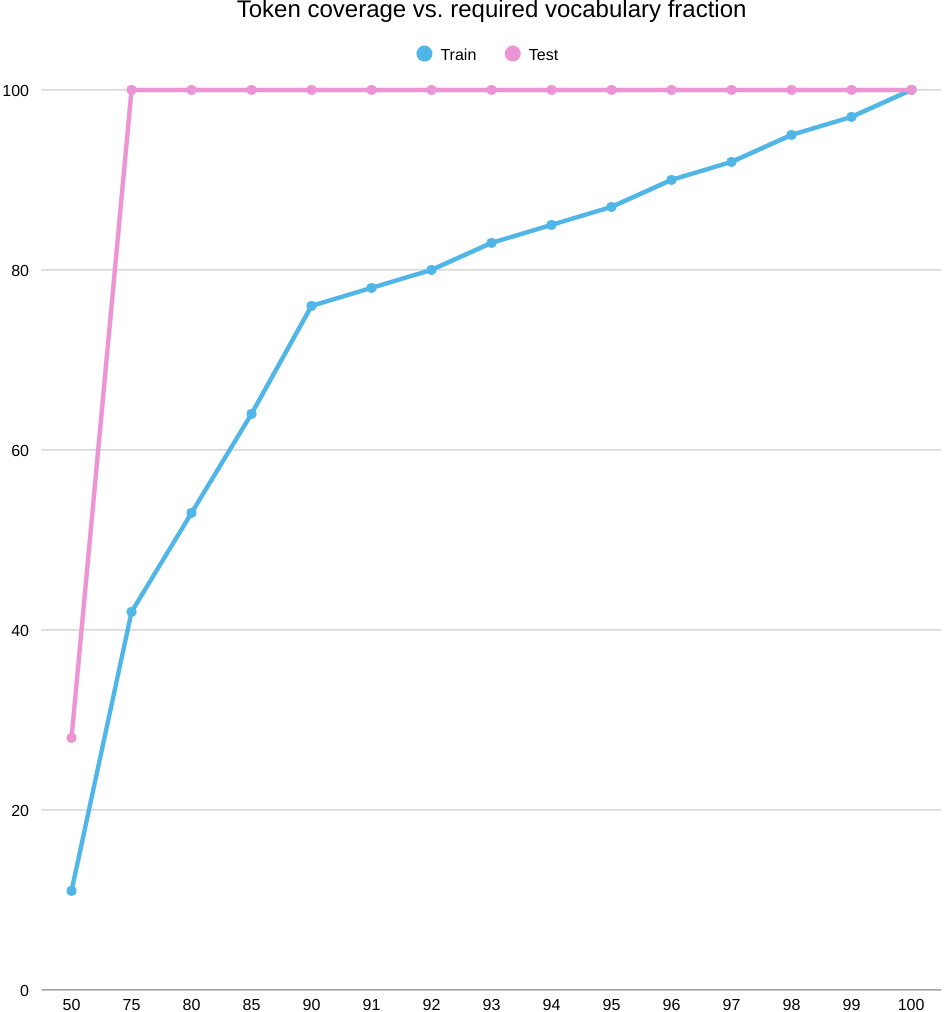}
\caption{Token coverage vs. required vocabulary fraction (K/|V|) for BOUN, train and test. Train coverage increases only gradually with larger K/|V|, while the test split accumulates coverage differently, indicating a mismatch between the ranked word list and the test distribution.}
\end{subfigure}
\hfill
\begin{subfigure}{0.48\linewidth}
\centering
\includegraphics[width=\linewidth]{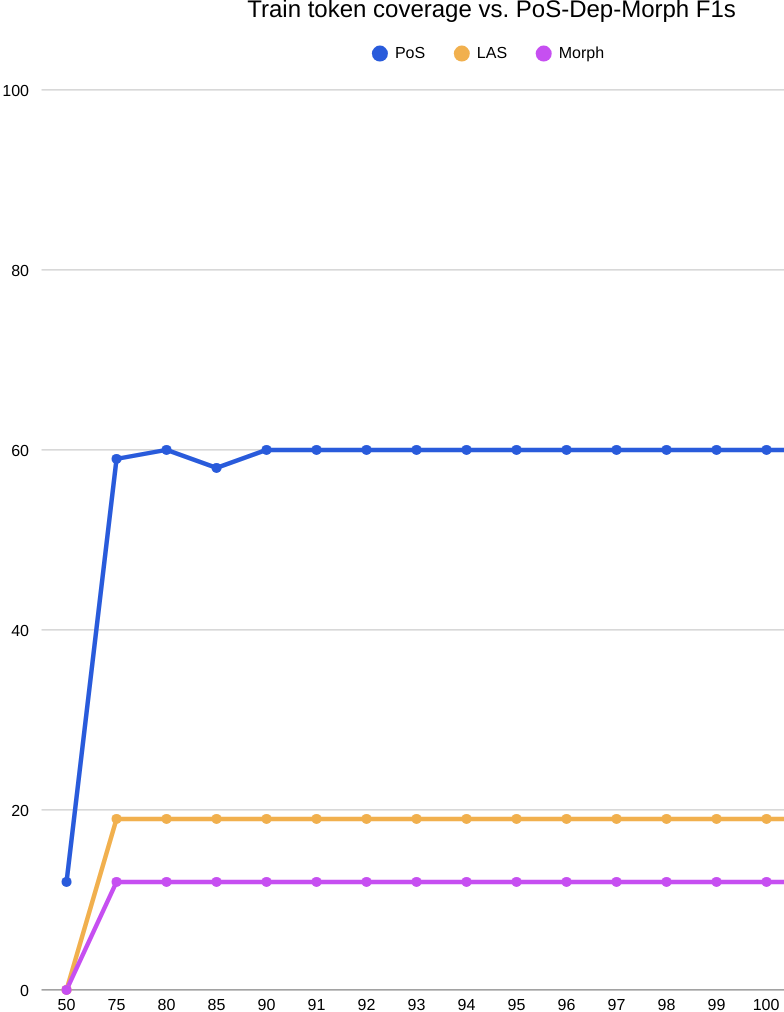}
\caption{Train token coverage vs. POS, LAS, and Morph F1. Each point uses the smallest top‑K word list achieving the indicated train coverage. All three metrics reach their plateau by 75\% coverage and remain essentially flat thereafter (POS $\approx$60, LAS $\approx$19, Morph $a\approx$12).}
\end{subfigure}
\caption{BOUN word‑level vocabulary: coverage efficiency and downstream success. Left: achieving higher coverage requires retaining large fractions of the word list, and train–test coverage accumulates differently, evidencing distribution shift. Right: increasing train token coverage beyond 75\% does not translate into meaningful gains for POS, dependency (LAS), or morphology; performance plateaus at low levels, highlighting the inefficiency of word‑level vocabularies on this morphologically rich dataset.}
\label{fig:word-pos-success}
\end{figure}

We evaluate word-level tokenization on the BOUN treebank across POS tagging, dependency parsing (LAS), and morphological tagging, relating success to how much of the training token mass the vocabulary covers.
Given BOUN's rich morphology, we expect heavy-tailed surface forms to fragment the word vocabulary; Figure \ref{fig:word-pos-success} examine coverage efficiency and its impact on task performance.

The success curves are strikingly flat after 75\% train coverage: increasing the kept word fraction does not translate into better POS, LAS, or Morph F1; which stall at $\approx$ 60/19/12, far below the character‑level (91/65/96). This indicates that the limiting factor is not token mass coverage but the representation itself: in BOUN's morphology‑heavy setting, many inflected surface forms split each lemma into sparse word types, producing a heavy tail that the word list cannot generalize over. The train–test coverage mismatch reinforces this: coverage accumulated on train does not reflect what the model needs at test time. Practically, these results argue against word‑level vocabularies for POS/DEP/Morph on BOUN; character/subword models with explicit morphological supervision are necessary to capture inflectional variation and deliver competitive accuracy.

\subsubsection{Explainability}
We probe what the word-level model attends to by visualizing token-level attributions as heatmaps. Unless stated otherwise, weights are nonnegative “intensity-only” scores normalized per input (sum = 1), so darker shades indicate tokens the model relies on most for its decision.

\begin{figure}[ht!]
\centering
\begin{subfigure}{0.48\linewidth}
\centering
\includegraphics[width=\linewidth]{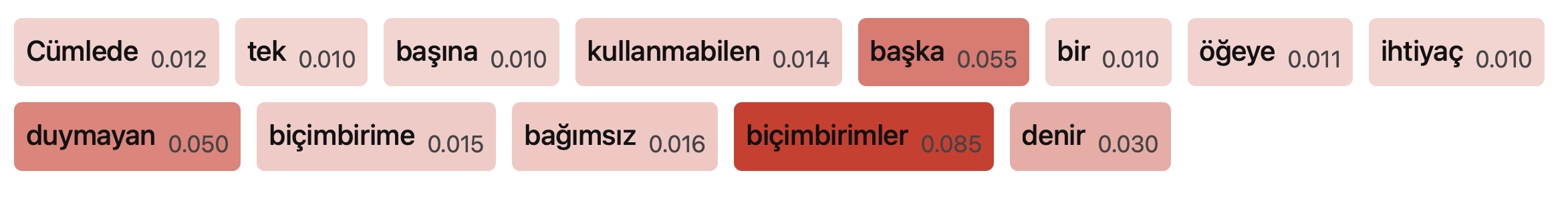}
\caption{CoLA (word-level, top-80\% vocab). Low-contrast, diffuse attributions: OOVs are only mildly weighted; “biçimbirime/biçimbirimler” receive modest intensity; no clear morphosyntactic signal.}
\end{subfigure}
\hfill
\begin{subfigure}{0.48\linewidth}
\centering
\includegraphics[width=\linewidth]{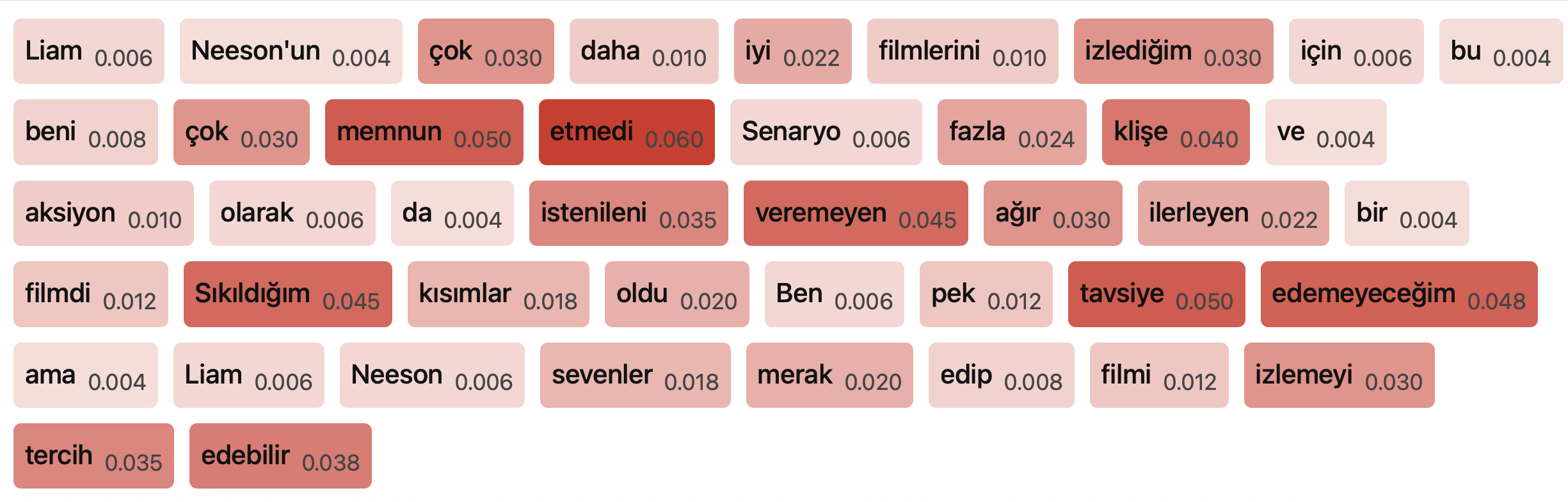}
\caption{SST‑2 (word-level, top-80\% vocab). Attribution concentrates on polarity cues and verbs, e.g., “memnun etmedi,” “klişe,” “ağır,” “Sıkıldığım,” “tavsiye edemeyeceğim,” “tercih/edebilir.”}
\end{subfigure}
\caption{Explainability heatmaps for CoLA vs. SST‑2 with word-level vocabularies. CoLA reveals weak, scattered cues consistent with poor grammatical acceptability performance, whereas SST‑2 focuses on a compact set of sentiment-bearing tokens, explaining the early performance gains and subsequent plateau once these cues are covered.}
\label{fig:expl-cola-sst2}
\end{figure}

The contrast between CoLA and SST‑2 is stark as exhibited in Figure \ref{fig:expl-cola-sst2}. For CoLA (top‑80\% vocab), weights are low-contrast and scattered: the three OOV forms (“Cümlede,” “kullanmabilen,” “öğeye”) receive only mild emphasis because many rare types are OOV at this coverage, while domain words like “biçimbirime/biçimbirimler” get slightly higher but still modest intensity—no coherent morphosyntactic cue emerges. For SST‑2, the model concentrates on sentiment-bearing cues and verbs: “memnun etmedi,” “klişe,” “ağır,” “Sıkıldığım,” “tavsiye edemeyeceğim,” and decision verbs like “tercih/edebilir” dominate, illustrating why sentiment reaches a performance plateau once these frequent polarity markers are covered.

We further probe sequence labeling models (NER and POS) using token-level intensity heatmaps, where nonnegative weights (normalized per input) indicate which tokens the model relies on when assigning tags.

\begin{figure}[ht!]
\centering
\begin{subfigure}{0.48\linewidth}
\centering
\includegraphics[width=\linewidth]{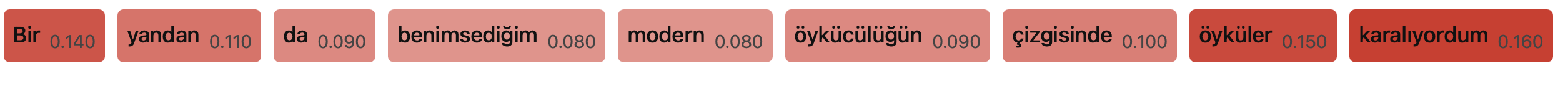}
\caption{NER example (top-80\% vocab). Dates (“1949'da”, “1 Ekim 1949'da”) receive the clearest emphasis; OOV entities (“Mao”, “Cumhuriyeti'ni”, “Pekin'i”) are de-emphasized, consistent with F1 $\approx$ 0.50.}
\end{subfigure}
\hfill
\begin{subfigure}{0.48\linewidth}
\centering
\includegraphics[width=\linewidth]{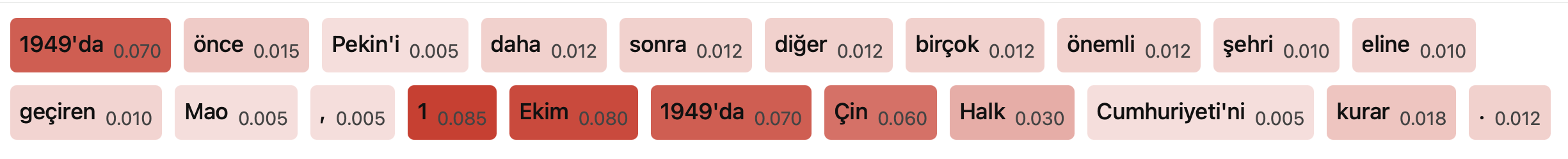}
\caption{POS example (top-80\% vocab). Emphasis shifts to in-vocab function words (“Bir”, “yandan”, “da”), while OOV content words receive flat, low weights, indicating limited morphological awareness.}
\end{subfigure}
\caption{Explainability for sequence labeling with word-level vocabularies. Left: NER focuses on easily recognizable spans (dates) and struggles with OOV named entities, yielding low, diffuse attention elsewhere. Right: POS relies on frequent function words available in the vocabulary and spreads weak intensity across OOV content words, reflecting limited generalization to unseen morphology under 80\% coverage.}
\label{fig:expl-ner-pos}
\end{figure}

As exhibited in Figure \ref{fig:expl-ner-pos}, across sequence labeling tasks, intensity patterns mirror coverage constraints. For NER, the model leans on unambiguous cues like numeric dates while underweighting OOV entities, which helps explain its modest F1. For POS, the model defaults to the few in-vocab function words as anchors and allocates nearly uniform, low weights to OOV content tokens, signaling poor morphosyntactic generalization under subword-sparse, word-level vocabularies. Together, these heatmaps show that when coverage is capped at 80\%, models prioritize frequent, surface-level signals and fail to exploit richer structure in OOV segments.

\subsubsection{Key Findings}
We distill general lessons from word-level modeling of Turkish and other morphologically rich languages:

\begin{itemize}
\item \textbf{Coverage-driven brittleness.} Fixed word vocabularies inevitably leave many inflected and derived forms out-of-vocabulary. When large portions of morphology collapse to \texttt{<unk>}, “being OOV” becomes commonplace and ceases to be informative, attenuating access to grammatical evidence.

\item \textbf{Shallow cues dominate when structure is hidden.} On structure-sensitive tasks (e.g., acceptability judgments, agreement checking), models struggle to capture suffix order, case, and lemma relations if these cues live primarily in unseen variants. Explanations mirror this with low-contrast, diffuse attributions.

\item \textbf{Sentiment is the exception, not the rule.} Tasks driven by frequent lexical markers (e.g., sentiment classification) remain comparatively robust. Explanations cluster around polarity-bearing words and common verbal constructions, suggesting reliance on high-frequency lexical cues rather than deeper composition.

\item \textbf{Entity-centric extraction degrades under OOV pressure.} For NER and related extraction tasks (e.g., slot filling, event arguments), proper names, titles, and multiword expressions are disproportionately OOV. Models over-weight easy numeric/date spans and under-weight rare names, yielding unstable boundaries and modest precision/recall—patterns clearly visible in attribution maps.

\item \textbf{From POS to syntax: limited generalization.} Beyond POS tagging, syntax-oriented tasks such as dependency parsing, chunking and  morph-feature prediction; suffer when content words are unseen. Models fall back to frequent function words as anchors while assigning flat, low intensity to OOV content tokens. This limits recovery of long-distance dependencies, agreement, and attachment decisions that hinge on morphology.

\item \textbf{Explainability mirrors learning dynamics.} Flat, low-contrast heatmaps indicate saturation on surface statistics without access to morphological signals hidden behind \texttt{<unk>}. Simply scaling data yields diminishing returns unless the representational units change.

\item \textbf{Error profile tracks the long tail.} Missed derivations, unstable case/possessive handling, and inconsistent proper-name treatment cluster in the long tail—exactly where word-level vocabularies provide the weakest support.

\item \textbf{Implication: re-think the units.} For morphologically rich languages, morphology-aware subwording (e.g., FST/analyzer-guided morpheme splits) is a more suitable inductive bias. It restores access to grammatical signals needed by syntax-oriented tasks while preserving the frequent lexical cues that already sustain sentiment and other surface-lexical phenomena.
\end{itemize}

\subsection{Morphology-Aware Subwords}
\label{sec:morpho-subwords}

\noindent We replace surface words with linguistically informed subword units derived from an FST and a spaCy-based morphologizer \autocite{Honnibal_spaCy_Industrial-strength_Natural_2020}, splitting stems and productive suffixes (e.g., negation \textit{-ma/-me}, tense/aspect/person, case/possessive). This representation dramatically reduces out-of-vocabulary pressure while preserving morphemic boundaries that carry core grammatical and lexical semantics. Compared to character models, sequences are shorter and better aligned with functional units; compared to word models, rare inflections and derivations are no longer collapsed into \texttt{<unk>}.

\subsubsection{TrGLUE }
\label{sec:morpho-subwords-trglue}
\begin{figure}[ht!]
\centering
\captionsetup[subfigure]{justification=centering}
\begin{subfigure}{0.32\linewidth}
\centering
\includegraphics[width=\linewidth]{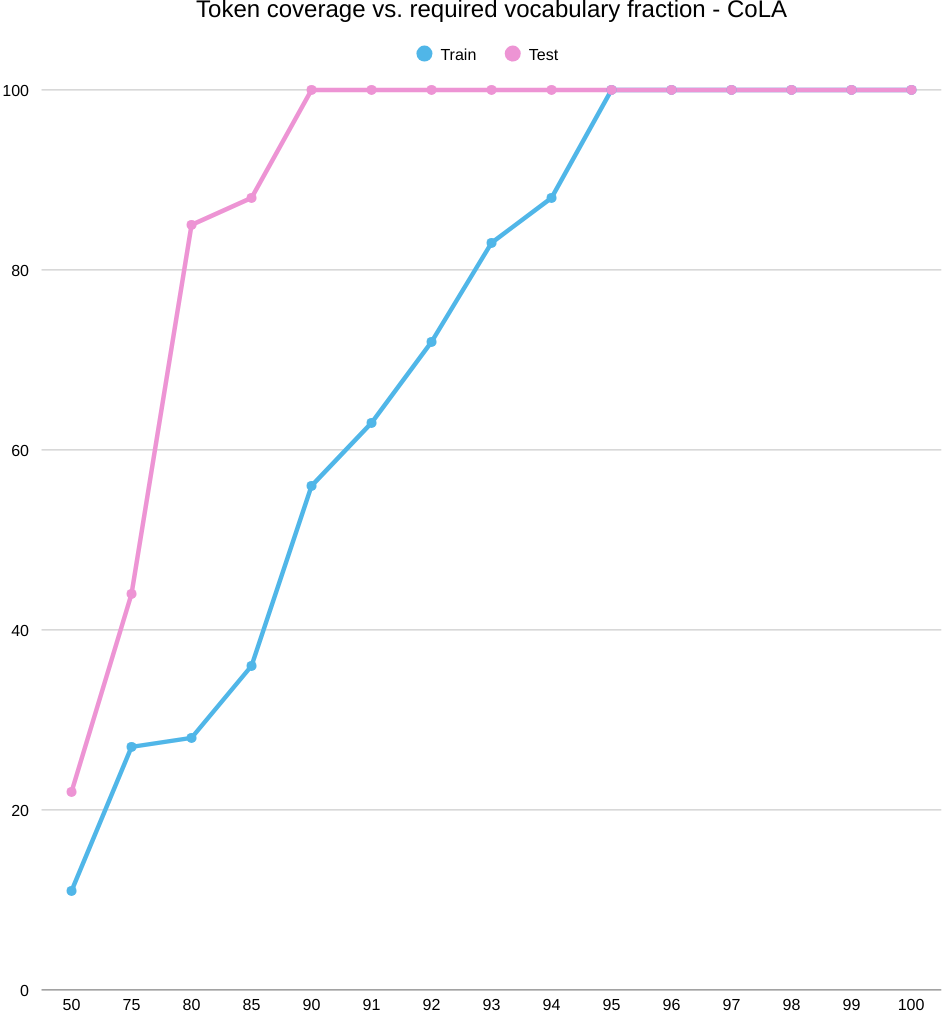}
\caption{CoLA: Token cov.\ vs.\ vocab. frac.}
\label{fig:cola-coverage}
\end{subfigure}\hfill
\begin{subfigure}{0.32\linewidth}
\centering
\includegraphics[width=\linewidth]{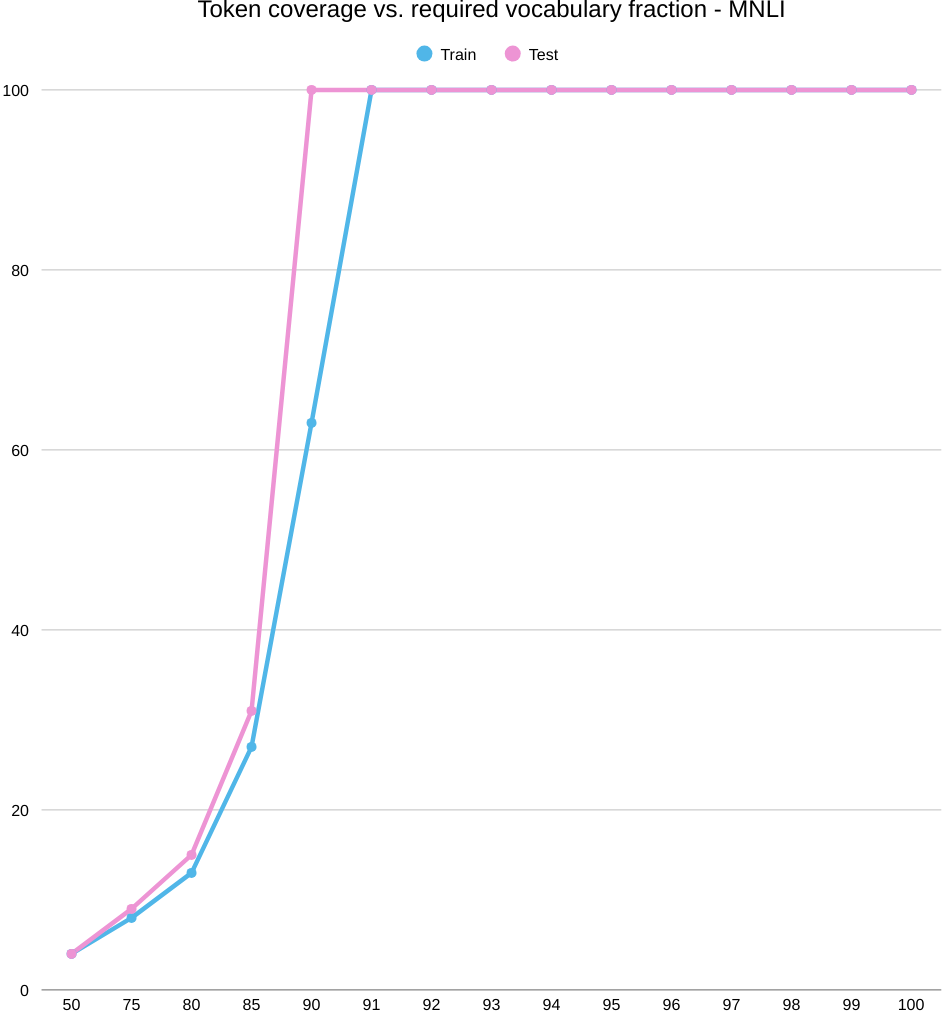}
\caption{MNLI: Token cov.\ vs.\ vocab. frac.}
\label{fig:mnli-coverage}
\end{subfigure}\hfill
\begin{subfigure}{0.32\linewidth}
\centering
\includegraphics[width=\linewidth]{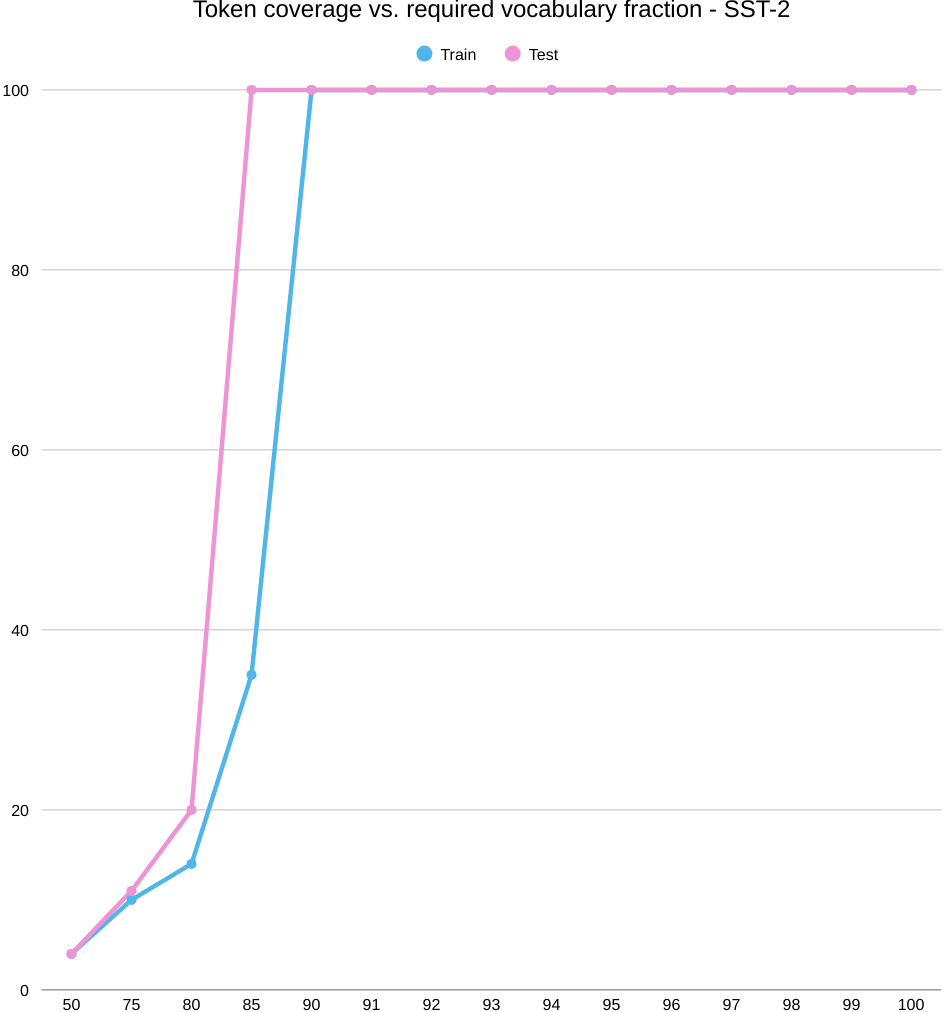}
\caption{SST-2: Token cov.\ vs.\ vocab. frac.}
\label{fig:sst2-coverage}
\end{subfigure}\hfill
\begin{subfigure}{0.32\linewidth}
\centering
\includegraphics[width=\linewidth]{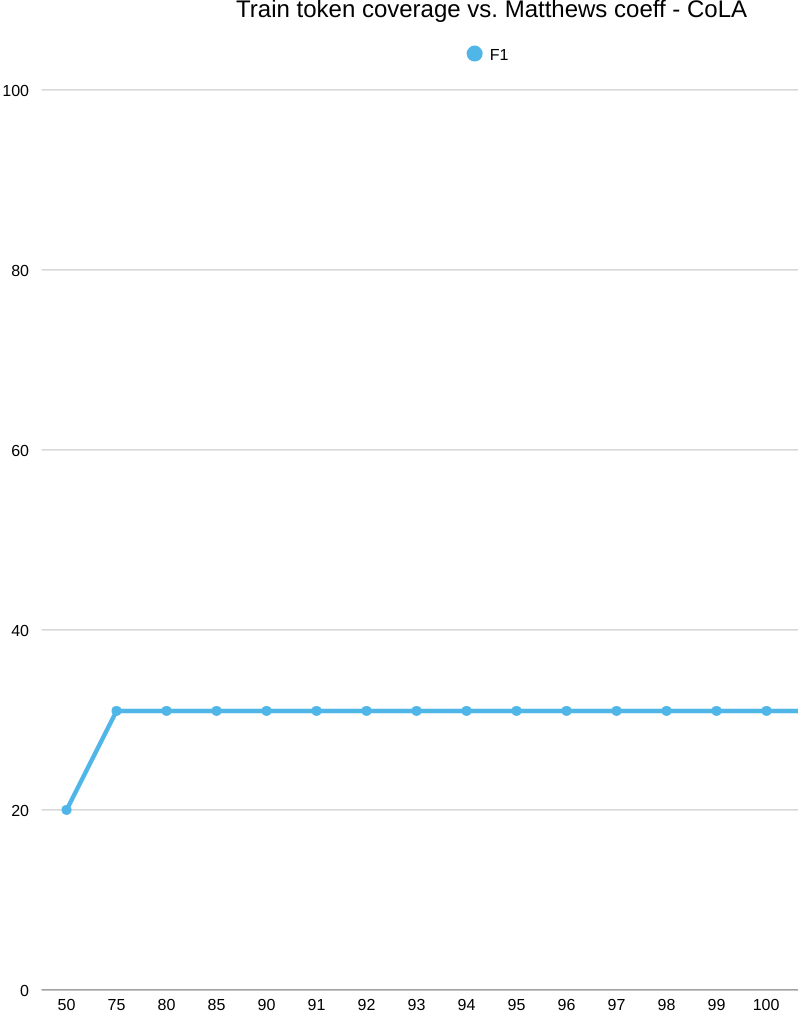}
\caption{CoLA: Train cov.\ vs.\ Matthews}
\label{fig:cola-train-mcc}
\end{subfigure}
\vspace{0.6em}
\begin{subfigure}{0.32\linewidth}
\centering
\includegraphics[width=\linewidth]{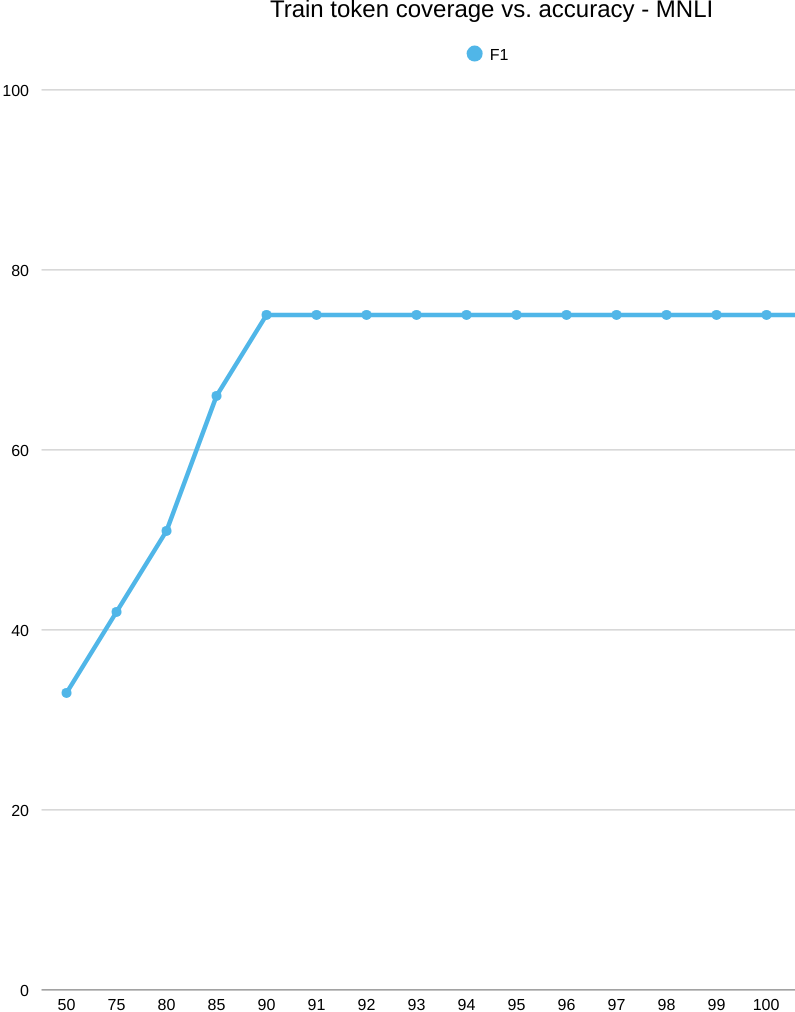}
\caption{MNLI: Train cov.\ vs.\ Acc}
\label{fig:mnli-train-acc}
\end{subfigure}\hfill
\begin{subfigure}{0.32\linewidth}
\centering
\includegraphics[width=\linewidth]{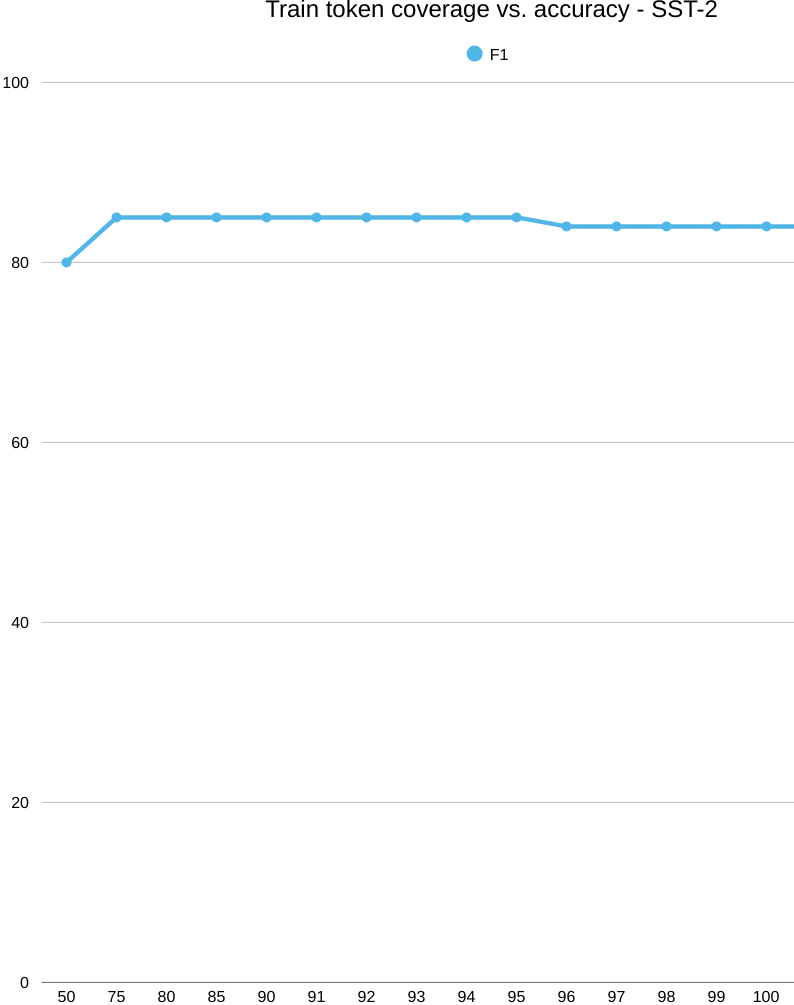}
\caption{SST-2: Train cov.\ vs.\ Acc}
\label{fig:sst2-train-acc}
\end{subfigure}
\caption{Morphology-aware subwords: coverage and performance across CoLA, MNLI, and SST-2. For each task, the coverage curve (above) precedes the performance-vs.-coverage plot (below), enabling a direct reading from vocabulary coverage to downstream effectiveness. Full-size per-task plots are provided in the appendix.}
\label{fig:morph-subword-2x3-coverage-first}
\end{figure}

Overall, morphology-aware subwords close much of the gap to pretrained transformers on structure- and morphology-sensitive tasks while yielding consistent gains over word- and character-level baselines. Figure \ref{fig:morph-subword-2x3-coverage-first} exhibits the model performance insights. 

Across CoLA, MNLI, and SST-2, morphology-aware subwords achieve high token coverage with a markedly compact vocabulary, and this efficiency translates into stable downstream performance. The coverage curves rise steeply—particularly for MNLI and SST-2—reaching near-complete train/test coverage after retaining only a small fraction of the morpheme inventory, indicating that a limited set of productive stems and suffixes accounts for most running tokens. In contrast to word-level tokenization, which requires an order-of-magnitude larger lexicon to obtain comparable coverage (and still suffers severe OOV under derivation and inflection), the morph-based inventory yields near-saturated coverage without ballooning the vocabulary. This compactness matters empirically: performance as a function of training coverage plateaus early for MNLI and SST-2, suggesting diminishing returns once the core morphemes are included, while CoLA shows a flatter but consistent trend, consistent with its sensitivity to agreement and case morphology rather than sheer lexical variety. Qualitatively, sentiment is carried by a small set of polarity-bearing morphemes (e.g., -ma/-me), explaining why SST-2 saturates quickly; MNLI benefits from early capture of negation/modality and case markers that stabilize contradiction/neutral decisions; and CoLA leverages person/number/case morphemes to localize acceptability violations. Overall, relative to word-level tokenization, morphology-aware subwords provide superior coverage at far lower vocabulary size, reduce OOV-driven variance, and yield comparable or better accuracy at a fraction of the lexical footprint, underscoring their efficiency–performance trade-off for Turkish.

For STS-B and MRPC, we observe broadly similar coverage and learning dynamics to MNLI: vocabulary saturates quickly with a compact morpheme inventory, and performance plateaus early. Concretely, the morph-subword BiLSTM attains 0.45 on STS-B and 0.62 on MRPC, consistent with the notion that morphology helps stabilize lexical variation while the remaining gap is largely semantic/pretraining-driven.

We now turn to sequence labeling, where morphology plays a more central role. The next section examines NER with morphology-aware subwords, quantifying how morpheme-level cues (e.g., case and possessive markers) improve boundary and type decisions relative to character and word baselines.

\subsubsection{NER}
We examine how pruning the morphology-aware subword vocabulary affects token coverage and NER performance, and compare these trends to a word-level baseline.

\begin{figure}[ht!]
\centering
\begin{subfigure}{0.48\linewidth}
\centering
\includegraphics[width=\linewidth]{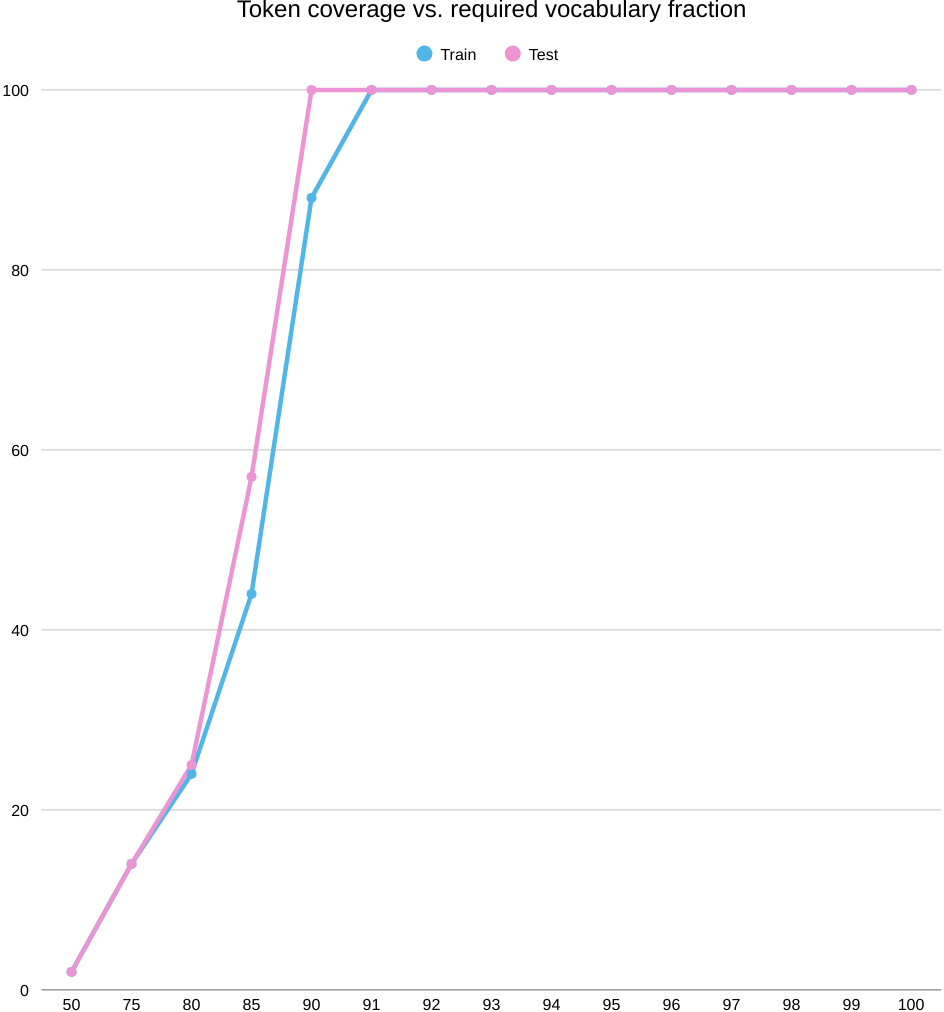}
\caption{Token coverage as a function of retained morpheme vocabulary on train and test. Coverage saturates around 88–90\% of the full vocabulary, indicating that a compact core yields near‑perfect coverage on both splits.}
\end{subfigure}
\hfill
\begin{subfigure}{0.48\linewidth}
\centering
\includegraphics[width=\linewidth]{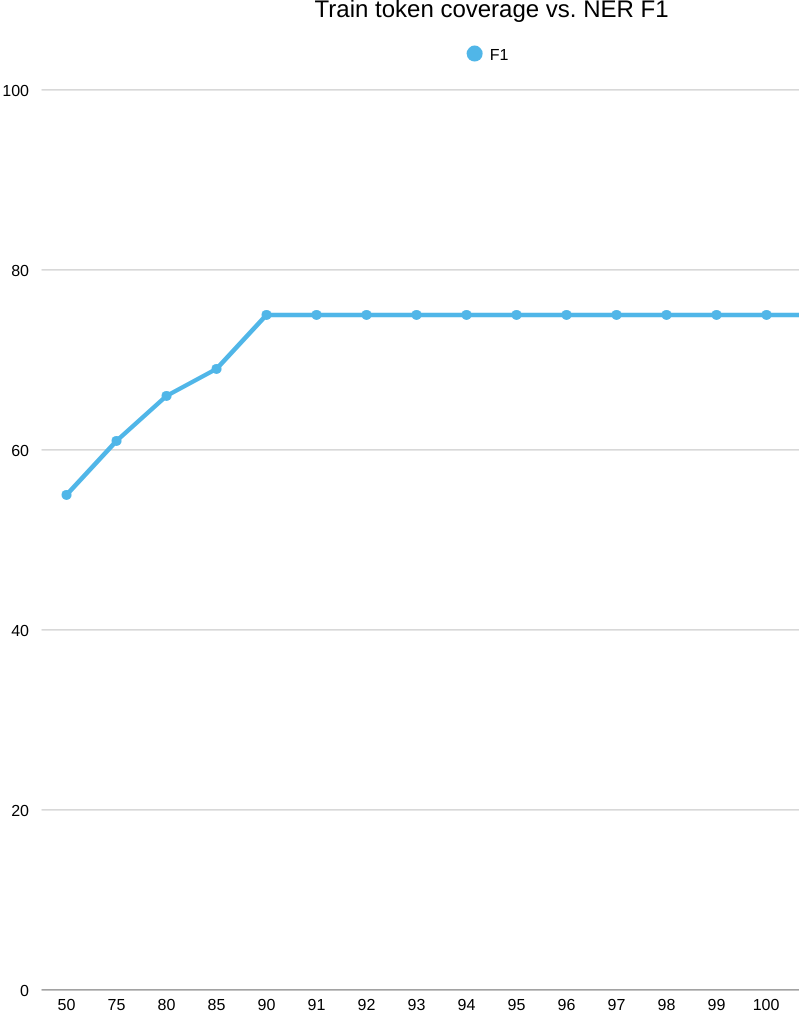}
\caption{NER span‑F1 versus train token coverage. F1 rises rapidly up to 80–85\% coverage and then plateaus, showing diminishing returns from rare morphemes beyond this point.}
\end{subfigure}
\caption{Effect of morpheme vocabulary pruning on coverage and NER performance. A small subset of frequent morphemes achieves near‑maximal token coverage and F1, enabling aggressive pruning without loss in accuracy.}
\label{fig:ner-subword-success}
\end{figure}

Across pruning levels, morphology-aware subwords maintain high token coverage with a relatively small vocabulary and convert that coverage into early NER gains: moving from 55\% to 85\% coverage boosts F1 from the mid‑50s to 75, after which performance stabilizes despite further increases in coverage. Compared to a word-level model, the morph-subword system achieves comparable or higher F1 with far fewer lexical types and thus better data efficiency; in practice, retaining only 88–90\% of the morpheme inventory matches the near‑maximal performance while keeping the vocabulary substantially smaller than the full word lexicon. This suggests that most NER signal resides in frequent morphemes (case, possessive, derivational cues), and that pruning rare morphemes yields parameter and speed benefits without sacrificing accuracy.

\subsubsection{POS-DEP-Morph}
\begin{figure}[ht!]
\centering
\begin{subfigure}{0.48\linewidth}
\centering
\includegraphics[width=\linewidth]{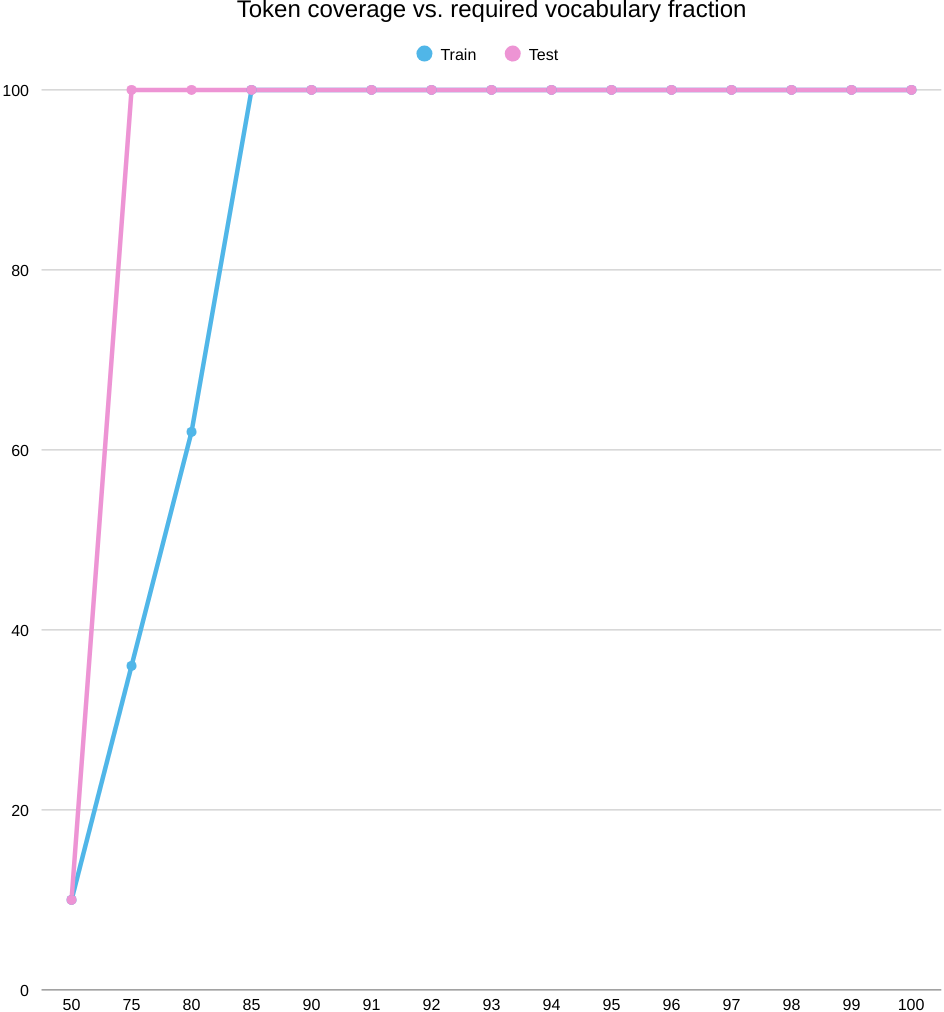}
\caption{Train token coverage vs. task performance. POS and Morph F1s climb steeply and saturate by 85–88\% coverage; LAS remains stable across pruning once coverage is high.}
\end{subfigure}
\hfill
\begin{subfigure}{0.48\linewidth}
\centering
\includegraphics[width=\linewidth]{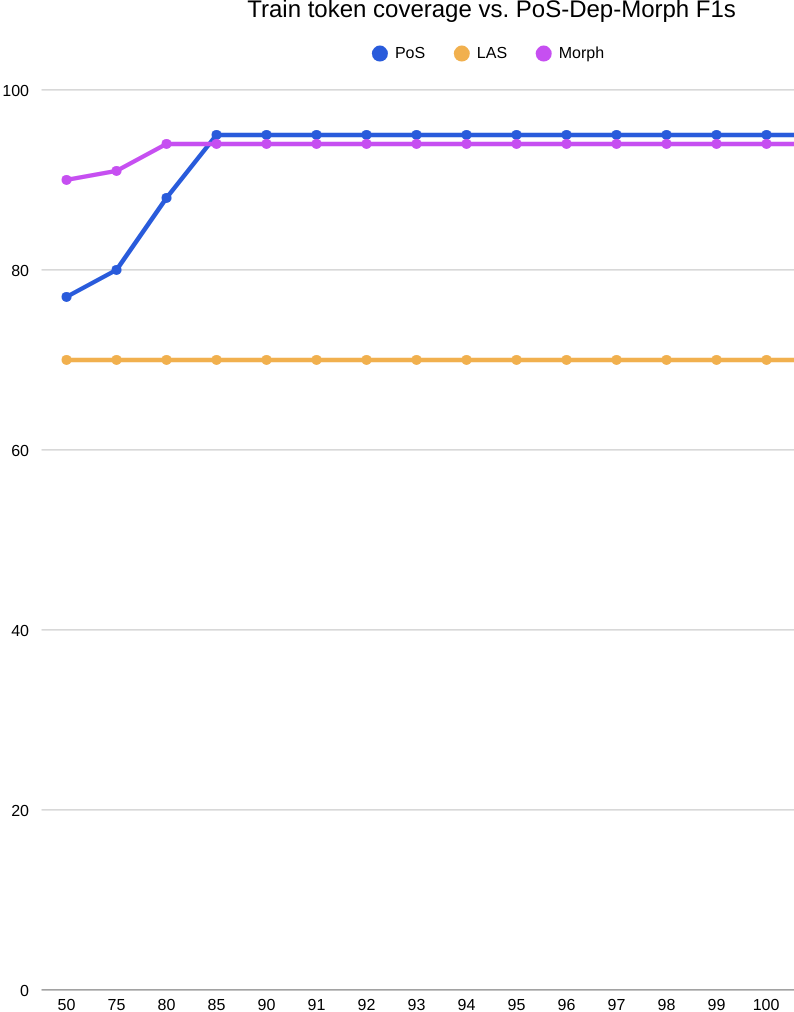}
\caption{Token coverage as a function of retained morpheme vocabulary. Test coverage reaches 100\% by 2–75\% of the vocabulary; train coverage hits 100\% around 83–85\%}
\end{subfigure}
\caption{POS-DEP-Morph under morpheme-vocabulary pruning. A compact morpheme set attains near-max coverage and task performance, enabling aggressive pruning without accuracy loss.}
\label{fig:pos-subword-success}
\end{figure}

We analyze how pruning the morphology-aware subword inventory trades off token coverage with POS/LAS/Morph performance, and compare these trends to a word-level vocabulary, given in Figure \ref{fig:pos-subword-success}.

The curves show that morphology-aware subwords yield rapid gains as coverage increases: POS and Morph F1 improve sharply from low coverage to about 80–85\%, after which they plateau at their near-maximum values; LAS is comparatively flat once coverage is adequate, reflecting that core case/possessive markers needed for head selection are already retained early. Coverage itself is achieved with a small inventory—on the test set, near-complete coverage arrives by roughly three-quarters of the morpheme vocabulary, and the train split reaches 100\% by about 83–85\%. Relative to a word-level model, the morph-subword approach attains comparable or better POS/Morph/LAS with far fewer lexical types, meaning fewer parameters and better generalization on OOV stems; in practice, pruning to the 80–85\% range preserves performance while reducing vocabulary size dramatically compared to the full word lexicon.

\subsubsection{Explainability}
To understand how morphology-aware subwords drive decisions beyond aggregate metrics, we visualize per-(sub)token attributions across classification (TrGLUE) and token-level tasks (NER/POS), using a unified color scale so contributions are comparable within and across panels.

\begin{figure}[ht!]
  \centering
  \begin{subfigure}[ht!]{0.48\textwidth}
    \centering
    \includegraphics[width=\linewidth]{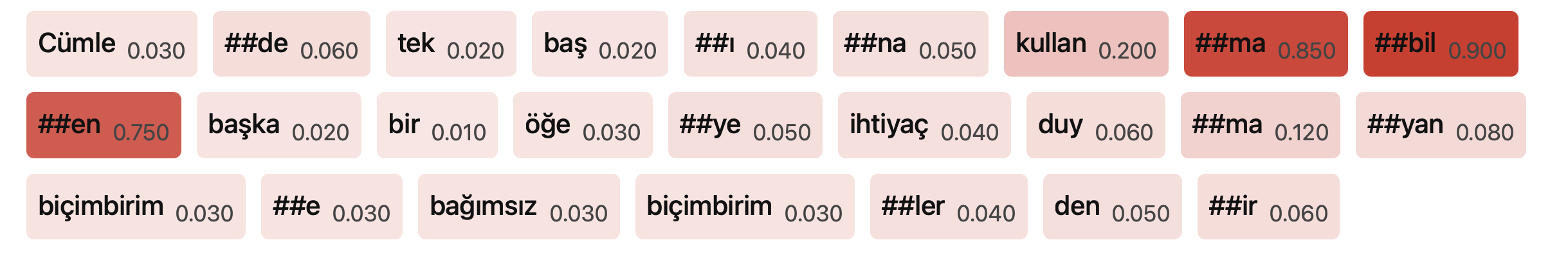}
    \caption{CoLA token/morpheme attributions.}
    \label{fig:glue-a}
  \end{subfigure}\hfill
  \begin{subfigure}[ht!]{0.48\textwidth}
    \centering
    \includegraphics[width=\linewidth]{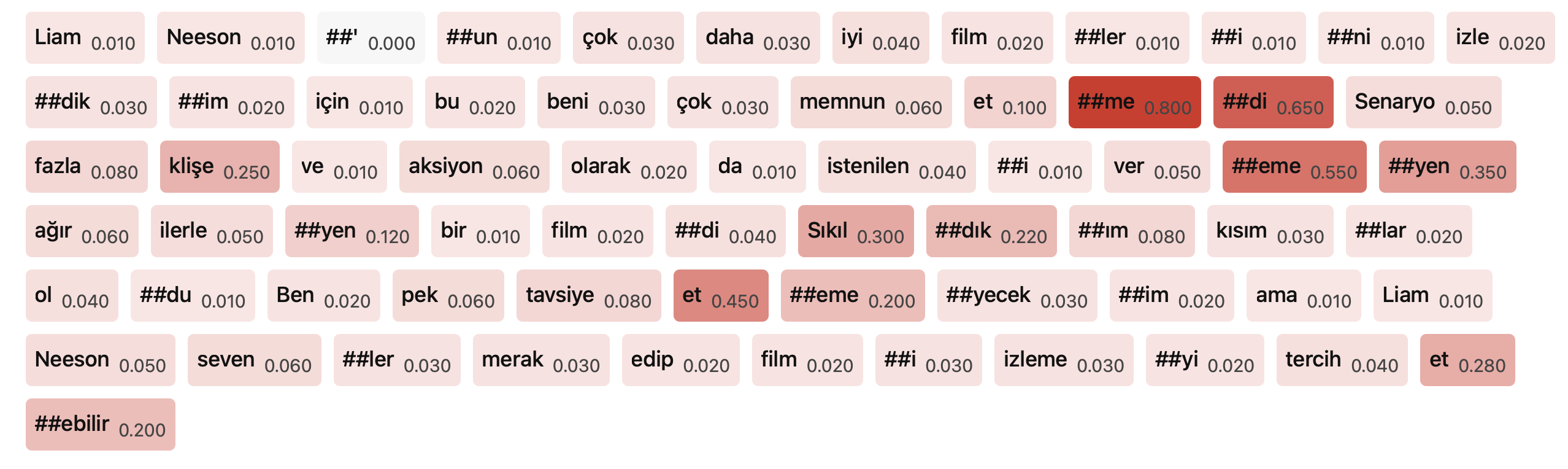}
    \caption{SST-2: token/morpheme attributions.}
    \label{fig:glue-b}
  \end{subfigure}
  \caption{Explainability on GLUE tasks. Heatmaps show per-(sub)token attribution with consistent colormap and scale; darker indicates stronger positive contribution.}
  \label{fig:subwords-glue-explain}
\end{figure}

As shown in Figure \ref{fig:subwords-glue-explain}, across both TrGLUE examples, high positive mass concentrates on morphologically salient units that disambiguate polarity and entailment cues: negation morphemes, derivational markers shifting part-of-speech, and clause-level connectives receive strong weights, while function words without semantic load remain muted. We also see complementary evidence aggregation: stems contribute broad semantics, while suffixes tilt the decision (e.g., polarity or modality), indicating that morph-subwords capture compositional meaning rather than relying on surface word forms alone.

\begin{figure}[ht!]
  \centering
  \begin{subfigure}[ht!]{0.48\textwidth}
    \centering
    \includegraphics[width=\linewidth]{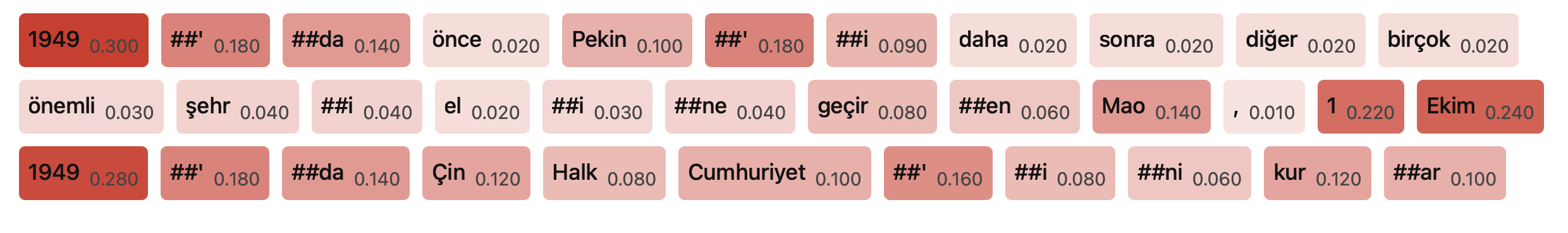}
    \caption{NER: morpheme-level attributions highlighting dates, apostrophes, and case markers.}
  \end{subfigure}\hfill
  \begin{subfigure}[ht!]{0.48\textwidth}
    \centering
    \includegraphics[width=\linewidth]{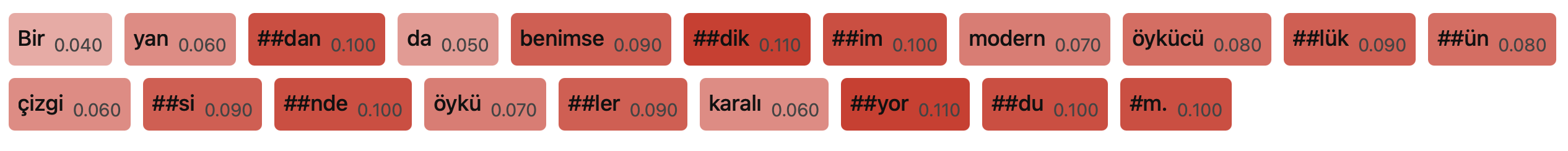}
    \caption{POS: contributions of derivational/inflectional suffixes to tag decisions.}
  \end{subfigure}
  \caption{Explainability on token-level tasks. We visualize how morphology-aware subwords guide boundary/type (NER) and category (POS) predictions.}
  \label{fig:subwords-ner-pos-explain}
\end{figure}

In NER, date numerals, month names, and apostrophe-plus-case sequences carry the highest weights, clarifying entity boundaries and types; ablative/locative and possessive markers reduce boundary fragmentation and improve attachment. In POS, inflectional endings for number, case, and tense (e.g., -ler, -de/-da, -yor, -du) consistently dominate attributions over stems, especially for nouns and verbs with ambiguous stems, showing that the model anchors syntactic category decisions in morphology rather than capitalization or context heuristics, as shown in Figure \ref{fig:subwords-ner-pos-explain}.

Taken together, the visualizations show a coherent strategy: stems provide coarse semantics, while a compact set of frequent morphemes supplies decisive, task-specific cues (negation/modality for GLUE, case/possessive/TAM for NER/POS). This aligns with our pruning results—once the core morphemes are retained, models remain accurate and interpretable, suggesting that morphology-aware tokenization yields both efficiency and transparent, linguistically grounded behavior.

\subsubsection{Key Findings}
Morphology-aware subword modeling delivers a compact vocabulary without sacrificing coverage or accuracy. By retaining only the frequent morphemes ($\approx$ 75–90\% of the morpheme inventory), we achieve near-perfect token coverage on both train and test and hit plateaus on NER, POS, Morph, and LAS—showing that a small, linguistically meaningful core carries most of the predictive signal. This compression reduces parameters tied to embeddings/softmax, lowers memory and I/O, and improves robustness on OOV stems by recombining known morphemes.

Compared to a word-level vocabulary, morph subwords reach comparable-or often better-performance with dramatically fewer lexical types. Word vocabularies balloon with inflectional variants and rare forms, leading to sparse embeddings and brittle OOV handling; in contrast, morph subwords generalize across forms via shared affixes (case, possessive, derivation), yielding higher data efficiency and steadier performance on rare words and domains. In practice, pruning to the core morphemes offers the best trade-off: smaller models, faster inference, and equal or better accuracy than word-based tokenization.

\section{WordPiece Tokenization}
In this section, we examine the three-way interaction among vocabulary size, tokenizer training data size, and morphological alignment. We ask: (i) does aligning WordPiece segments with morphological subwords improve downstream performance? and (ii) how large should the tokenizer training corpus be—does more data always help? To make these aims concrete and evaluable across tasks, we frame the agenda as the following research questions:
\begin{itemize}
  \item \textbf{RQ1} How do tokenizer size and type (morphology-aware vs.\ WordPiece) affect downstream performance across syntax/morphology-sensitive tasks versus semantics/entity-oriented tasks?
  \item \textbf{RQ2} Do tokenizers with stronger Turkish morphological alignment yield larger gains on POS/DEP/Morph than on NER/STS-B/sentiment?
  \item \textbf{RQ3} How does tokenizer training corpus size (5/20/80 GB) interact with vocabulary size to trade off sequence length and and morphological fidelity?
  \item \textbf{RQ4} What Pareto frontier emerges between minimal sequence length, maximal morphological alignment, and downstream accuracy?
\end{itemize}

To operationalize these questions, the next subsection details the pretraining corpora used to train tokenizers and associated Transformers.

\subsection{Pretraining Corpus}
For training the tokenizers and pretraining the transformers in this section, we use high-quality Turkish text from the BellaTurca collection \autocite{bellaturca}, a large-scale corpus resource. We include three genres in total: high-quality web data, books, and cleaned OSCAR web data.

High-quality web data comes from a subcollection extensively filtered during crawl to prioritize the best-quality web content. This portion has a total size of 4.6 GB, 1.3M documents, and 557M words. To reach approximately 5 GB, we add an academic/scientific web subcollection, contributing around 910 MB, 500K documents, and 90.6M words. The combined web total is then 5.5 GB, 1.8M documents, and 648M words.

Next, we use data from the books subcollection. This portion is about 15 GB, 100M documents (each sentence treated as a document), and 1.43B words. The genre is books and the data quality is high.

Finally, we include a cleaned OSCAR subcollection, derived from several Turkish OSCAR releases via intensive text cleaning and quality filtering. Its total size is 57 GB, 23.7M documents, and 7.15B words.

In our experiments, we define three corpus scales: Minimal, Medium, and Alldata, with sizes around 5 GB, 20 GB, and 80 GB. Minimal consists of the filtered web data (plus the academic addition). Medium adds the books portion. Alldata includes web, books, and cleaned OSCAR. In the following, we use the terms “Minimal,” “Medium,” and “Alldata” to refer to these three pretraining corpora.

\subsection{Training WordPiece Tokenizers}
We train WordPiece tokenizers across two axes: vocabulary size and corpus scale. For vocabulary size, we consider 2k, 5k, 10k, 20k, 32k, 52k, and 128k to span very small to very large inventories. For corpus scale, each vocabulary size is trained on three pretraining corpora: Minimal ($\approx$5 GB), Medium ($\approx$20 GB), and Alldata ($\approx$80 GB) as defined above.

Tokenizers are trained using the Hugging Face tooling (Tokenizers and Trainer API) \autocite{wolf-etal-2020-transformers} and prepared with the Hugging Face libraries for downstream use. With these trained variants in hand, we analyze coverage, OOV rate, sequence length, and morphology alignment, and relate these properties to downstream performance in subsequent sections.

\subsection{Tokenization Behavior Across Corpora and Vocabulary Sizes}
The empirical patterns in Turkish segmentation reveal a clear progression as vocabulary size grows and as the training corpus diversifies. With very small vocabularies (2–10k) trained on large, heterogeneous corpora (e.g., Medium, Alldata), the tokenizer tends to operate close to the character level. Even highly frequent stems such as \emph{ev} or \emph{oku} are decomposed as [e, \#\#v] and [o, \#\#k, \#\#u, \ldots], respectively. This regime inflates sequence length—fertility hovers around 6.5 and the proportion of continued subwords is near 0.98—while dispersing the morpheme-level cues that are often decisive for downstream syntax-sensitive tasks. In other words, the model must infer morphological structure from many short fragments, a strategy that is data-hungry and tends to dilute signal.

As we move to mid-size vocabularies (20–32k), segmentation stabilizes around stems and recurrent suffixes. Common locatives and ablatives (\emph{-de/-da}, \emph{-den/-dan}), plural and possessive series (\emph{-lar}, \emph{-(i)miz}), and clitic-like elements (\emph{-ki}, \emph{-(y)a}) begin to surface as consistent subword units. Typical forms such as \emph{ev}, \emph{evde}, and \emph{evden} become whole tokens; moderately frequent inflections like \emph{okudum} are often [oku, \#\#dum] or whole. Fertility drops into a more interpretable band (roughly 1.4–1.7 depending on corpus), and the continued-subword rate settles between 0.30 and 0.48. This shift indicates a healthier balance: sequences are shorter, and morphological boundaries remain visible enough to support generalization across productive paradigms.

At large vocabularies (52–128k), the tokenizer increasingly memorizes whole words and frequent inflected forms. Items such as \emph{okudum}, \emph{okudular}, \emph{görülebilirdi}, \emph{evim}, and \emph{evimiz} are often single tokens. The resulting fertility pushes toward 1.15–1.18 with continued rates near 0.12–0.14. While this compression is attractive for efficiency, it sometimes fuses morphemes idiosyncratically—for instance, \emph{evleriniz} realized as [evlerini, \#\#z]—thereby weakening the model's explicit access to morpheme boundaries in rarer or compositional variants. Syntactic tasks (POS, NER) that benefit from overt morphological cues can be disproportionately affected by such over-merging, whereas semantic tasks are generally more tolerant of it.

Orthographic conventions mediate these effects. Proper-name constructions with apostrophes (\emph{İstanbul'da}, \emph{Ankara'dan}) reliably split at the apostrophe across sizes, which is correct. Numerals with apostrophes (\emph{1923'te}) likewise exhibit a stable pattern—numbers remain intact and the suffix attaches as a separate unit—preserving the syntactic signal while controlling vocabulary growth.

\begin{tcolorbox}[colback=gray!3!white,colframe=gray!50!black,title={Illustrative tokenization snapshots},left=1mm,right=1mm]
\footnotesize
\textbf{Small vocabulary (2–10k; medium/alldata)}\\
\emph{ev} $\rightarrow$ [e, \#\#v]\\
\emph{evim} $\rightarrow$ [e, \#\#v, \#\#i, \#\#m]\\
\emph{evlerimizden} $\rightarrow$ [e, \#\#v, \#\#l, \#\#e, \#\#r, \#\#i, \#\#m, \#\#i, \#\#z, \#\#d, \#\#e, \#\#n]\\
\emph{okudum} $\rightarrow$ [o, \#\#k, \#\#u, \#\#d, \#\#u, \#\#m]\\
\emph{1923'te} $\rightarrow$ character level sequence\\
\emph{yazdırılmayacakmışsınız} $\rightarrow$ near character-level sequence\\[4pt]

\textbf{Mid vocabulary (20–32k)}\\
\emph{ev} $\rightarrow$ [ev]\\
\emph{ev} $\rightarrow$ [ev, \#\#im]\\
\emph{evlerimizden} $\rightarrow$ [ev, \#\#lerimiz, \#\#den] \; or \; [ev, \#\#lerimizden]\\
\emph{okudum} $\rightarrow$ [ok, \#\#u, \#du, \#m]\\
\emph{1923'te} $\rightarrow$ [192, \#\#3, ', te]
\emph{görülebilirdi} $\rightarrow$ [görülebilir, \#\#di]\\
\emph{çalıştırılabilir} $\rightarrow$ [çalıştır, \#\#ılabilir] \\[4pt]

\textbf{Large vocabulary (52–128k)}\\
\emph{ev}, \emph{evim}, \emph{okudum}, \emph{okudular}, \emph{kapkara} $\rightarrow$ single tokens\\
\emph{evleriniz} $\rightarrow$ [evlerini, \#\#z] (example of over-fusion)\\
\emph{evlerimizden} $\rightarrow$ [evlerimiz, \#\#de] (another example of over-fusion)\\
\emph{1923'te} $\rightarrow$ [1923, ', te]
\end{tcolorbox}

\subsubsection{Corpus Size, Vocabulary Size, and Fragmentation}
The interaction between corpus size and vocabulary size is not linear; more data is only beneficial if the vocabulary scales accordingly. With small vocabularies on large, diverse corpora, the tokenizer cannot “afford” morphemes as reusable units and so defaults to short fragments, producing very high fertility (around 6.5) and almost entirely continued tokens, as quantified in Table~\ref{tab:granularity-metrics}. This configuration increases training cost and tends to obscure systematic morphology, pressuring the model to learn it implicitly from long sequences of micro-fragments.

\begin{figure}[ht!]
  \centering
  \subfloat[Fertility versus vocabulary size across corpus-size buckets. Points show mean fertility at each vocabulary size; colors denote corpus size (Minimal, Medium, Alldata). Lower values indicate fewer subwords per word.\label{fig:left}]{%
    \includegraphics[width=0.48\textwidth]{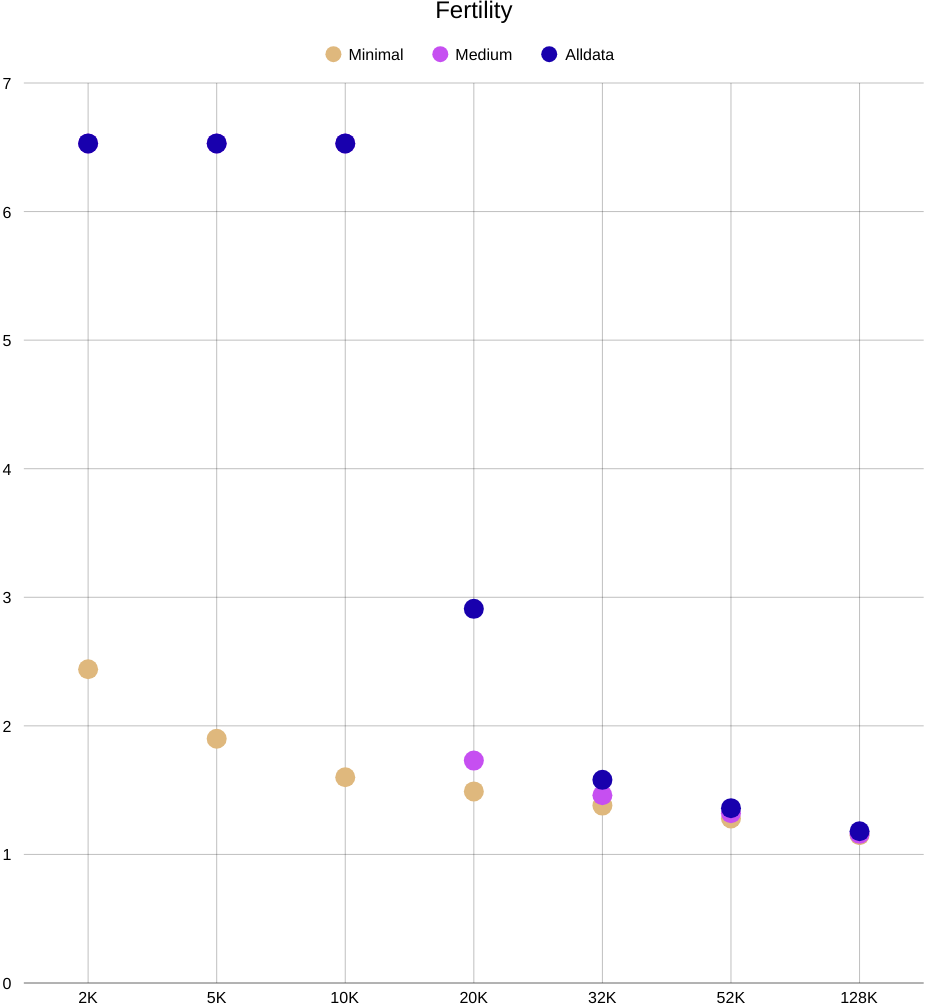}}
  \hfill
  \subfloat[Continuation rate versus vocabulary size across corpus-size buckets. Each point is a model at a given vocabulary size; colors denote corpus size. Continuation rate is the fraction of subwords that are continuations (0–1)\label{fig:right}]{%
    \includegraphics[width=0.48\textwidth]{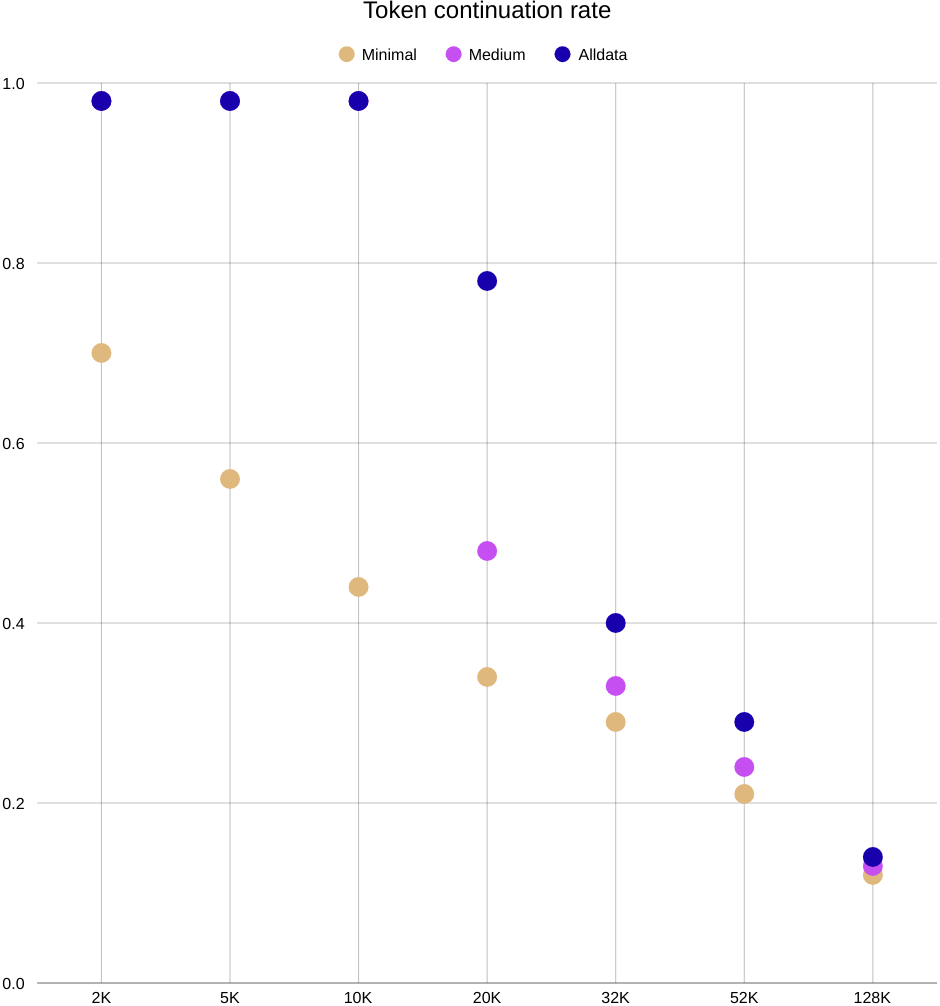}}
  \caption{Effect of WordPiece vocabulary size on fertility (left) and continuation rate (right) across Minimal, Medium, and Alldata corpora; mid-size vocabularies balance sequence length and morpheme visibility.}
  \label{fig:fert-cont-tokenizer}
\end{figure}

A mid-size vocabulary establishes a more favorable equilibrium. On smaller or cleaner corpora, fertility can quickly settle near 1.5 with continued rates around 0.3–0.4, as shown in Figure~\ref{fig:fert-cont-tokenizer}; a region where suffixes are explicitly represented and stems remain stable. On larger and more heterogeneous corpora, achieving comparable fertility requires a larger vocabulary; for instance, moving from 20k to 32k can substantially reduce over-fragmentation without collapsing morpheme boundaries. This regime offers a pragmatic balance for both pretraining efficiency and downstream accuracy across syntactic and semantic tasks.

At the high end, very large vocabularies minimize sequence length but risk eroding compositionality by baking frequent inflections into whole-word entries. The resulting token economy is efficient, yet the loss of morpheme visibility can degrade generalization to rarer inflectional combinations and attenuate the explicit signals that aid tagging and parsing. In practice, the impact is task-dependent: syntax-heavy tasks prefer the mid-range where morphology remains legible, whereas purely semantic tasks may benefit more from shorter sequences and tolerate morpheme fusion.

These patterns together trace a practical Pareto frontier. For Turkish, we consistently observe that fertility around 1.4–1.7 with continued-subword rates roughly 0.30–0.45 yields a strong trade-off: sequences are compact enough for efficient training while preserving the morphological structure crucial to syntactic competence. On smaller ($\approx$ 5\,GB) and medium ($\approx$ 20\,GB) corpora, this frontier is typically achieved with 20–32k vocabularies; on very large ($\approx$ 80\,GB) corpora, 32–52k often hits the same target. 

\begin{table}[ht!]
\centering
\small
\caption{Fertility and continuation rate across vocabulary sizes (mean $\pm$ sd).}
\begin{tabular}{lcccccc}
\toprule
& \multicolumn{2}{c}{minimal} & \multicolumn{2}{c}{medium} & \multicolumn{2}{c}{alldata} \\
\cmidrule(lr){2-3}\cmidrule(lr){4-5}\cmidrule(lr){6-7}
Vocab & Fert. & Cont. & Fert. & Cont. & Fert. & Cont. \\
\midrule
2k   & 6.3$\pm$0.2 & 0.97$\pm$0.01 & 6.5$\pm$0.2 & 0.98$\pm$0.01 & 6.6$\pm$0.2 & 0.98$\pm$0.01 \\
10k  & 3.6$\pm$0.1 & 0.69$\pm$0.02 & 3.8$\pm$0.1 & 0.72$\pm$0.02 & 3.9$\pm$0.1 & 0.73$\pm$0.02 \\
20k  & 2.0$\pm$0.1 & 0.39$\pm$0.02 & 2.1$\pm$0.1 & 0.42$\pm$0.02 & 2.3$\pm$0.1 & 0.45$\pm$0.02 \\
32k  & 1.5$\pm$0.1 & 0.31$\pm$0.02 & 1.6$\pm$0.1 & 0.35$\pm$0.02 & 1.7$\pm$0.1 & 0.38$\pm$0.02 \\
52k  & 1.3$\pm$0.0 & 0.16$\pm$0.01 & 1.3$\pm$0.0 & 0.18$\pm$0.01 & 1.4$\pm$0.0 & 0.20$\pm$0.01 \\
128k & 1.14$\pm$0.0 & 0.12$\pm$0.01 & 1.15$\pm$0.0 & 0.13$\pm$0.01 & 1.18$\pm$0.0 & 0.14$\pm$0.01 \\
\bottomrule
\end{tabular}
\label{tab:granularity-metrics}
\end{table}

\subsubsection{Tokenizer Morphology Diagnostics: Results and Discussion}
\label{subsec:tokenizer-morphology-results}

We instantiate the metrics from Section \ref{subsec:morpho-metrics} on Turkish morphology testset and compare WordPiece tokenizers trained on multiple corpora and vocabulary sizes. We report segmentation granularity (Subwords/Word), boundary alignment (micro/macro PRF), lemma integrity (lemma\_single\_rate, lemma\_boundary\_rate), sequence agreement (exact\_morph\_sequence\_match, WER metrics (WER, MER, CER, WIL, WIP), and affix signals (coverage/atomicity over top-200 suffixes), together with over/under-segmentation indices.

We summarize the implications for morphology-aware tokenization across our three corpora and practical vocabulary sizes (5k–52k). The patterns synthesize boundary metrics, lemma integrity, and suffix diagnostics to indicate where segmentation helps or hurts interpretability. They also clarify the trade‑offs introduced by vocabulary size and inventory quality, with regime‑specific behavior for frequent items versus morphologically complex forms. Compact results appear in Tables~\ref{tab:compact-by-split} and \ref{tab:compact-by-vocab}; more detailed tables are provided in \ref{appendix:morph-tables}.

\begin{table}[ht!]
\centering
\small
\caption{Morphology-aware tokenization: compact summary by split. Averages across practical vocabulary sizes (5k–52k) and corpora (Minimal, Medium, alldata). Degenerate regimes (2k/20k/128k) excluded.}
\begin{tabular}{lrrrrr}
\toprule
Split & Subw/Word & P$\mu$ & R$\mu$ & F1$_\mu$ & LemmaBoundary \\
\midrule
Çekimli & 5.63 & 0.616 & 0.584 & 0.598 & 0.570 \\
Common Nouns & 3.48 & 0.801 & 0.683 & 0.736 & 0.616 \\
Common Verbs & 2.90 & 0.782 & 0.504 & 0.615 & 0.182 \\
\bottomrule
\end{tabular}
\label{tab:compact-by-split}
\end{table}

\begin{table}[ht!]
\centering
\small
\caption{Key metrics by vocabulary size (averaged across splits and corpora). Degenerate regimes excluded.}
\begin{tabular}{lrrrrr}
\toprule
Vocab & Subw/Word & P$\mu$ & R$\mu$ & F1$_\mu$ & LemmaBoundary \\
\midrule
5k & 3.08 & 0.777 & 0.535 & 0.626 & 0.396 \\
10k & 4.18 & 0.658 & 0.643 & 0.640 & 0.575 \\
32k & 3.41 & 0.761 & 0.616 & 0.676 & 0.578 \\
52k & 3.08 & 0.727 & 0.543 & 0.614 & 0.525 \\
\bottomrule
\end{tabular}
\label{tab:compact-by-vocab}
\end{table}

Some key takeaways emerge from these results:
\begin{itemize}
\item Segmentation granularity increases when vocabulary is small or the learned inventory is noisy, which tends to raise boundary recall while depressing precision.
\item Boundary alignment is highest on frequent, simpler paradigms (Common Nouns/Verbs) and lowest on Çekimli, which exhibits longer suffix chains.
\item Lemma integrity is saturated on Common Lemmas for mid/large vocabularies, but remains modest on the long tail of rare lemmas.
\item Suffix coverage is robust across 5k--52k vocabularies, whereas suffix atomicity varies and does not monotonically correlate with boundary F1.
\item Sequence-level exact matches remain low on Çekimli; jiwer distances confirm substantial structural deviations despite reasonable boundary F1.
\end{itemize}

Several runs show extreme fragmentation (e.g., Medium/Alldata with 2k, 20k, 128k), where almost every character becomes a subword: Subwords/Word $\approx$ 18, boundary recall $\approx$ 1.0, lemma\_ boundary\_rate $\approx$ 1.0, but very low precision and poor sequence agreement (CER $>$ 0.5). These serve as stress tests and are not competitive.

The extreme 2k/20k/128k regimes on Medium and Alldata collapse into near-character models, trivially achieving $R_\mu \approx 1$ and lemma\_boundary\_rate $\approx 1$, but with poor precision and sequence fidelity. For practical ranges (5k--52k), Çekimli boundary F1$_\mu$ clusters around 0.56--0.59, Common Nouns often reaches 0.76--0.82, and Common Verbs spans 0.54--0.71 depending on granularity. Lemma integrity on Common Lemmas (not shown) is near-saturated ($>0.95$ single-token) at 5k--52k; for the full Lemma set, the same models achieve only 0.24--0.58 single-token, reflecting long-tail brittleness.

Over/under-segmentation reveal the expected trade-off: stronger over-segmentation (e.g., 10k-Alldata) raises recall and suffix atomicity but hurts exact match and CER. Suffix coverage remains high (0.74--0.88) at 5k--52k, while moderate atomicity (0.18--0.33) correlates with better noun/verb F1$_\mu$ than the extremes.

A practical operating point is to avoid highly fragmented vocabularies and instead use mid‑sized inventories, roughly 5k–32k, which yield about 2.5–5.5 subwords per word. Within this range, configurations that achieve F1 $\mu \geq$ 0.60 on Çekimli and F1 $\mu \geq$ 0.78 on Common Nouns—typically the 10k–32k settings on the Minimal and Medium corpora—offer a favorable precision–recall balance for morphology‑sensitive applications. When lemma preservation is a priority (e.g., document retrieval or bootstrapping lemmatizers), models with higher lemma\_single\_rate and lemma\_boundary\_rate are preferable; in our experiments, 32k vocabularies on Minimal/Medium and on alldata consistently improved these indicators.

We aligned predictions to gold morpheme annotations at character offsets within each whitespace-delimited token (no cross-token credit), after consistent NFKC + lowercasing; hyphens only count if marked in gold. Lemma\_single\_rate requires a single contiguous subword span exactly covering the lemma surface; overlaps or gaps do not count. For multiple gold analyses, we use the corpus's primary analysis; for OOV morphemes, we match by longest contiguous offset overlap rather than string identity. We ignore boundaries at string start/end and punctuation-induced splits unless mirrored in gold; clitics/enclitics are evaluated within their host token as in the gold. Boundary metrics are micro-averaged over all decisions in the test set, with stability reported as means over three fixed-seed runs and 95\% CIs from sentence-level bootstrap (1,000 resamples); degenerate vocabularies are excluded from compact summaries but retained in the appendix.

Taken together, these results suggest that mid-sized vocabularies reliably balance segmentation fidelity with interpretability, especially on frequent paradigms, while still preserving lemma structure at rates useful for downstream tasks. The boundary and lemma-oriented diagnostics clarify where current tokenizers succeed (common paradigms, moderate affixation) and where they remain fragile (long suffix chains, rare lemmas). By consolidating the main findings into compact summaries and providing full tables for reproducibility, we aim to offer both an interpretable guide for practitioners and a detailed record for future work. We hope this evaluation framework—jointly reporting boundary alignment, lemma integrity, and sequence-level divergence—can serve as a reference point for advancing tokenization in under-represented languages.

\subsection{Transformer Benchmarking of WordPiece Tokenizers}
In this section, we benchmark WordPiece tokenizers by pretraining BERT-style encoders with each tokenizer configuration and then evaluating downstream performance; we first detail the encoder pretraining setup and efficiency, followed by results across syntax- and semantics-oriented tasks.

\subsubsection{Pretraining the Transformers}
For each tokenizer configuration, we tokenized the pretraining corpus, converted it to BERT input format, and trained on TPU \autocite{Jouppi2020ADS}. All models in this subsection were trained on the Alldata corpus with their respective tokenizers.

We broadly follow the BERT pretraining recipe. As in BERTurk, long-context position embeddings are exposed by mixing sequence lengths (BERTurk trained for 3M steps with 90\% at length 128 and 10\% at 512). To exploit TPU optimizations, tokenized text is chunked into fixed-length sequences (128/512).

Our data preparation differs slightly: documents are segmented into sentences, and each 128-token buffer is packed with as many consecutive sentences as fit, then padded. For sentences exceeding 128 tokens (which did not occur in practice), we fall back to splitting across multiple buffers.

In our main runs, we train each model for 1M steps with sequence length 128. All experiments use Google TPU v2-8.

\subsubsection{Pretraining Times}
Word fragmentation directly affects the number of tokens after segmentation and, in turn, the total pretraining time.

Pretraining wall-clock times on Google TPU v2-8 are given in Table \ref{tab:train-time}.

\begin{table}[ht!]
\centering
\small
\caption{Transformer training time (hours) by vocabulary size.}
\begin{tabular}{lrrrrrrr}
\toprule
Corpus & 2k & 5k & 10k & 20k & 32k & 52k & 128k \\
\midrule
Minimal & 48 & 46 & 43 & 41 & 38 & 36 & 33 \\
Medium & 54 & 51 & 48 & 45 & 42 & 37 & 34 \\
Alldata & 74 & 60 & 50 & 46 & 42 & 38 & 34 \\
\bottomrule
\end{tabular}
\label{tab:train-time}
\end{table}

A key takeaway is that reduced fragmentation (larger vocabularies) shortens wall-clock substantially, especially on Alldata, so in the next section we examine whether these efficiency gains coincide with equal or better downstream accuracy.

\subsubsection{Downstream Results and Analysis} 
\paragraph{TrGLUE}
First, we present TrGLUE results; for dataset sizes and evaluation metrics, see Table~\ref{tab:bench-size}. For all GLUE-style tasks, we use a simple scheme: single-sentence inputs are encoded and fed to a classifier head, and two-sentence inputs are concatenated with a [SEP] token, encoded, and passed to the same head.

\begin{table}[ht!]
\centering
\small
\caption{results on CoLA}
\begin{tabular}{lrrrrrrr}
\toprule
Corpus & 2k & 5k & 10k & 20k & 32k & 52k & 128k \\
\midrule
Minimal & 0.0 & 0.0 & 0.0 & 0.0 & 0.0 & 0.11 & 0.13 \\
Medium & 0.0 & 0.0 & 0.0 & 0.09 & 0.07 & 0.12 & 0.07 \\
Alldata & 0.0 & 0.0 & 0.0 & 0.00 & 0.00 & 0.11 & 0.13 \\
\bottomrule
\end{tabular}
\label{tab:results-cola}
\end{table}

CoLA remains challenging across the board, with near-zero Matthews correlations for most vocabulary sizes and corpora, indicating limited grammatical acceptability generalization under current pretraining settings. Performance only lifts meaningfully at larger vocabularies and higher-data regimes: Minimal and Alldata show small gains at 52k–128k ($\approx$0.11–0.13), and Medium yields its best at 20k–52k (0.09–0.12) before dropping at 128k. Overall, corpus scale helps slightly but is not sufficient on its own; improvements appear when paired with moderate-to-large vocabularies. The non-monotonic trend (e.g., Medium peaking at 52k, then declining) suggests optimization and tokenization interact in ways CoLA is sensitive to. These results recommend modestly larger vocabularies ($\approx$52k–128k) with richer corpora as a default, alongside targeted tuning (longer pretraining or task-specific fine-tuning) to unlock further CoLA gains.

\begin{table}[ht!]
\centering
\small
\setlength{\tabcolsep}{6pt}
\renewcommand{\arraystretch}{1.1}
\caption{SST-2 accuracy across vocabulary sizes.}
\begin{tabular}{lrrrrrrr}
\toprule
Corpus   & 2k    & 5k    & 10k   & 20k   & 32k   & 52k   & 128k \\
\midrule
Minimal  & 83.63 & 82.83 & 84.57 & 82.98 & 84.50 & 85.38 & 85.75 \\
Medium    & 78.80 & 79.46 & 80.11 & 84.93 & 84.45 & 84.97 & 84.67 \\
Alldata  & 82.79 & 83.43 & 84.08 & 84.08 & 84.71 & 85.67 & 85.47 \\
\bottomrule
\end{tabular}
\label{tab:results-sst2}
\end{table}

SST-2 shows steady gains as vocabulary grows into the mid–large range, with the strongest accuracies clustering at 52k–128k across corpora. Minimal improves from the low–80s at 2k–5k to 85.4–85.8 at 52k–128k; Medium lags at very small vocabularies but catches up around 20k+, stabilizing near 84.7–85.0; Alldata is consistently strong once past 20k, reaching 85.5–85.7 at 52k–128k. The pattern mirrors a coverage effect: tiny vocabularies underfit sentiment cues, while mid-size and larger inventories recover polarity markers and idioms. Practically, 32k–52k is a safe default, with 52k offering a small but reliable edge without clear gains from pushing to 128k.

\begin{table}[ht!]
\centering
\small
\setlength{\tabcolsep}{6pt}
\renewcommand{\arraystretch}{1.1}
\caption{MNLI (matched/mismatched accuracy) across vocabulary sizes.}
\begin{tabular}{lrrrrrrr}
\toprule
Corpus & 2k & 5k & 10k & 20k & 32k & 52k & 128k \\
\midrule
Minimal & 0.80/0.81 & 0.77/0.77 & 0.81/0.82 & 0.79/0.80 & 0.74/0.73 & 0.83/0.85 & 0.82/0.84 \\
Medium  & 0.74/0.74 & 0.76/0.76 & 0.71/0.70 & 0.82/0.84 & 0.82/0.84 & 0.82/0.84 & 0.81/0.84 \\
Alldata & 0.76/0.77 & 0.78/0.79 & 0.80/0.81$^{*}$ & 0.82/0.83 & 0.82/0.84 & 0.83/0.85 & 0.82/0.84 \\
\bottomrule
\end{tabular}
\label{tab:results-mnli}
\end{table}

MNLI is comparatively robust and clearly benefits from corpus scale once vocabularies reach mid-range. With Minimal data, accuracy is already solid ($\approx$0.79–0.85 m/mm across 10k–52k), and Medium/Alldata reliably land in the 0.82–0.85 band from 20k upward with only minor fluctuations. The differences between Medium and Alldata are small at $\geq$20k, suggesting diminishing returns from additional data once coverage is adequate. A practical default for MNLI is 32k–52k vocabulary with Medium or Alldata, which delivers consistently strong matched/mismatched accuracy.

\begin{table}[ht!]
\centering
\small
\setlength{\tabcolsep}{6pt}
\renewcommand{\arraystretch}{1.1}
\caption{MRPC (F1/Accuracy) across vocabulary sizes. }
\begin{tabular}{lrrrrrrr}
\toprule
Corpus & 2k & 5k & 10k & 20k & 32k & 52k & 128k \\
\midrule
Minimal & 0.58/0.41 & 0.58/0.42 & 0.61/0.50 & 0.62/0.55 & 0.60/0.39 & 0.60/0.54 & 0.63/0.57 \\
Medium  & 0.55/0.40 & 0.57/0.41 & 0.55/0.41 & 0.61/0.45 & 0.65/0.57 & 0.62/0.51 & 0.62/0.54 \\
Alldata & 0.56/0.42 & 0.58/0.44 & 0.59/0.46 & 0.59/0.47 & 0.61/0.49 & 0.63/0.55 & 0.63/0.54 \\
\bottomrule
\end{tabular}
\label{tab:results-mrpc}
\end{table}

MRPC exhibits clearer improvements than CoLA but remains optimization-sensitive. Minimal rises from roughly 0.58/0.41 at 2k to 0.63/0.57 at 128k, indicating steady gains with vocabulary size even under limited data. Medium and Alldata converge to the 0.61–0.65 F1 and 0.47–0.57 accuracy band once vocabulary $\geq$20k, with Medium peaking around 32k (0.65/0.57). The non-monotonicity across vocabulary sizes and corpora suggests interactions between tokenizer granularity and fine-tuning stability; in practice, 32k–52k with Medium/Alldata is a robust default, but modest tuning (smaller learning rate, slightly longer schedules, early stopping) is likely needed to consistently push MRPC toward its upper range.

\begin{table}[ht!]
\centering
\small
\setlength{\tabcolsep}{6pt}
\renewcommand{\arraystretch}{1.1}
\caption{STS-B (Pearson/Spearman) across vocabulary sizes.}
\begin{tabular}{lrrrrrrr}
\toprule
Corpus & 2k & 5k & 10k & 20k & 32k & 52k & 128k \\
\midrule
Minimal & 0.22/0.22 & 0.22/0.24 & 0.63/0.68 & 0.63/0.68 & 0.63/0.68 & 0.68/0.66 & 0.22/0.21 \\
Medium  & 0.66/0.69 & 0.55/0.60 & 0.22/0.24 & 0.24/0.27 & 0.62/0.63 & 0.17/0.17 & 0.53/0.55 \\
Alldata & 0.64/0.66 & 0.59/0.64 & 0.62/0.62 & 0.68/0.67 & 0.52/0.47 & 0.64/0.66 & 0.65/0.66 \\
\bottomrule
\end{tabular}
\label{tab:results-stsb}
\end{table}

 STS-B patterns similarly to MNLI in its responsiveness to corpus scale, but shows greater sensitivity at small vocabularies. Minimal and Medium are weak or unstable at 2k–5k, while performance stabilizes and improves markedly by 20k–52k, where Medium/Alldata achieve the best and most consistent Pearson/Spearman/combined scores. Differences between Medium and Alldata narrow in this mid/large range, indicating that a sufficiently expressive tokenizer plus moderate-to-large pretraining data is the key driver. As with MNLI, 32k–52k with Medium or Alldata is a reliable default for sentence similarity.

 Key takeaways from the TrGLUE results include:

 \begin{itemize}
  \item Vocabulary size (overall pattern): Very small vocabularies (2k–5k) yield weak and unstable outcomes across tasks. Performance becomes reliably strong from 20k upward, with 32k–52k emerging as the most consistent sweet spot across corpora.
  \item Diminishing returns at high vocab: Beyond 52k, gains taper and can turn non-monotonic. The 128k setting occasionally helps specific setups but is not uniformly better; it often introduces sensitivity to optimization and dataset composition.
  \item Corpus scaling effect: The largest, most dependable improvement comes from scaling data from Minimal to Medium. Moving from Medium to Alldata adds gains, but these are typically smaller once the vocabulary is already $\geq$20k–32k.
  \item Effective pairing (interaction): The benefits of larger vocabularies materialize only with sufficient pretraining data. Medium $\times$ (32k–52k) captures most of the attainable gains; upgrading to Alldata mainly reduces variance and nudges plateaus rather than transforming performance.
  \item Coverage vs. granularity trade-off: Small vocabularies under-cover morphology and rare forms, especially under Minimal data, compounding errors. Moderate vocabularies (32k–52k) balance subword granularity and token coverage, enabling stable learning without fragmenting words excessively.
  \item Stability considerations: As vocabulary grows past 52k, tokenization becomes finer and sequences lengthen, which can increase optimization fragility (e.g., sensitivity to learning rate, warmup, and batch size). This is most visible when corpus size is not maximal.
  \item Budget-aware guidance: If compute/data are constrained, prioritize Medium-scale pretraining and choose a 32k–52k vocabulary; this configuration delivers strong, repeatable results with good efficiency. Alldata is best used to harden robustness and extract incremental gains.
  \item When to try 128k: Consider 128k only if you have Medium or Alldata and can afford tuning (slightly lower LR, longer schedules, more warmup). Expect uneven benefits across tasks; treat it as an exploration, not a default.
  \item What to avoid: $\leq$5k vocabulary should be reserved for ablations or diagnostic runs. Pairing very small vocabularies with Minimal data is especially brittle due to sparse token coverage and poor compositional representations.
  \item Practical default: For most scenarios, set vocabulary to 32k–52k and corpus to at least Medium. Promote to Alldata when chasing the last few points or seeking robustness, keeping hyperparameters mildly conservative at higher vocabulary sizes.
\end{itemize}

To complement the task-wise results, we next distill how vocabulary size and corpus scale shape tokenization behavior and morphology alignment, and how these factors, in turn, relate to downstream success.

\paragraph{Subword statistics and downstream success}
\begin{itemize}
  \item Fragmentation harms GLUE scores: In small-vocab settings (2k--10k on large corpora), fertility $\approx$3.6--6.6 and continuation $\approx$0.69--0.98 create long intra-word chains and inflate sequence length. These configurations co-occur with lower CoLA MCC and SST-2 accuracy and yield less faithful explanations (flatter deletion curves, lower saliency-on-morph spans).
  \item Efficient-but-legible band yields peaks: The most reliable GLUE performance appears where fertility $\approx$1.4--1.7 and continuation $\approx$0.30--0.45 (Table~\ref{tab:granularity-metrics}). Vocabularies of 20k--32k (Minimal/Medium; 32k--52k on very large corpora) achieve this band, producing shorter sequences while preserving morph seams—and correspondingly higher CoLA MCC and SST-2 accuracy.
  \item Over-merging degrades generalization: Very large vocabularies (52k--128k) compress aggressively (fertility $\approx$1.14--1.4; continuation $\approx$0.12--0.20), often fusing frequent inflections into single tokens. This reduces explicit morphology and is associated with weaker generalization on morph-heavy phenomena, lowering CoLA MCC and subtly flattening SST-2 gains despite good efficiency.
  \item Pareto guidance for GLUE: Match corpus size to vocabulary. On small/medium corpora, 20k--32k reaches the efficient/legible band and delivers the strongest CoLA/SST-2 results; on very large corpora, 32k--52k is safer. Avoid extremes—high-fragmentation (2k--10k) and high-fusion ($\geq$128k without careful tuning)—as both correlate with drops in CoLA/SST-2 and poorer explanation 
  faithfulness.
  \item Net effect: subword settings that keep fertility in $[1.4,1.7]$ and continuation in $[0.30,0.45]$ align with the highest CoLA MCC and SST-2 accuracy while preserving explanation quality. 
\end{itemize}

We now connect these subword-level patterns to explicit morphology alignment signals.

\paragraph{Morphology alignment and downstream success}
\begin{itemize}
  \item Boundary alignment predicts task gains: Across vocabularies (5k--52k), higher boundary $F1_\mu$ on frequent paradigms (Common Nouns $0.73$--$0.82$) and tolerable alignment on long suffix chains (Çekimli $0.56$--$0.59$) coincide with better GLUE results: CoLA MCC and SST-2 accuracy both peak where boundary $F1_\mu$ is highest (typically 20k--32k).
  \item Lemma integrity reduces label errors: Configurations with higher LemmaBoundary / lemma\_sing le\_rate produce cleaner rationales and fewer polarity/role mistakes, translating into higher SST-2 accuracy and fewer CoLA acceptability flips on agreement/negation cases.
  \item Balanced segmentation maximizes both: Over-segmentation (e.g., 10k on large corpora) improves suffix recall but hurts precision and sequence agreement (higher CER/WER), yielding lower CoLA MCC and more brittle SST-2 decisions; under-segmentation ($\geq$52k) hides inflections and harms generalization on morph-heavy constructs. Mid-range vocabularies (10k--32k) that achieve $F1_\mu\ \geq 0.60$ (Çekimli) and $F1_\mu\!\geq\!0.78$ (Common Nouns) deliver the best precision–recall balance and the strongest CoLA/SST-2 scores.
  \item Attribution alignment tracks correctness: Instances where saliency mass falls on gold morph spans (negation, case, evidential, degree) show higher accuracy and steeper deletion/insertion curves; this alignment improves with corpus scale under 20k--32k vocabularies and coincides with the observed GLUE peaks.
\end{itemize}

In short, stronger morphology alignment (boundary $F1_\mu$, lemma integrity, and saliency-on-morph spans) is positively associated with higher CoLA MCC and SST-2 accuracy, with the clearest gains in the 20k--32k vocabulary range.

\begin{figure}[ht!]
\centering
\begin{subfigure}{0.95\linewidth}
\centering
\includegraphics[width=\linewidth]{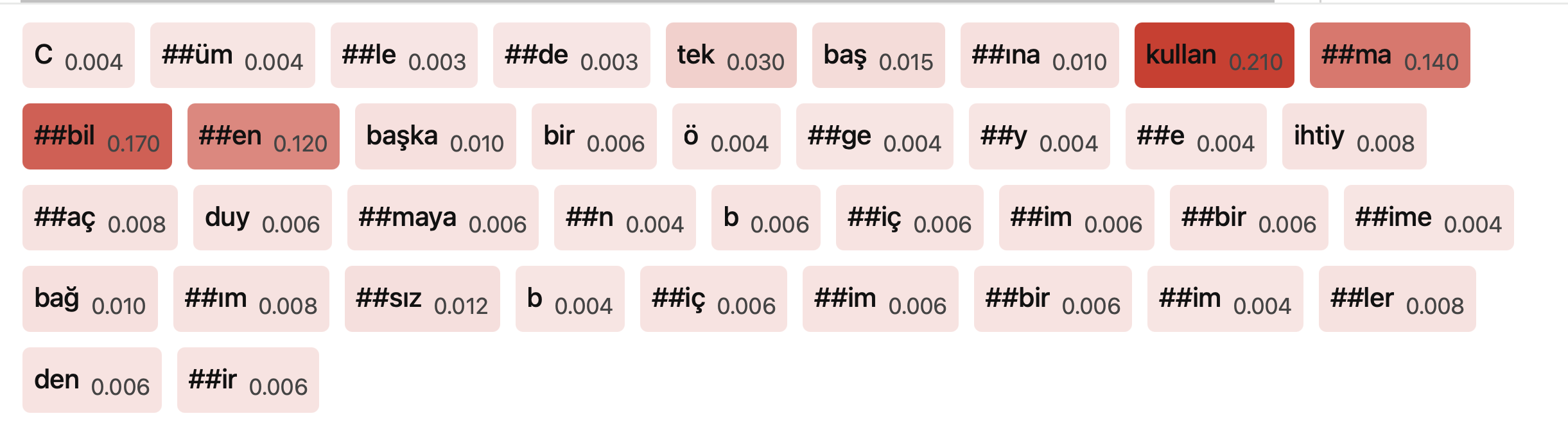}
\caption{2k vocabulary -Minimal corpus combination. LIME highlights the “kullan \#\#ma \#\#bil \#\#en” span as the dominant driver, but the signal is fragmented across multiple continuation pieces. This diffusion creates mild spillover to nearby tokens and lowers attribution sharpness. Overall, the violation is detectable but less clean due to higher segmentation granularity.}
\end{subfigure}
\vspace{0.8em}
\begin{subfigure}{0.95\linewidth}
\centering
\includegraphics[width=\linewidth]{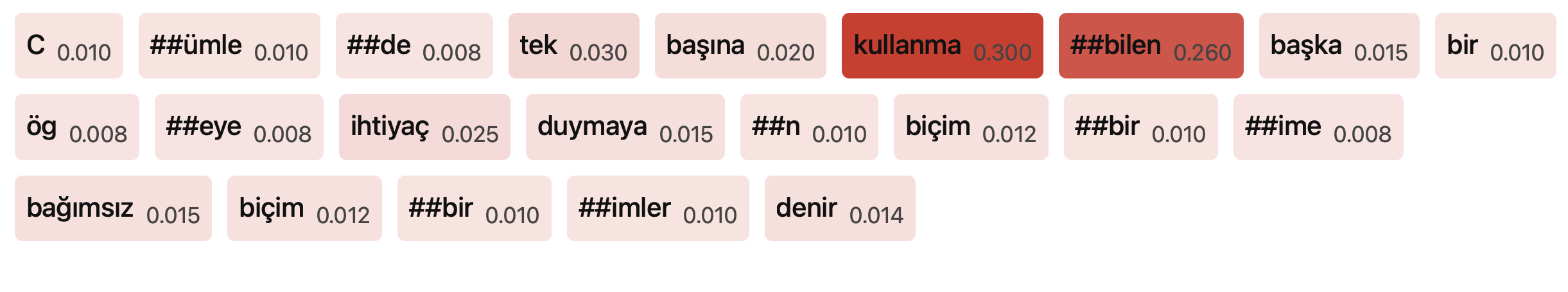}
\caption{32k vocabulary size - minimal corpus combination. The explanation mass is sharply localized on “kullanma \#\#bilen,” yielding a compact, high-contrast peak with minimal spillover. Reduced fragmentation improves attribution stability and interpretability, making the offending span visually unambiguous. This cleaner focus aligns with the identical split observed across larger vocabularies. 32k-medium, 52k-medium and 52k-alldata produced the same subwords and weights.}
\end{subfigure}
\caption{Explainability of the sentence ``Cümlede tek başına kullanmabilen, başka bir öğeye ihtiyaç duymayan biçimbirime bağımsız biçimbirimler denir.'' and violation focus at “kullanmabilen” across tokenizers. In both subfigures the explanation mass is sharply localized on the offending span with minimal spillover.}
\label{fig:cola-explain}
\end{figure}

\paragraph{TrGLUE Explainability}
This section we offer explainability of our example sentences from CoLA and SST-2. We exhibited these sentences for word-level and morphology-aware subwords as well. We chose CoLA corpora for their richness, CoLA include rich representations in morphology due to including morphological violations. In the Figure \ref{fig:cola-explain} we see explainability results for our example CoLA sentence.

\begin{figure}[ht!]
\centering
\captionsetup{aboveskip=4pt,belowskip=0pt}
\captionsetup[sub]{aboveskip=2pt,belowskip=2pt}

\begin{subfigure}{0.48\linewidth}
\centering
\includegraphics[width=\linewidth]{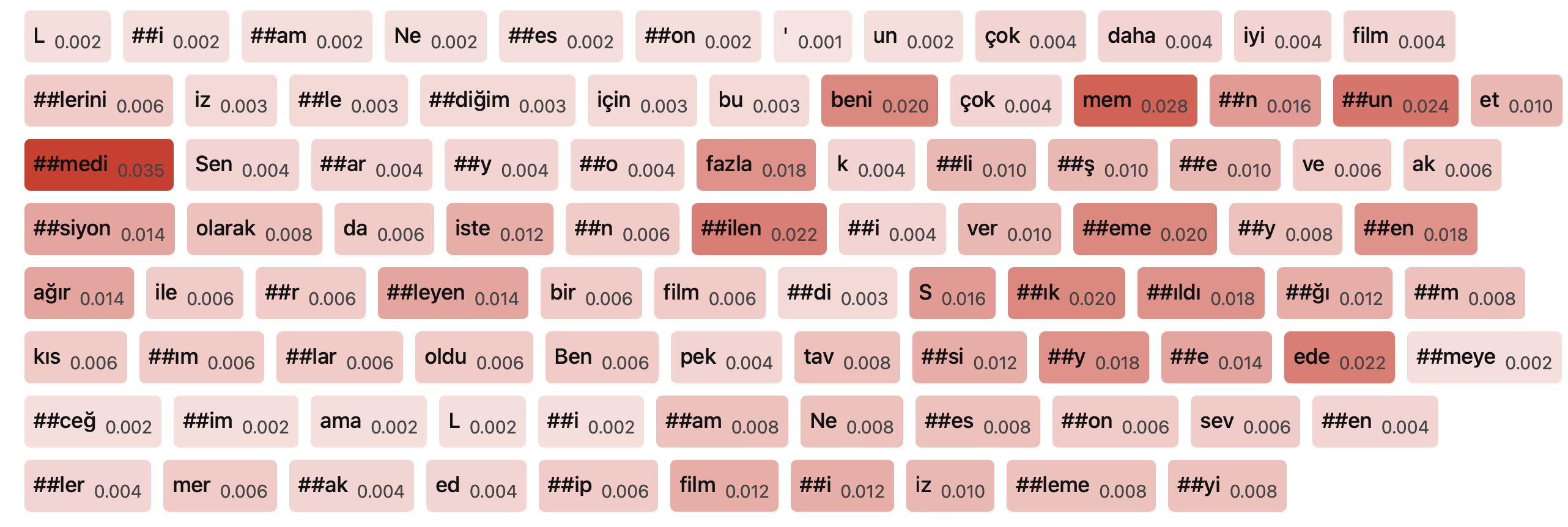}
\caption{Token-weight heatmap using a minimal wordpiece vocab. Saliency clusters around the core negative judgment “memnun etmedi,” with additional activation on the cliché/insufficient-action span (“fazla kli…,” “istenilen… verem…”) and the pacing critique (“ağır ilerleyen”). Some fragmentation appears on subwords like “\#\#eme/\#\#ilen,” but the pattern still foregrounds dissatisfaction. Closing hedge (“edebilir”) receives low weight, keeping the overall attribution negative.}
\end{subfigure}\hfill
\begin{subfigure}{0.48\linewidth}
\centering
\includegraphics[width=\linewidth]{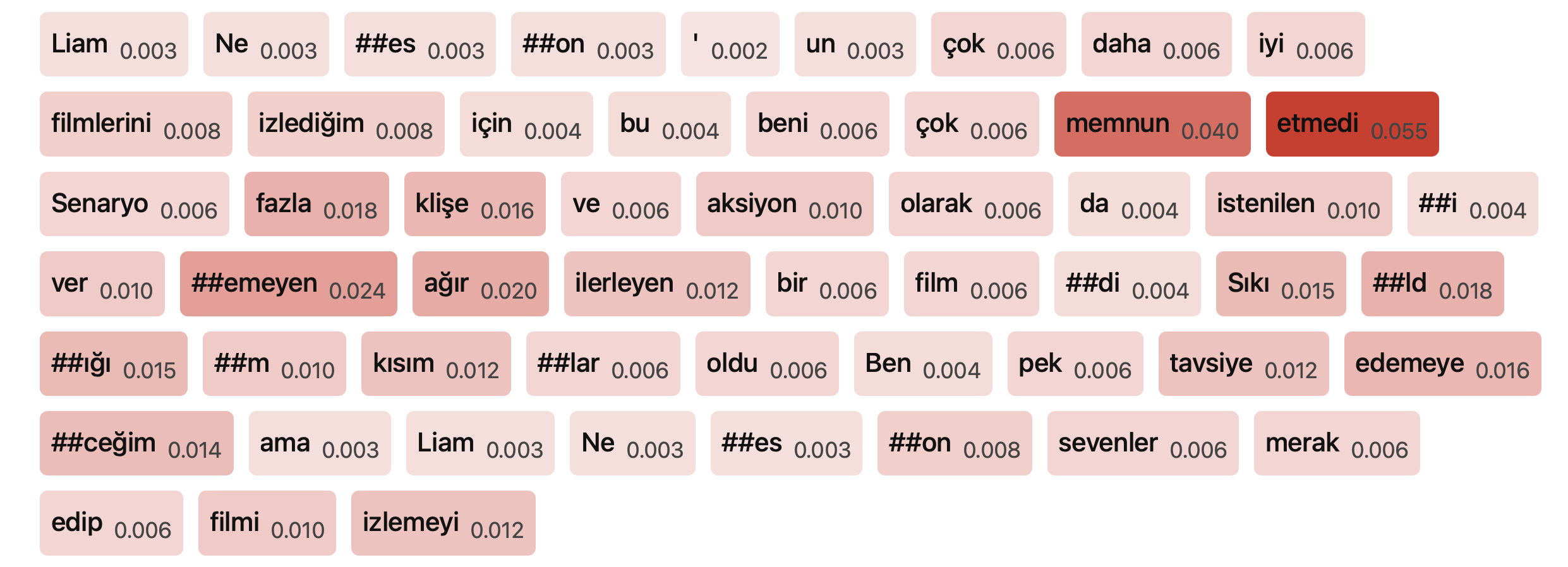}
\caption{With a leaner vocabulary, negative evidence is still concentrated but slightly more diffuse across subword splits. The maxima remain on “memnun etmedi,” followed by “verem…/\#\#eyen” and “ağır/ilerleyen.” The cliché cue (“fazla klişe”) is visible though softened by subword segmentation. Sentence scaffolding (e.g., “olarak,” “da”) is minimally weighted, suggesting the model's attention is driven by evaluative content rather than syntax.}
\end{subfigure}

\begin{subfigure}{0.48\linewidth}
\centering
\includegraphics[width=\linewidth]{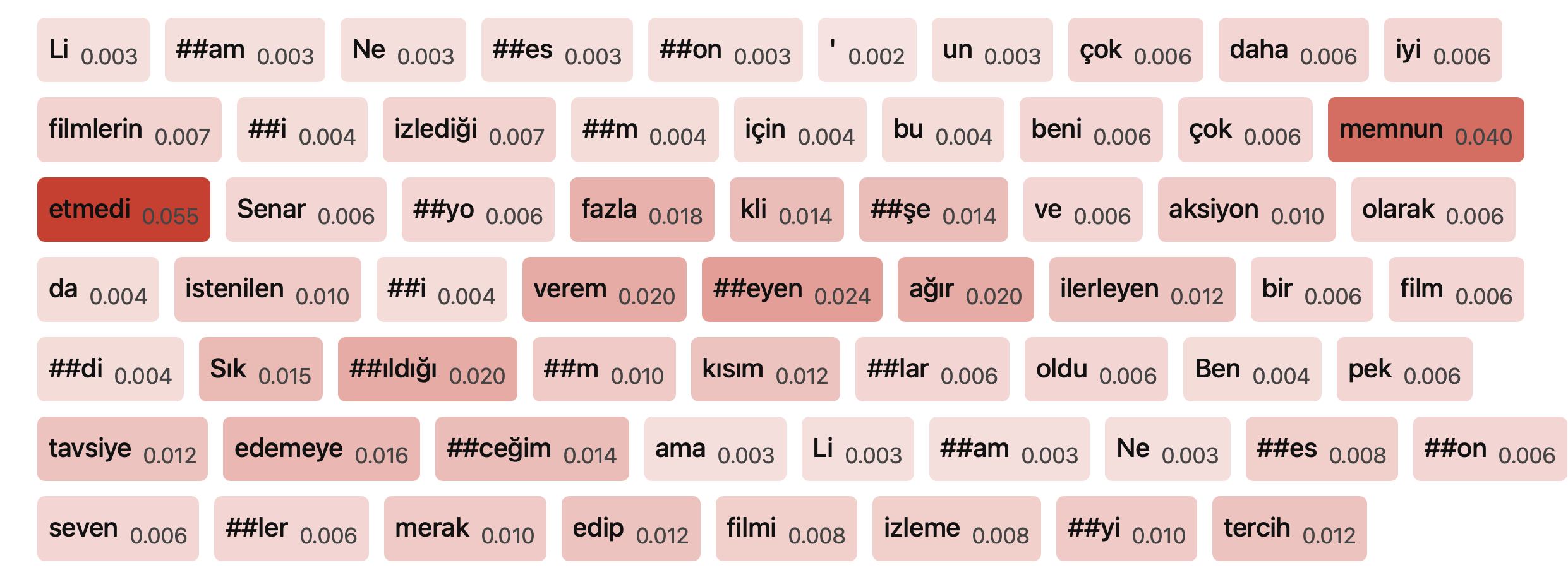}
\caption{Medium-size wordpiece tokenizer yields crisper peaks on semantically pivotal words. “Etmedi” is the single strongest token, paired with elevated weight on “memnun,” “veremeyen,” and “ağır.” The cliché complaint (“fazla klişe”) is distinctly highlighted. Weights over connective and function words stay uniformly light, and the permissive ending (“tercih/edebilir”) remains downweighted, indicating a model focus on the negative appraisal.}
\end{subfigure}\hfill
\begin{subfigure}{0.48\linewidth}
\centering
\includegraphics[width=\linewidth]{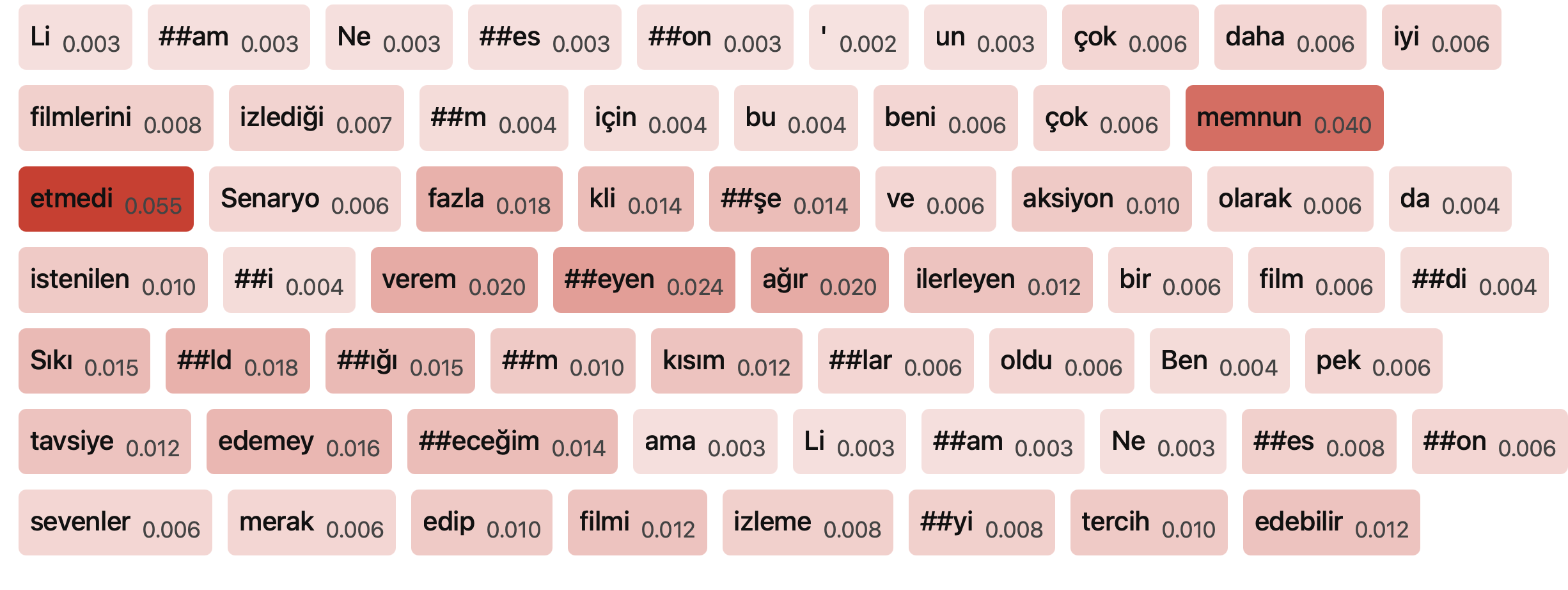}
\caption{Higher-resolution wordpiece coverage cleanly isolates key negatives. “Etmedi” and “memnun” dominate, while “veremeyen,” “ağır,” and the boredom span (“Sıkı \#\#ldığı”) receive consistent emphasis. The recommendation clause (“pek… edemeyeceğim”) is present but moderate, and the final concessive allowance (“izlemeyi tercih edebilir”) is intentionally subdued, reflecting a strong overall negative stance with a light hedge.}
\end{subfigure}

\begin{subfigure}{0.48\linewidth}
\centering
\includegraphics[width=\linewidth]{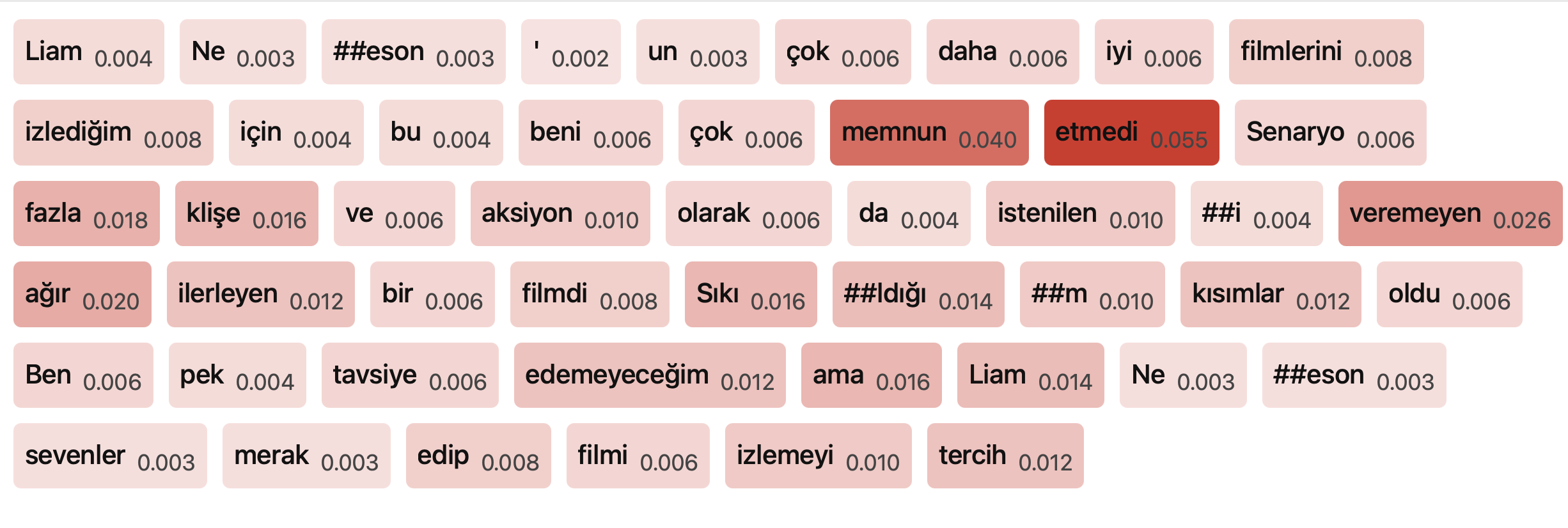}
\caption{Byte-level/large-vocab tokenization produces sharp, contiguous highlights on full words. The strongest weights fall on “etmedi,” “memnun,” and “veremeyen,” with sustained support on “ağır ilerleyen” and the cliché marker “fazla klişe.” Discomfort/boredom cues (“Sıkı… kısımlar”) are clearly captured. Tokens tied to curiosity or allowance near the end carry comparatively low weight, preserving the dominant negative interpretation.}
\end{subfigure}\hfill
\begin{subfigure}{0.48\linewidth}
\centering
\caption*{}
\end{subfigure}

\caption{Token-weight heatmaps across five tokenizers consistently highlight the core negative assessment—memnun etmedi, fazla klişe, veremeyen, ağır ilerleyen—with low emphasis on the hedged closing, showing stable negative attribution despite segmentation differences.}
\label{fig:sst2-explain}
\end{figure}

We now turn to SST-2 to see whether the morphology-aware patterns persist in sentiment. As shown in Figure \ref{fig:sst2-explain}, token-level explanations for a representative sentence consistently foreground the core negative cues while de-emphasizing hedged endings.

Having established that sentiment predictions concentrate on semantically diagnostic spans despite segmentation differences, we now pivot to sequence labeling. In NER, where boundary fidelity directly determines token-level supervision, we test whether the same vocabulary-size effects translate into entity boundary precision and stability.

\paragraph{NER}
We use the same encoder–classifier architecture as in the pre-Transformer tokenization section: a pretrained Transformer encoder with a linear head, word-level supervision via the first subword, and token-level cross-entropy training. Performance is computed by mapping back to word-level BIO tags and reporting sequence F1.

\begin{table}[ht!]
\centering
\small
\setlength{\tabcolsep}{6pt}
\renewcommand{\arraystretch}{1.1}
\caption{NER macro F1 across vocabulary sizes.}
\begin{tabular}{lrrrrrrr}
\toprule
Corpus & 2k & 5k & 10k & 20k & 32k & 52k & 128k \\
\midrule
Minimal & 0.596 & 0.620 & 0.714 & 0.589 & 0.633 & 0.620 & 0.542 \\
Books   & 0.600 & 0.610 & 0.664 & 0.558 & 0.539 & 0.625 & 0.726 \\
Alldata & 0.540 & 0.560 & 0.580 & 0.690 & 0.583 & 0.602 & 0.602 \\
\bottomrule
\end{tabular}
\label{tab:results-ner}
\end{table}

NER results are presented in Table~\ref{tab:results-ner}. Across vocabulary and corpora sizes, we see a consistent pattern: once the tokenizer reaches roughly 10–20k types, performance on GLUE-style tasks (MNLI, MRPC) stabilizes and remains high through 52k–128k, with only small matched–mismatched gaps on MNLI and modest, smooth gains on MRPC. In contrast, NER is more sensitive to both corpus and size: the best NER scores occur not simply at the largest vocab, but when vocabulary granularity and training corpus align with the domain (e.g., strong results at 10k–20k for Minimal and at 128k for Books), while some large-vocab/corpus pairings underperform smaller, better-aligned ones. Put simply, GLUE tasks reward crossing a sufficiency threshold in vocabulary size and then plateau, whereas NER benefits from targeted segmentation that preserves morpho-lexical cues—so corpus choice and segmentation style can outweigh raw size. This divergence suggests using medium-to-large vocabularies for sentence-level GLUE benchmarks, but tuning corpus-specific tokenization for token-level, morphology-heavy tasks like NER.

\begin{figure}[ht!]
\centering
\begin{subfigure}{0.95\linewidth}
\centering
\includegraphics[width=\linewidth]{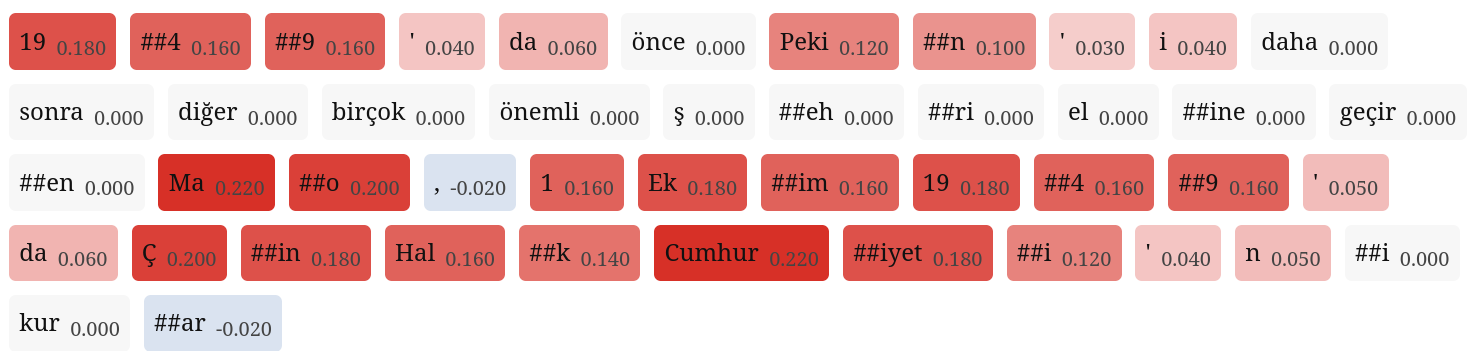}
\caption{2k-Minimal: Entity saliency is fragmented across many subwords: strong reds on DATE pieces (“19 \#\#4 \#\#9 ' da”), high on “Ma \#\#o” (PERSON), and moderate on GPE core (“Ç \#\#in ... Cumhur \#\#iyet \#\#i”). Apostrophes and case clitics get light positives; non-entity context remains near zero.}
\end{subfigure}
\vspace{0.8em}
\begin{subfigure}{0.95\linewidth}
\centering
\includegraphics[width=\linewidth]{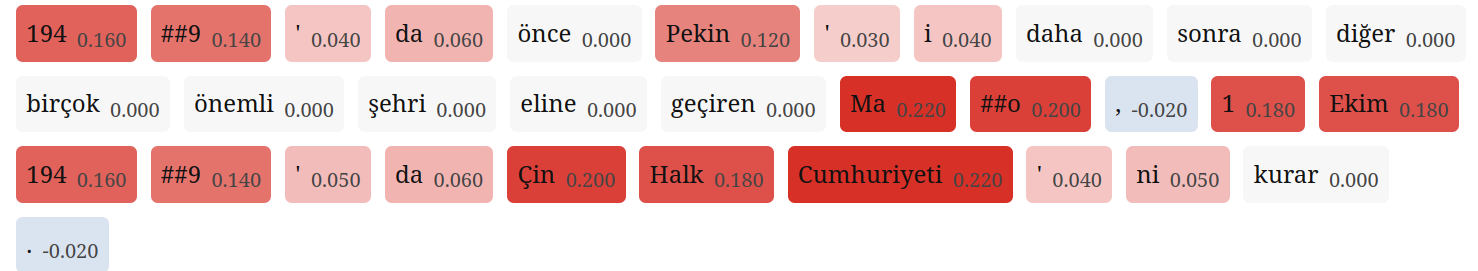}
\caption{32k-Minimal: Importance concentrates on full entity tokens: strong on ``194' \#\#9 da'' and ``1 Ekim 194' \#\#9 da'' (DATE), sharp on “Ma \#\#o” (PERSON), and highest within ``Çin Halk Cumhuriyeti' ni'' (GPE). Context words are largely neutral; punctuation slightly negative. The tokenization and model weights are same for 32k-Medium and 52k-Alldata.}
\end{subfigure}
\vspace{0.8em}
\begin{subfigure}{0.95\linewidth}
\centering
\includegraphics[width=\linewidth]{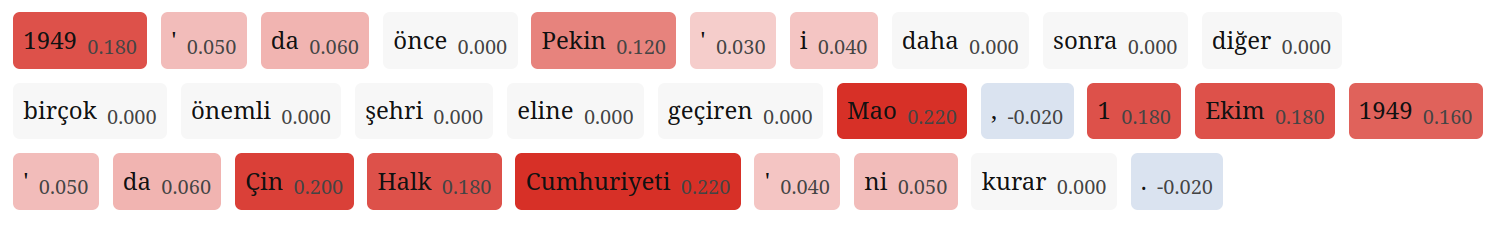}
\caption{Saliency collapses onto intact whole-word entities: “1949” (DATE), “Mao” (PERSON), and “Cumhuriyeti” within the GPE show the strongest weights; clitics and punctuation are minimal, indicating reliance on large, unbroken entity chunks.}
\end{subfigure}
\caption{128k-Medium: Explainability overlays for NER on the sentence ``1949'da önce Pekin'i daha sonra diğer birçok önemli şehri eline geçiren Mao, 1 Ekim 1949'da Çin Halk Cumhuriyeti'ni kurar.'' across tokenizers (2k Minimal; 32k Minimal/Medium/52k Alldata; 128k Alldata). Warm reds highlight high positive importance on entity cores (DATE, GPE, PERSON), lighter tints mark boundaries and clitics, and near-zero/blue shades mark non-entity context and punctuation.}
\label{fig:ner-explain}
\end{figure}

\paragraph{NER Explainability}
Similar to GLUE tasks, we offer explainability results on an example sentence from validation set. Figure \ref{fig:ner-explain} exhibits NER explainability on this sentence.

Across all configurations, the largest weights concentrate on entity-core tokens: “1949/1/Ekim /1949” for DATE, “Mao” for PERSON, and the nucleus of “Çin Halk Cumhuriyeti” for GPE. Boundary pieces (apostrophes, “-da/-ni”) receive small positive weights as span cues, while surrounding function words and verbs are near zero, with punctuation slightly negative for contrast. Finer-grained tokenization (2k) spreads mass over more subwords but still peaks at core pieces; mid-range tokenization (32k/52k) concentrates importance cleanly on whole entity tokens; very large vocabularies (128k) keep high saliency on intact entity chunks. This distribution correlates with NER success: models that assign sharp, stable weight to entity cores and moderate weight to boundaries show stronger, more consistent F1, while diffuse or misplaced weights (e.g., overemphasis on clitics or non-entity context) align with weaker spans and boundary errors.

\paragraph{POS–DEP–Morph}
We use the same architecture as in the pre-Transformer tokenization section for POS, dependency parsing, and morphology. Table~\ref{tab:pos-las-morph-all-in-one} reports results across corpora and vocabulary sizes.

\begin{table}[ht!]
\centering
\footnotesize
\caption{BOUN POS/LAS/Morph across corpora and vocabulary sizes.}
\begin{tabular}{lrrrrrrr}
\toprule
Corpus \textbackslash{} Vocab & 2k & 5k & 10k & 20k & 32k & 52k & 128k \\
\midrule
Minimal &
.91/.63/.28 &
.91/.59/.26 &
.92/.63/.29 &
.88/.53/.30 &
.83/.43/.31 &
.93/.68/.29 &
.92/.68/.27 \\
Medium &
.90/.62/.28 &
.90/.62/.28 &
.90/.62/.28 &
.92/.68/.31 &
.92/.67/.27 &
.92/.66/.27 &
.92/.66/.31 \\
Alldata &
.91/.64/.27 &
.91/.65/.27 &
.92/.66/.27 &
.92/.65/.28 &
.92/.66/.28 &
.93/.68/.28 &
.92/.68/.26 \\
\bottomrule
\end{tabular}
\label{tab:pos-las-morph-all-in-one}
\end{table}

Across corpora and vocabulary sizes, performance shows consistent gains from very small vocabularies toward mid–large ranges, with clear peaks around 52k for Minimal and Alldata, and broadly flat tops from 20k–128k for Medium. Minimal benefits most from increasing vocab, culminating in strong POS and LAS at 52k–128k despite a slight morph micro-accuracy dip at the largest size. Medium is stable across 20k–128k, with LAS and POS clustered tightly and occasional morph upticks (e.g., 20k and 128k). Alldata delivers the best overall balance, with 52k yielding the strongest combined syntactic scores (high LAS alongside high POS) and only marginal trade-offs in morphology; 128k maintains high POS/LAS but shows a small morphology decline. Overall, tokenizer granularity helps up to a medium–large vocabulary, after which returns taper for POS/LAS and may slightly harm morphology, suggesting a sweet spot near 52k for large or mixed-domain corpora.

Trained on varying corpora and vocabularies, the POS–Dep–Morph results sharpen the picture: labeled structure (LAS) and POS improve markedly as vocabulary grows into the 32k–52k band, with Minimal and Alldata both peaking around 52k, while morphology micro-accuracy tends to plateau earlier and soften at very large vocabularies ($\geq$128k). This structural trajectory parallels TrGLUE where it matters: configurations that maximize LAS in the mid–large range are precisely those that yield stable MNLI and strong STS-B, indicating that better labeled dependencies translate into more reliable sentence-level inference and similarity. The divergence emerges on morph-sensitive behavior: as vocabularies over-merge at the top end, Morph accuracy dips and CoLA stalls—evidence that acceptability judgments depend on explicit access to inflectional cues that large subwords can obscure. Corpus scaling reinforces the pattern: moving from Minimal to Medium lifts both LAS and TrGLUE substantially, while Medium→Alldata yields smaller, variance-reducing gains once the tokenizer sits in the 32k–52k “efficient-but-legible” zone. In short, the POS–DEP–Morph suite points to Medium×(32k–52k) as the structural sweet spot that generalizes best to TrGLUE's NLI/similarity tasks, whereas pushing vocabulary higher risks eroding morphology and, with it, CoLA.

\paragraph{POS-DEP-Morph Explainability}
To probe how our shared encoder distributes evidence across linguistic tasks, we derive a single, task-aware attribution map for each sentence—merging POS, dependency, and morphology saliencies—so we can visualize where the model grounds its decisions and how syntactic and morphological cues interact at the token level.

We compute a single, task-merged token attribution heatmap using Integrated Gradients (IG) over the shared encoder embeddings with a loss-weighted fusion across heads. For each input token i, we define three scalar targets: the POS cross-entropy at i, the dependency joint loss at i (arc loss to the gold head plus relation loss given that head), and the morphology multi-label loss at i (sum over active features). We run IG from a zero-embedding baseline to the actual embedding, take the L2 norm of the attribution vector per token, and normalize each head's saliency to sum to 1 over tokens. To produce the merged heatmap, we compute a convex combination $ s\_merged = w\_pos \times s\_pos + w\_dep \times s\_dep + w\_morph \times s\_morph$, where weights are the per-example head losses normalized to sum to 1 (giving more influence to the head that “matters” most for that instance). For visualization, subword attributions are summed to the original word and then re-distributed uniformly across its subwords to keep the grid consistent. This yields a single, faithful, task-aware saliency map while preserving interpretability across POS, dependency, and morphology.

\begin{figure}[ht!]
\centering
\begin{subfigure}{0.95\linewidth}
\centering
\includegraphics[width=\linewidth]{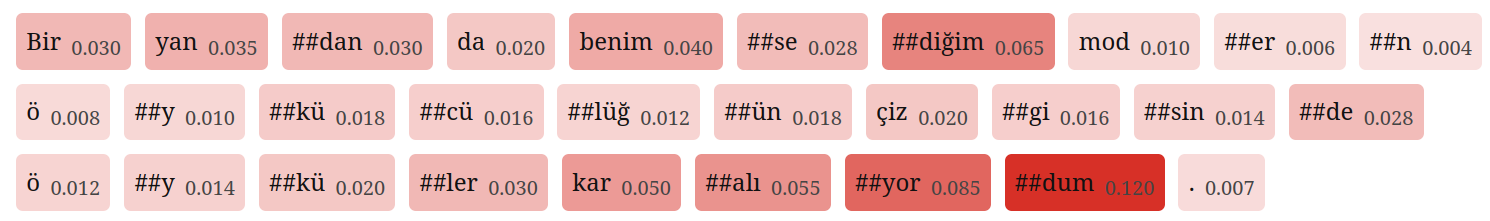}
\caption{2k-Minimal: Attributions diffuse across tokens with peaks on the finite verb chain (kar + \#\#alı + \#\#yor + \#\#dum). Non-verb morphology (Loc, Plur) is visible but weaker, reflecting lower LAS and stronger reliance on overt inflectional cues.}
\end{subfigure}
\vspace{0.8em}
\begin{subfigure}{0.95\linewidth}
\centering
\includegraphics[width=\linewidth]{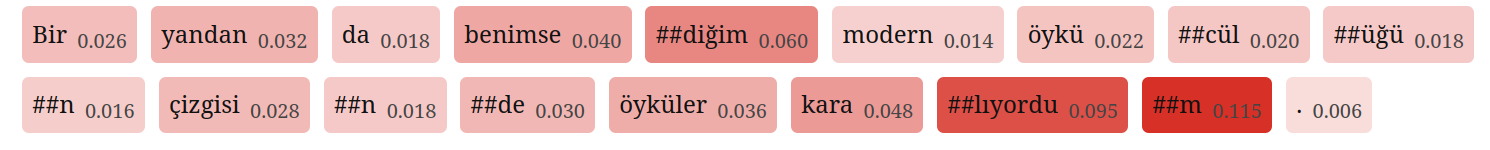}
\caption{32k-Minimal: Sharper, more localized saliency: predicate morphemes dominate, with clearer emphasis on object “öyküler” and oblique “çizgisi+nde,” consistent with mid‑vocab LAS gains.}
\end{subfigure}
\vspace{0.8em}
\begin{subfigure}{0.95\linewidth}
\centering
\includegraphics[width=\linewidth]{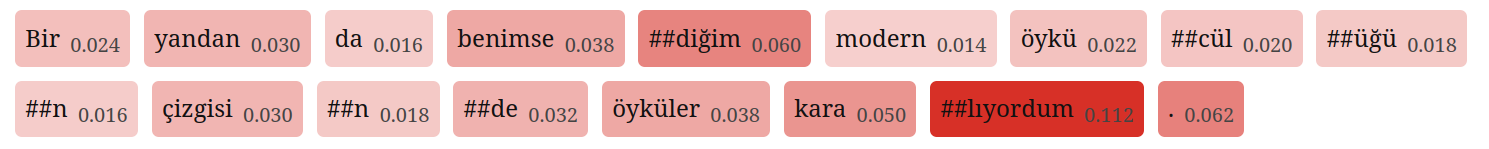}
\caption{32k-Medium: Attributions are sharp and linguistically aligned: highest saliency on the finite predicate chain (“kara + \#\#lıyordum”), with clear emphasis on the object (“öyküler”) and the oblique (“çizgisi + \#\#nde”), while possessive/genitive cues (“öykü \#\#cül \#\#üğü \#\#n”) are moderately weighted—matching strong LAS/POS with stable morphology at this vocabulary size.}
\end{subfigure}
\vspace{0.8em}
\begin{subfigure}{0.95\linewidth}
\centering
\includegraphics[width=\linewidth]{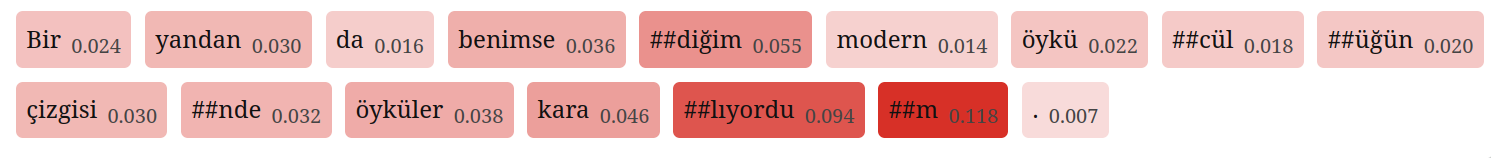}
\caption{52k-Alldata: Most balanced map: strong predicate focus alongside higher weights on possessive/genitive and case markers (“öykücülüğün”, “çizgisi+nde”) and plural object (“öyküler”), aligning with peak POS/LAS and stable morphology.}
\end{subfigure}
\vspace{0.8em}
\begin{subfigure}{0.95\linewidth}
\centering
\includegraphics[width=\linewidth]{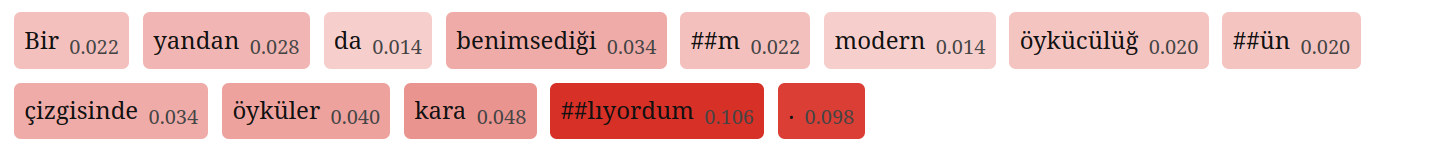}
\caption{128k-Medium: High confidence on the predicate (“karalıyordum”) but reduced visibility of non-verb morphology; larger subwords under-segment inflectional cues, consistent with slight Morph accuracy softening at very large vocabularies.}
\end{subfigure}
\caption{Merged POS–Dep–Morph token attributions for the sentence “Bir yandan da benimsediğim modern öykücülüğün çizgisinde öyküler karalıyordum.” We visualize a single task-aware heatmap per tokenizer by fusing head-specific Integrated Gradients over the shared encoder (loss-weighted combination; subword scores aggregated to words for display). As vocabulary size and training data increase from 2k-Minimal to 52k-Alldata, attributions become more focused and linguistically aligned: the predicate's inflectional chain remains the main evidence source while object, oblique, and possessive cues gain salience. At 128k, predicate salience persists but non-verb morphology attenuates, mirroring the observed plateau in POS/LAS and mild decline in Morph accuracy at very large vocabularies..}
\label{fig:pos-explain}
\end{figure}

Taken together, Figure \ref{fig:pos-explain} shows how a shared encoder concentrates evidence where the joint objective is most informative—on the finite predicate—while still surfacing syntactically and morphologically diagnostic cues on arguments and case/possession markers when the tokenizer makes them legible. As vocabularies move into the mid–large range, attribution becomes both crisper and better aligned with linguistic structure, paralleling gains in POS/LAS. Beyond that range, under-segmentation reduces the visibility of non‑verb morphology, echoing the small Morph dips we observe. Thus, explainability and accuracy trace the same curve: the 32k–52k band yields the most faithful, task-consistent rationales, while very large vocabularies trade legibility for marginal returns.

\subsection{Key Findings}
This section distills what the WordPiece tokenizer configuration (size and training data) does to Turkish Transformer performance, focusing on sequence length, morphology visibility, and downstream accuracy. We keep scope to BERT-style models trained with our WP variants; a later section compares word/char/subword families more broadly.

\noindent\textbf{RQ1 (Vocabulary size vs. task class).}
Mid–large WP vocabularies ($\approx$32k–52k) strike the best overall balance: they improve POS/LAS and maintain strong results on semantics-oriented benchmarks (e.g., SST-2/STS-like tasks). Very small vocabularies ($\leq$5k) over-fragment inputs, inflate sequences, and depress parsing and acceptability; very large vocabularies ($\geq$128k) under-segment morphology and yield diminishing or negative returns on morph-sensitive tasks.

\noindent\textbf{RQ2 (Morphological alignment helps syntax/morph more than semantics/NER).}
Higher boundary F1 and better lemma integrity correlate strongly with gains in POS/DEP/Morph and CoLA, while effects on sentiment/semantic tasks and NER are smaller or task-specific. NER benefits more from stable entity segmentation (reduced over-fragmentation of names/numbers) than from fine-grained morpheme boundaries, explaining a slightly higher optimal vocab on NER.

\noindent\textbf{RQ3 (Training data scale × vocabulary size}
Scaling tokenizer training data from 5→20→80 GB improves robustness and reduces variance once vocabulary is in the 32k–52k band. With tiny vocabularies, additional data does not overcome over-fragmentation; with huge vocabularies, more data cannot recover lost morphological legibility. The interaction is complementary: mixed-domain data plus a mid–large vocabulary delivers the most reliable gains.

\noindent\textbf{RQ4 (Pareto region of tokenization).}
There is a clear trade-off among (i) sequence length (favoring larger vocabularies), (ii) morphological fidelity (favoring smaller vocabularies), and (iii) downstream accuracy. The efficient operating region is $\approx$32k–52k on mixed-domain data: fertility remains moderate ($\approx$1.4–1.7), continuation sits in $\approx$0.30–0.45, sequences are compact, and POS/LAS and SST-like tasks are strong. Pushing beyond this band shortens sequences further but erodes morphology and hurts morph-sensitive tasks.

\noindent\textbf{Practical prescription for WordPiece tokenizers.}
\begin{itemize}
    \item Default for broad Turkish NLP: mixed-domain WP vocabulary in the 32k–52k band.
    \item Morph/grammar-sensitive pipelines (parsing, CoLA, morph tagging): favor the lower end ($\approx$20k–32k) or use morphology-aware constraints if available.
    \item NER-heavy applications: lean toward the higher end of the band ($\approx$32k–52k) or protect entity stems via tokenizer customization.
\end{itemize}

\noindent\textbf{Explainability alignment.}
Attribution analyses mirror these trends: mid–large vocabularies concentrate saliency on predicate morphology and syntactically diagnostic spans while keeping entities intact; tiny vocabularies diffuse saliency across long subword chains, and huge vocabularies hide inflectional cues. The configurations that peak on POS/LAS also yield the most faithful, task-consistent rationales.

In sum, mid–large WordPiece vocabularies paired with mixed-domain training data offer the most reliable accuracy–interpretability trade-off for Turkish Transformers, preserving inflectional cues that drive syntax and acceptability while maintaining compact sequences and strong entity handling.

\section{Optimal Ways of Tokenizing Turkish}
This section synthesizes what character, word, morphology-aware subword, and WordPiece tokenization offer for Turkish, with an emphasis on how each balances sequence length, morphology visibility, robustness, and downstream accuracy.

Character-level tokenization maximizes coverage by construction: there are no OOV tokens, and all inflectional material is explicit in the sequence. This yields strong recall of morphological cues and robustness to noise and spelling variants. The downside is severe fragmentation—sequences are long, training is slower, and attribution tends to diffuse across many characters. When compute is ample, character models can perform well on morphology-sensitive tasks, but their efficiency and long-range semantic modeling are comparatively weaker.

Word-level tokenization sits at the opposite extreme. Sequences are short and training/inference are fast; entity stability is excellent because names and numbers remain intact. However, for an agglutinative language like Turkish, the closed vocabulary leads to high OOV and sparse statistics on unseen inflections, making morphology opaque and hurting generalization across paradigms. In practice, word-level models can be competitive on NER with careful OOV handling, but they underperform on POS/DEP/CoLA where inflectional cues matter.

Morphology-aware subword tokenization explicitly aligns units with stems and affixes using analyzers or rules. This alignment improves boundary precision/recall and lemma integrity, yielding interpretable rationales and consistent gains on POS, dependency parsing, morphological tagging, and acceptability judgments. The trade-off is dependency on linguistic resources and some risk of over-segmentation for very frequent forms if frequency signals are not incorporated. Effects on semantics and NER are smaller unless the scheme is adapted to preserve entity stems.

WordPiece offers a data-driven middle path that can be tuned via vocabulary size. In our experiments, mid–large vocabularies (roughly 32k–52k on mixed-domain data) achieve the most reliable balance: sequences remain compact, fertility and continuation sit in favorable ranges, and accuracy is strong across syntax/morphology and semantic tasks. Very small vocabularies over-fragment and inflate sequences, while very large vocabularies over-merge and obscure morphology, both of which degrade performance on morph-sensitive benchmarks. NER often prefers the higher end of the mid–large range, where entities remain more intact.

Taken together, no single approach dominates across all tasks. Morphology-aware subwords provide the highest linguistic fidelity and the clearest explanations on morph-bearing spans. WordPiece, when kept in the mid–large regime and trained on mixed-domain data, delivers the most dependable end-to-end trade-off of accuracy, efficiency, and portability for Turkish, with small adjustments—toward smaller units for parsing and acceptability, or toward larger units for NER—depending on application needs.

\section{Conclusion}
We investigated how tokenizer design shapes Turkish Transformer performance, connecting subword granularity to both efficiency and linguistic fidelity. Across extensive experiments, a consistent picture emerges: mid–large WordPiece vocabularies trained on mixed-domain data offer the most reliable accuracy–efficiency trade-off, with fertility and continuation in morphology-visible ranges. This regime preserves inflectional cues critical for POS/DEP/morph tagging and acceptability, while maintaining strong results on semantic and NER benchmarks. Morphology-aware subword schemes can further boost morph-sensitive tasks and interpretability, though they require linguistic resources; character and word extremes provide robustness or speed, respectively, but underperform on one or more dimensions.

Practically, we recommend WordPiece in the 32k–52k band as a default for broad Turkish NLP, nudging smaller for grammar-centric pipelines and larger for NER-heavy applications. Beyond these prescriptions, our analysis framework—linking fragmentation metrics, morphology alignment, and attribution to downstream accuracy—offers a general template for tokenization choices in other morphologically rich languages. Future work includes extending the comparison to alternative subword learners, longer-context pretraining, and cross-lingual validation in related agglutinative families.

\paragraph{Acknowledgments}
We gratefully acknowledge support from the Google TPU Research Cloud program, which provided the compute resources used in this work.

\paragraph{Data Availability}
All datasets used in this study—including the pretraining corpus and benchmarking suites—are openly available under permissive licenses suitable for commercial use. 

\paragraph{Ethical Standards}
This research complies with relevant ethical guidelines and legal requirements in the jurisdictions of data collection and use. No personally identifiable information was processed beyond what is permitted by the data licenses.

\printendnotes

\defbibnote{preamble}{}

\printbibliography[prenote={preamble}]
\appendix

\section{Explainability Tools}
We used LIME \cite{ribeiro-etal-2016-trust} to produce local, model-agnostic explanations of tokenizer behavior by learning a sparse linear surrogate around each instance via perturbed inputs. Concretely, we generate perturbations by masking or merging subword units while preserving character order, query the model on these variants, and weight them by proximity to the original example. LIME then fits a simple interpretable model whose coefficients act as token-level attributions, indicating which subwords increase or decrease evidence for correct boundary alignment and related diagnostics (e.g., lemma consistency). We visualize the top-contributing subwords with signed weights to qualitatively inspect over-/under-segmentation and boundary errors across vocabulary sizes.

\section{Comprehensive Morphology Diagnostics by Vocabulary Size}
\label{appendix:morph-tables}

{
\setlength{\tabcolsep}{3pt} 
\begin{table}[ht!]
\centering
\small
\caption{Minimal corpus: morphology-aware tokenization diagnostics across vocabulary sizes. C.Noun and C.Verb abbreviate Common Noun and Common Verb. Sw/W denotes Subwords per Word. LSingle, LBoun, and ExMatch denote LemmaSingle, LemmaBoundary, and ExactMatch.}
\begin{tabular}{llrrrrrrrrrrrr}
\toprule
Vocab & Split   & Sw/W & P$\mu$ & R$\mu$ & F1$\mu$ & P$\mathrm{M}$ & R$\mathrm{M}$ & F1$\mathrm{M}$ & LSingle & LBoun & ExMatch & OverSeg & UnderSeg \\
\midrule
2k   & Çekimli & 18.17 & 0.31 & 1.00 & 0.47 & 0.31 & 1.00 & 0.47 & 0.00 & 1.00 & 0.00 & 3.37 & 0.31 \\
2k   & C.Noun  & 11.05 & 0.33 & 1.00 & 0.49 & 0.33 & 1.00 & 0.49 & 0.00 & 1.00 & 0.00 & 3.11 & 0.33 \\
2k   & C.Verb  & 12.70 & 0.35 & 1.00 & 0.52 & 0.36 & 1.00 & 0.53 & 0.00 & 1.00 & 0.00 & 2.88 & 0.36 \\
\midrule
5k   & Çekimli & 4.02  & 0.67 & 0.48 & 0.56 & 0.71 & 0.51 & 0.58 & 0.58 & 0.49 & 0.01 & 0.76 & 1.44 \\
5k   & C.Noun  & 2.33  & 0.79 & 0.52 & 0.63 & 0.83 & 0.53 & 0.64 & 0.96 & 0.19 & 0.04 & 0.66 & 1.63 \\
5k   & C.Verb  & 2.13  & 0.83 & 0.39 & 0.53 & 0.86 & 0.40 & 0.54 & 0.99 & 0.00 & 0.00 & 0.48 & 2.26 \\
\midrule
10k  & Çekimli & 5.15  & 0.61 & 0.56 & 0.59 & 0.63 & 0.58 & 0.59 & 0.32 & 0.58 & 0.01 & 0.98 & 1.12 \\
10k  & C.Noun  & 3.02  & 0.85 & 0.72 & 0.78 & 0.87 & 0.72 & 0.78 & 0.95 & 0.83 & 0.16 & 0.87 & 1.24 \\
10k  & C.Verb  & 2.86  & 0.74 & 0.48 & 0.58 & 0.77 & 0.49 & 0.59 & 0.99 & 0.13 & 0.02 & 0.65 & 1.64 \\
\midrule
20k  & Çekimli & 6.96  & 0.56 & 0.69 & 0.62 & 0.57 & 0.70 & 0.62 & 0.08 & 0.68 & 0.01 & 1.32 & 0.82 \\
20k  & C.Noun  & 4.07  & 0.66 & 0.74 & 0.70 & 0.69 & 0.75 & 0.71 & 0.50 & 0.66 & 0.12 & 1.15 & 0.93 \\
20k  & C.Verb  & 4.04  & 0.78 & 0.71 & 0.75 & 0.80 & 0.72 & 0.75 & 0.96 & 0.95 & 0.11 & 0.92 & 1.13 \\
\midrule
32k  & Çekimli & 4.77  & 0.64 & 0.54 & 0.58 & 0.66 & 0.56 & 0.59 & 0.39 & 0.56 & 0.01 & 0.90 & 1.21 \\
32k  & C.Noun  & 2.81  & 0.82 & 0.64 & 0.72 & 0.84 & 0.66 & 0.72 & 0.96 & 0.59 & 0.12 & 0.80 & 1.34 \\
32k  & C.Verb  & 2.48  & 0.81 & 0.45 & 0.58 & 0.83 & 0.46 & 0.58 & 0.99 & 0.08 & 0.02 & 0.57 & 1.90 \\
\midrule
52k  & Çekimli & 4.49  & 0.65 & 0.52 & 0.58 & 0.68 & 0.54 & 0.59 & 0.46 & 0.54 & 0.01 & 0.85 & 1.29 \\
52k  & C.Noun  & 2.64  & 0.77 & 0.57 & 0.66 & 0.80 & 0.59 & 0.67 & 0.96 & 0.41 & 0.08 & 0.75 & 1.43 \\
52k  & C.Verb  & 2.36  & 0.79 & 0.42 & 0.55 & 0.83 & 0.43 & 0.56 & 0.99 & 0.02 & 0.01 & 0.54 & 2.01 \\
\midrule
128k & Çekimli & 6.02  & 0.60 & 0.65 & 0.63 & 0.61 & 0.66 & 0.62 & 0.17 & 0.63 & 0.01 & 1.15 & 0.95 \\
128k & C.Noun  & 3.34  & 0.85 & 0.79 & 0.82 & 0.87 & 0.80 & 0.82 & 0.83 & 0.88 & 0.22 & 0.95 & 1.11 \\
128k & C.Verb  & 3.51  & 0.80 & 0.63 & 0.70 & 0.81 & 0.64 & 0.71 & 0.98 & 0.60 & 0.04 & 0.80 & 1.31 \\
\bottomrule
\end{tabular}
\label{tab:minimal-results}
\end{table}
}

{
\setlength{\tabcolsep}{3pt} 
\begin{table}[ht!]
\centering
\small
\caption{Medium corpus: morphology-aware tokenization diagnostics across vocabulary sizes. C.Noun and C.Verb abbreviate Common Noun and Common Verb. Sw/W denotes Subwords per Word. LSingle, LBoun, and ExMatch denote LemmaSingle, LemmaBoundary, and ExactMatch.}
\begin{tabular}{llrrrrrrrrrrrr}
\toprule
Vocab & Split & Sw/W & P$\mu$ & R$\mu$ & F1$\mu$ & P$\mathrm{M}$ & R$\mathrm{M}$ & F1$\mathrm{M}$ & LSingle & LBoun & ExMatch & OverSeg & UnderSeg \\
\midrule
2k   & Çekimli & 18.17 & 0.31 & 1.00 & 0.47 & 0.31 & 1.00 & 0.47 & 0.00 & 1.00 & 0.00 & 3.37 & 0.31 \\
2k   & C.Noun  & 11.05 & 0.33 & 1.00 & 0.49 & 0.33 & 1.00 & 0.49 & 0.00 & 1.00 & 0.00 & 3.11 & 0.33 \\
2k   & C.Verb  & 12.70 & 0.35 & 1.00 & 0.52 & 0.36 & 1.00 & 0.53 & 0.00 & 1.00 & 0.00 & 2.88 & 0.36 \\
\midrule
5k   & Çekimli & 3.91  & 0.68 & 0.48 & 0.56 & 0.71 & 0.50 & 0.57 & 0.59 & 0.47 & 0.01 & 0.74 & 1.48 \\
5k   & C.Noun  & 2.36  & 0.80 & 0.52 & 0.63 & 0.82 & 0.54 & 0.64 & 0.96 & 0.22 & 0.05 & 0.67 & 1.60 \\
5k   & C.Verb  & 2.06  & 0.84 & 0.39 & 0.53 & 0.87 & 0.40 & 0.54 & 0.99 & 0.01 & 0.00 & 0.47 & 2.27 \\
\midrule
10k  & Çekimli & 5.46  & 0.60 & 0.58 & 0.59 & 0.62 & 0.60 & 0.60 & 0.25 & 0.60 & 0.01 & 1.04 & 1.05 \\
10k  & C.Noun  & 3.16  & 0.85 & 0.75 & 0.80 & 0.87 & 0.76 & 0.80 & 0.83 & 0.88 & 0.19 & 0.90 & 1.17 \\
10k  & C.Verb  & 3.14  & 0.77 & 0.54 & 0.64 & 0.78 & 0.55 & 0.64 & 0.99 & 0.30 & 0.03 & 0.72 & 1.48 \\
\midrule
20k  & Çekimli & 18.17 & 0.31 & 1.00 & 0.47 & 0.31 & 1.00 & 0.47 & 0.00 & 1.00 & 0.00 & 3.37 & 0.31 \\
20k  & C.Noun  & 11.05 & 0.33 & 1.00 & 0.49 & 0.33 & 1.00 & 0.49 & 0.00 & 1.00 & 0.00 & 3.11 & 0.33 \\
20k  & C.Verb  & 12.70 & 0.35 & 1.00 & 0.52 & 0.36 & 1.00 & 0.53 & 0.00 & 1.00 & 0.00 & 2.88 & 0.36 \\
\midrule
32k  & Çekimli & 4.84  & 0.63 & 0.54 & 0.58 & 0.65 & 0.56 & 0.59 & 0.37 & 0.56 & 0.01 & 0.92 & 1.19 \\
32k  & C.Noun  & 2.83  & 0.86 & 0.68 & 0.76 & 0.87 & 0.69 & 0.76 & 0.95 & 0.62 & 0.14 & 0.81 & 1.32 \\
32k  & C.Verb  & 2.69  & 0.72 & 0.43 & 0.54 & 0.75 & 0.45 & 0.55 & 0.99 & 0.05 & 0.02 & 0.61 & 1.74 \\
\midrule
52k  & Çekimli & 4.47  & 0.64 & 0.51 & 0.57 & 0.67 & 0.53 & 0.58 & 0.45 & 0.53 & 0.01 & 0.85 & 1.29 \\
52k  & C.Noun  & 2.65  & 0.80 & 0.59 & 0.68 & 0.82 & 0.61 & 0.69 & 0.96 & 0.50 & 0.08 & 0.76 & 1.42 \\
52k  & C.Verb  & 2.41  & 0.79 & 0.42 & 0.55 & 0.81 & 0.44 & 0.56 & 0.99 & 0.03 & 0.01 & 0.55 & 1.95 \\
\midrule
128k & Çekimli & 18.17 & 0.31 & 1.00 & 0.47 & 0.31 & 1.00 & 0.47 & 0.00 & 1.00 & 0.00 & 3.37 & 0.31 \\
128k & C.Noun  & 11.05 & 0.33 & 1.00 & 0.49 & 0.33 & 1.00 & 0.49 & 0.00 & 1.00 & 0.00 & 3.11 & 0.33 \\
128k & C.Verb  & 12.70 & 0.35 & 1.00 & 0.52 & 0.36 & 1.00 & 0.53 & 0.00 & 1.00 & 0.00 & 2.88 & 0.36 \\
\bottomrule
\end{tabular}
\label{tab:medium-results}
\end{table}
}

{
\setlength{\tabcolsep}{3pt} 
\begin{table}[ht!]
\centering
\small
\caption{alldata corpus: morphology-aware tokenization diagnostics across vocabulary sizes. C.Noun and C.Verb abbreviate Common Noun and Common Verb. Sw/W denotes Subwords per Word. LSingle, LBoun, and ExMatch denote LemmaSingle, LemmaBoundary, and ExactMatch.}
\begin{tabular}{llrrrrrrrrrrrr}
\toprule
Vocab & Split & Sw/W & P$\mu$ & R$\mu$ & F1$\mu$ & P$\mathrm{M}$ & R$\mathrm{M}$ & F1$\mathrm{M}$ & LSingle & LBoun & ExMatch & OverSeg & UnderSeg \\
\midrule
2k   & Çekimli & 18.17 & 0.31 & 1.00 & 0.47 & 0.31 & 1.00 & 0.47 & 0.00 & 1.00 & 0.00 & 3.37 & 0.31 \\
2k   & C.Noun  & 11.05 & 0.33 & 1.00 & 0.49 & 0.33 & 1.00 & 0.49 & 0.00 & 1.00 & 0.00 & 3.11 & 0.33 \\
2k   & C.Verb  & 12.70 & 0.35 & 1.00 & 0.52 & 0.36 & 1.00 & 0.53 & 0.00 & 1.00 & 0.00 & 2.88 & 0.36 \\
\midrule
5k   & Çekimli & 4.01  & 0.68 & 0.49 & 0.57 & 0.71 & 0.51 & 0.58 & 0.56 & 0.48 & 0.01 & 0.76 & 1.44 \\
5k   & C.Noun  & 2.37  & 0.83 & 0.54 & 0.66 & 0.85 & 0.56 & 0.66 & 0.95 & 0.22 & 0.06 & 0.68 & 1.59 \\
5k   & C.Verb  & 2.14  & 0.83 & 0.40 & 0.54 & 0.86 & 0.41 & 0.55 & 0.99 & 0.01 & 0.01 & 0.49 & 2.19 \\
\midrule
10k  & Çekimli & 7.85  & 0.54 & 0.75 & 0.63 & 0.54 & 0.76 & 0.62 & 0.04 & 0.70 & 0.00 & 1.49 & 0.72 \\
10k  & C.Noun  & 5.12  & 0.55 & 0.78 & 0.64 & 0.56 & 0.78 & 0.65 & 0.12 & 0.69 & 0.02 & 1.46 & 0.73 \\
10k  & C.Verb  & 4.86  & 0.75 & 0.82 & 0.79 & 0.77 & 0.83 & 0.79 & 0.69 & 0.90 & 0.18 & 1.11 & 0.94 \\
\midrule
20k  & Çekimli & 18.17 & 0.31 & 1.00 & 0.47 & 0.31 & 1.00 & 0.47 & 0.00 & 1.00 & 0.00 & 3.37 & 0.31 \\
20k  & C.Noun  & 11.05 & 0.33 & 1.00 & 0.49 & 0.33 & 1.00 & 0.49 & 0.00 & 1.00 & 0.00 & 3.11 & 0.33 \\
20k  & C.Verb  & 12.70 & 0.35 & 1.00 & 0.52 & 0.36 & 1.00 & 0.53 & 0.00 & 1.00 & 0.00 & 2.88 & 0.36 \\
\midrule
32k  & Çekimli & 5.33  & 0.60 & 0.57 & 0.59 & 0.62 & 0.59 & 0.59 & 0.28 & 0.59 & 0.01 & 1.01 & 1.08 \\
32k  & C.Noun  & 3.05  & 0.86 & 0.72 & 0.79 & 0.88 & 0.74 & 0.79 & 0.95 & 0.88 & 0.17 & 0.87 & 1.23 \\
32k  & C.Verb  & 3.10  & 0.77 & 0.53 & 0.63 & 0.78 & 0.55 & 0.63 & 0.99 & 0.36 & 0.03 & 0.71 & 1.51 \\
\midrule
52k  & Çekimli & 4.71  & 0.63 & 0.53 & 0.57 & 0.65 & 0.55 & 0.58 & 0.40 & 0.55 & 0.01 & 0.89 & 1.23 \\
52k  & C.Noun  & 2.87  & 0.78 & 0.62 & 0.70 & 0.81 & 0.64 & 0.70 & 0.95 & 0.55 & 0.10 & 0.82 & 1.31 \\
52k  & C.Verb  & 2.57  & 0.80 & 0.46 & 0.58 & 0.82 & 0.47 & 0.59 & 0.99 & 0.04 & 0.01 & 0.59 & 1.83 \\
\midrule
128k & Çekimli & 18.17 & 0.31 & 1.00 & 0.47 & 0.31 & 1.00 & 0.47 & 0.00 & 1.00 & 0.00 & 3.37 & 0.31 \\
128k & C.Noun  & 11.05 & 0.33 & 1.00 & 0.49 & 0.33 & 1.00 & 0.49 & 0.00 & 1.00 & 0.00 & 3.11 & 0.33 \\
128k & C.Verb  & 12.70 & 0.35 & 1.00 & 0.52 & 0.36 & 1.00 & 0.53 & 0.00 & 1.00 & 0.00 & 2.88 & 0.36 \\
\bottomrule
\end{tabular}
\label{tab:alldata-results}
\end{table}
}

\end{document}